\documentclass[10pt,twocolumn,letterpaper]{article}

\usepackage{wacv}              

\usepackage{graphicx}
\usepackage{amsmath}
\usepackage{amssymb}
\usepackage{booktabs}
\usepackage{multicol}
\usepackage{algpseudocode}
\usepackage{algorithm}
\usepackage[accsupp]{axessibility}  
\usepackage{array}
\usepackage{diagbox}
\usepackage{wrapfig}
\usepackage{float}
\usepackage{xcolor}
\usepackage{caption}

\usepackage[pagebackref,breaklinks,colorlinks]{hyperref}

\usepackage[capitalize]{cleveref}
\crefname{section}{Sec.}{Secs.}
\Crefname{section}{Section}{Sections}
\Crefname{table}{Table}{Tables}
\crefname{table}{Tab.}{Tabs.}

\def\wacvPaperID{241} 
\def\confName{WACV}
\def\confYear{2025}

\begin{document}

\title{DreamBlend: Advancing Personalized Fine-tuning of \\ Text-to-Image Diffusion Models}

\author{%
    Shwetha Ram, 
    Tal Neiman, 
    Qianli Feng, 
    Andrew Stuart, 
    Son Tran, 
    Trishul Chilimbi \\
    Amazon \\
    {\tt\small \{shweram, taneiman, fengq, andrxstu, sontran, trishulc\}@amazon.com}
}

\twocolumn[{%
\renewcommand\twocolumn[1][]{#1}%
\maketitle
\begin{center}
    \centering
    \vspace{-20pt} 
    \captionsetup{type=figure}
    \includegraphics[width=0.8\linewidth]{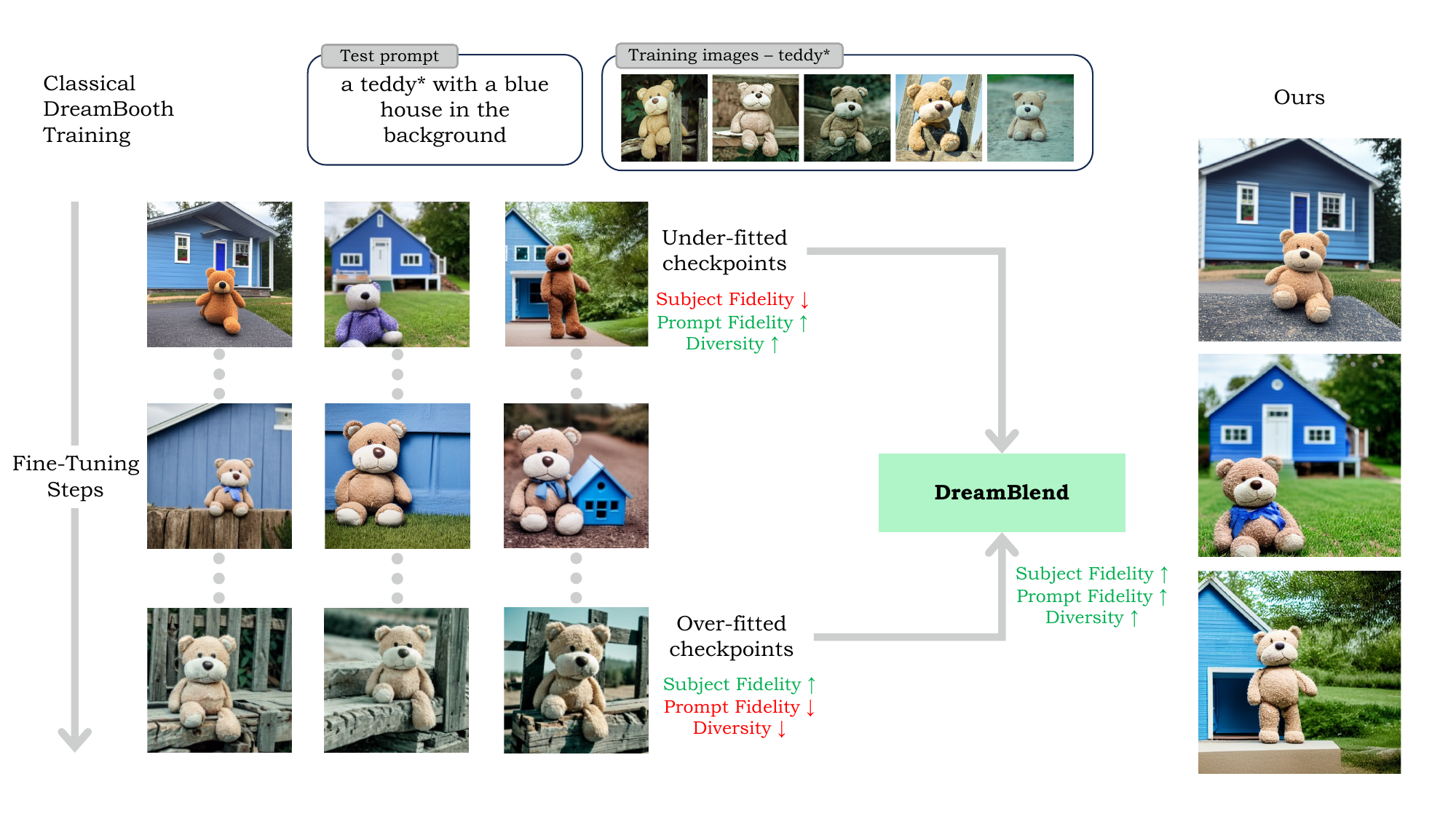}
    \caption{DreamBlend merges prompt fidelity and diversity from underfit checkpoints with subject fidelity from overfit checkpoints during image generation. The generated images have the layout of the underfit images and the subject fidelity of the overfit images, achieving better subject fidelity, prompt fidelity and diversity.}
    \label{fig:dreamblend}
\end{center}%
}]

\begin{abstract}
   Given a small number of images of a subject, personalized image generation techniques can fine-tune large pre-trained text-to-image diffusion models to generate images of the subject in novel contexts, conditioned on text prompts. In doing so, a trade-off is made between prompt fidelity, subject fidelity and diversity. As the pre-trained model is fine-tuned, earlier checkpoints synthesize images with low subject fidelity but high prompt fidelity and diversity. In contrast, later checkpoints generate images with low prompt fidelity and diversity but high subject fidelity. This inherent trade-off limits the prompt fidelity, subject fidelity and diversity of generated images. In this work, we propose \emph{DreamBlend} to combine the prompt fidelity from earlier checkpoints and the subject fidelity from later checkpoints during inference. We perform a cross attention guided image synthesis from a later checkpoint, guided by an image generated by an earlier checkpoint, for the same prompt. This enables generation of images with better subject fidelity, prompt fidelity and diversity on challenging prompts, outperforming state-of-the-art fine-tuning methods.
\end{abstract}

\section{Introduction}
\label{sec:intro}

\begin{figure*}
\centering
\begin{tabular}{c|c} 
  \begin{tabular}{@{}c@{}c@{}c@{}}
    \includegraphics[width=0.08\textwidth]{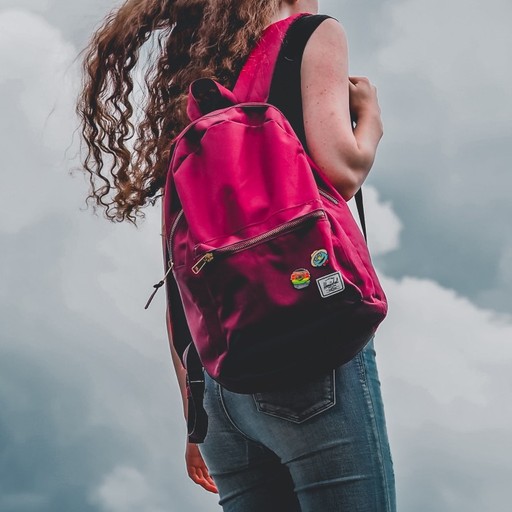} &
    \includegraphics[width=0.08\textwidth]{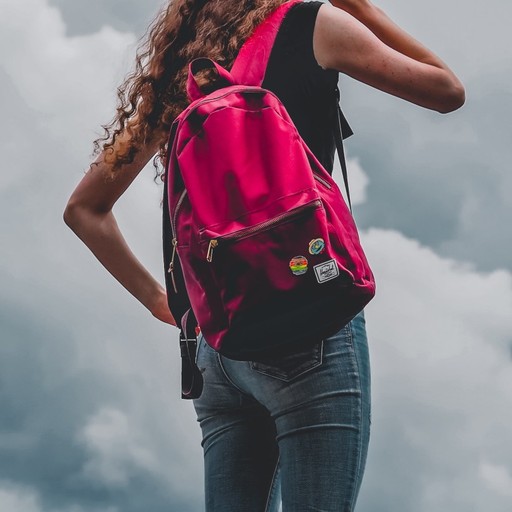} \\
    \includegraphics[width=0.08\textwidth]{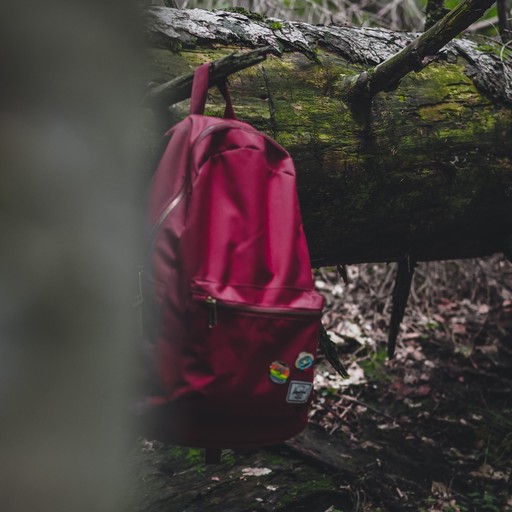} &
    \includegraphics[width=0.08\textwidth]{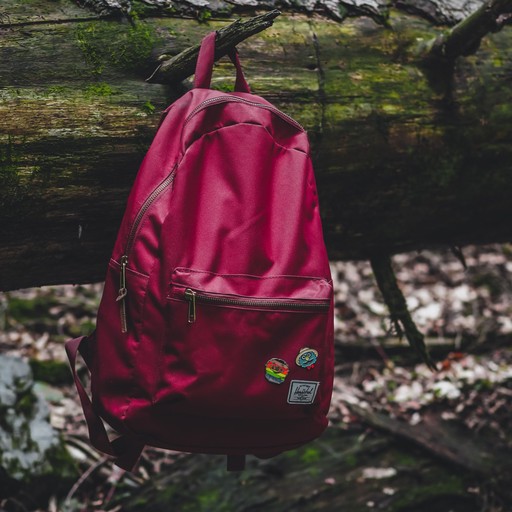} \\
    {Input} & {$backpack^*$}  \\
  \end{tabular}
&
\begin{tabular}{@{}c@{\hspace{2pt}}c@{\hspace{2pt}}c@{\hspace{2pt}}c@{\hspace{2pt}}c@{\hspace{2pt}}c@{\hspace{2pt}}c@{\hspace{2pt}}c@{\hspace{2pt}}c@{\hspace{2pt}}c@{\hspace{2pt}}c@{}}
    \includegraphics[width=0.08\textwidth]{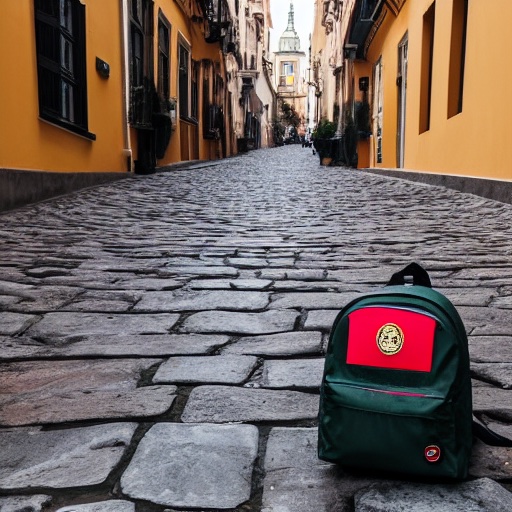} &
    \includegraphics[width=0.08\textwidth]{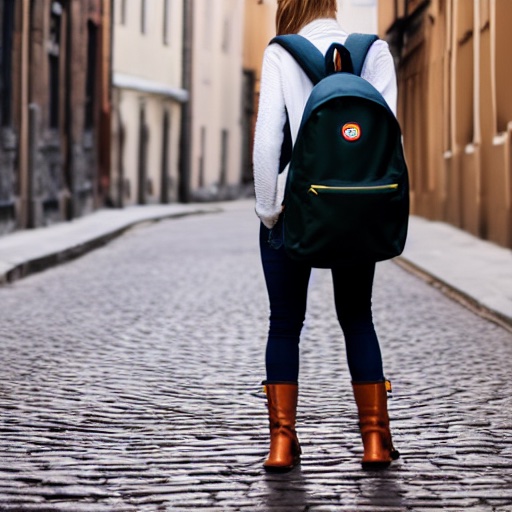} &
    \includegraphics[width=0.08\textwidth]{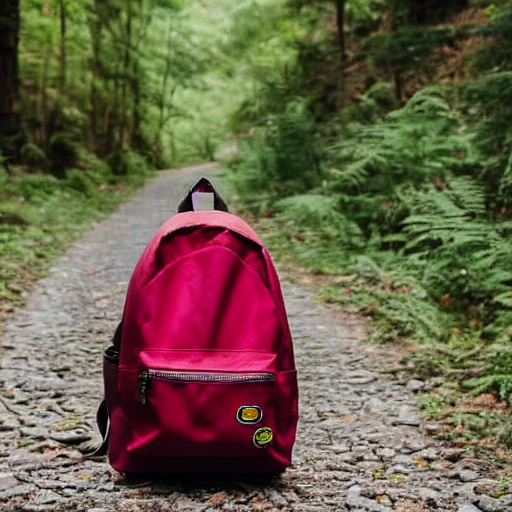} &
    \includegraphics[width=0.08\textwidth]{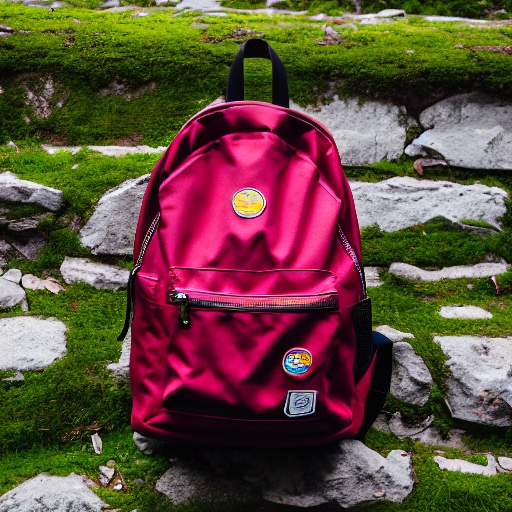} &
    \includegraphics[width=0.08\textwidth]{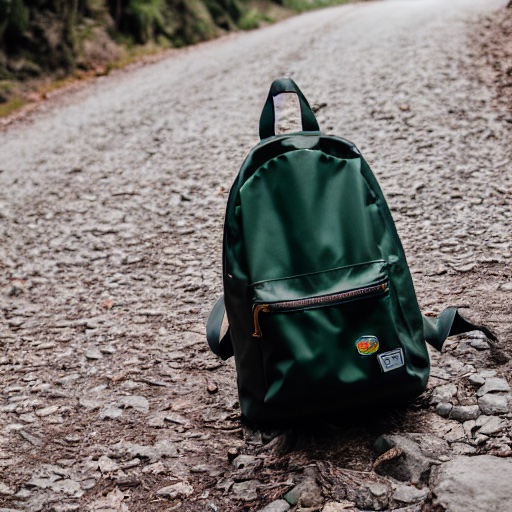} &
    \includegraphics[width=0.08\textwidth]{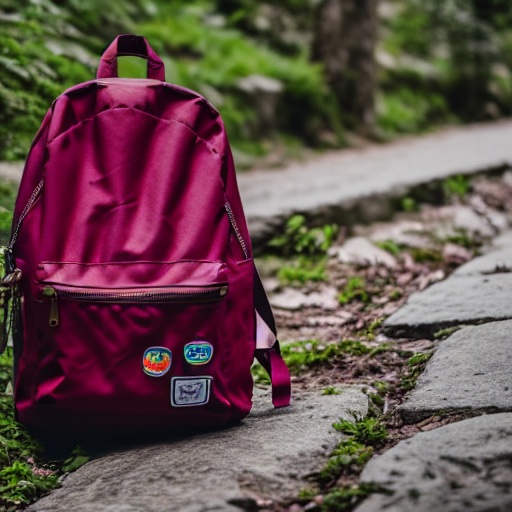} &
    \includegraphics[width=0.08\textwidth]{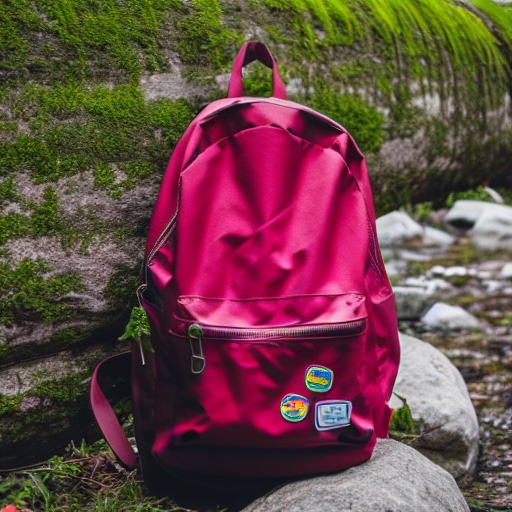} &
    \includegraphics[width=0.08\textwidth]{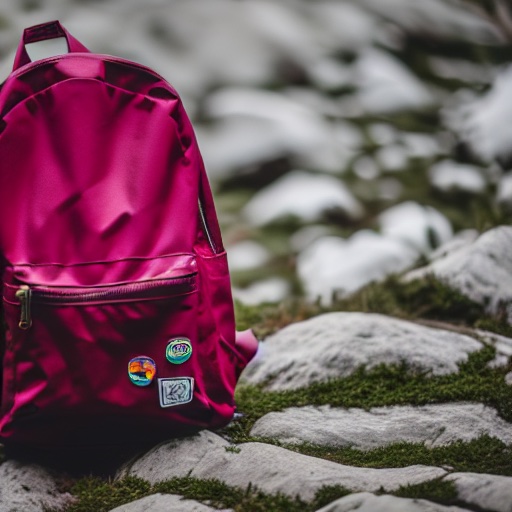} & \dots &
    \includegraphics[width=0.08\textwidth]{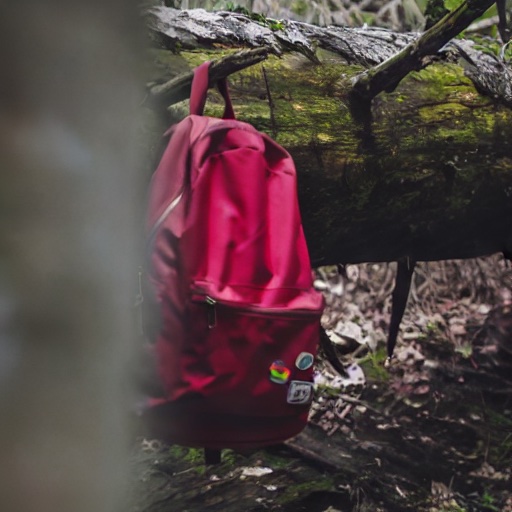} \\

    \includegraphics[width=0.08\textwidth]{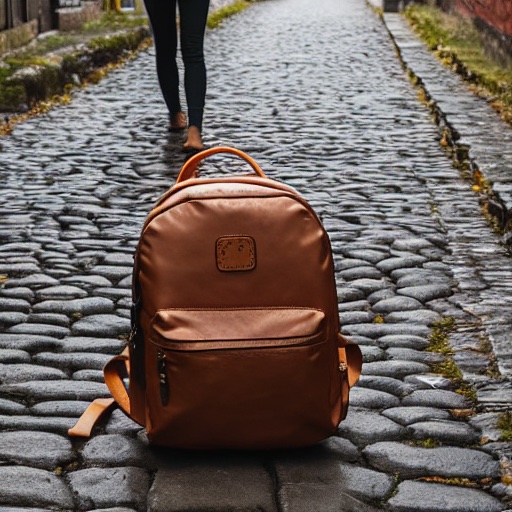} &
    \includegraphics[width=0.08\textwidth]{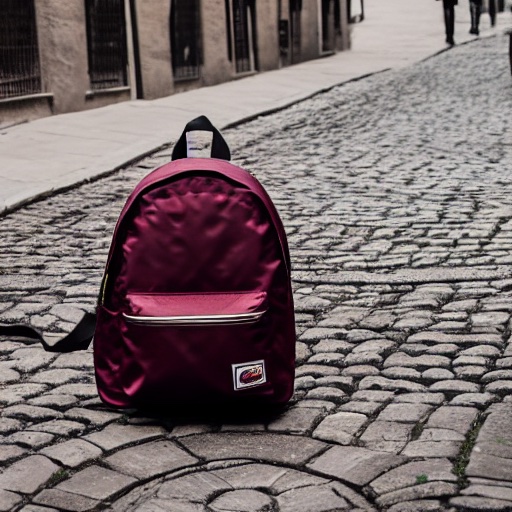} &
    \includegraphics[width=0.08\textwidth]{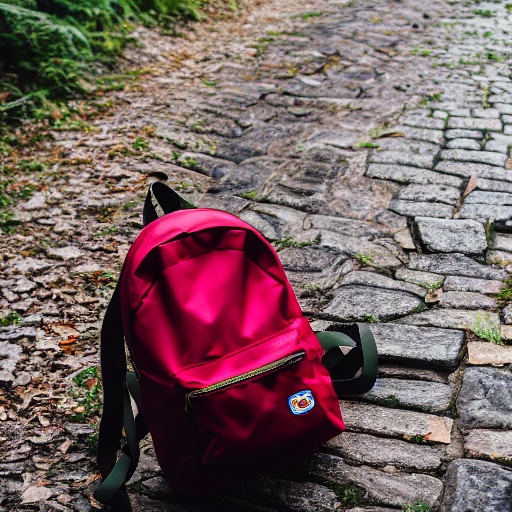} &
    \includegraphics[width=0.08\textwidth]{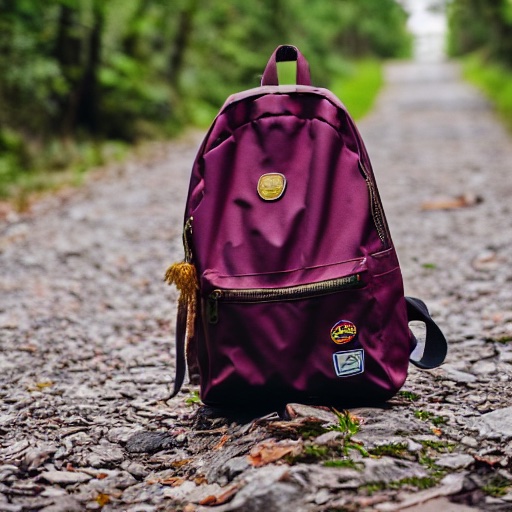} &
    \includegraphics[width=0.08\textwidth]{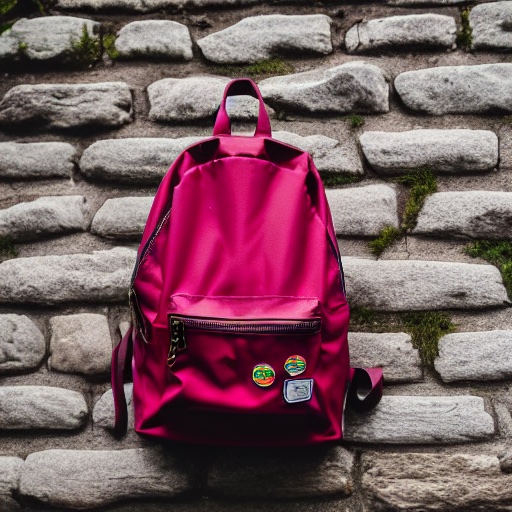} &
    \includegraphics[width=0.08\textwidth]{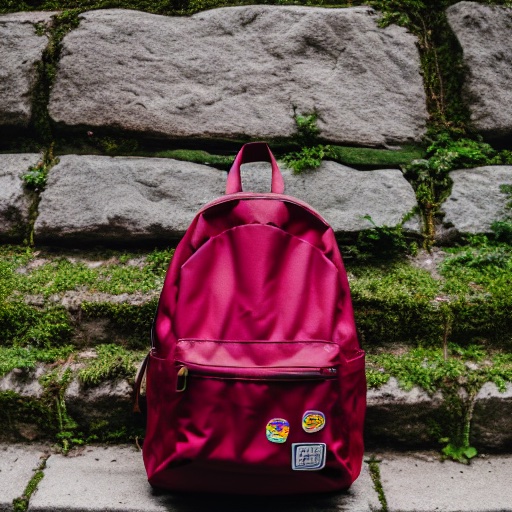} &
    \includegraphics[width=0.08\textwidth]{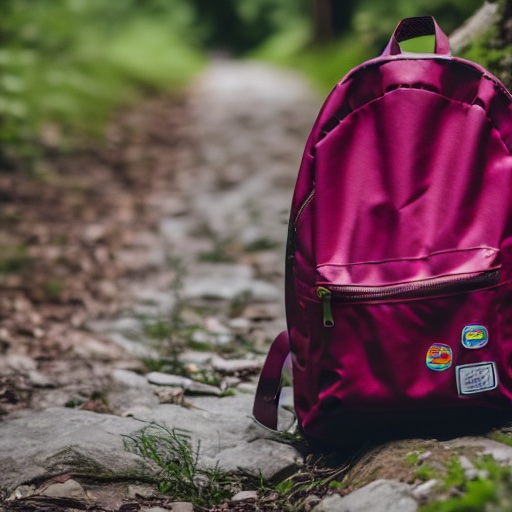} &
    \includegraphics[width=0.08\textwidth]{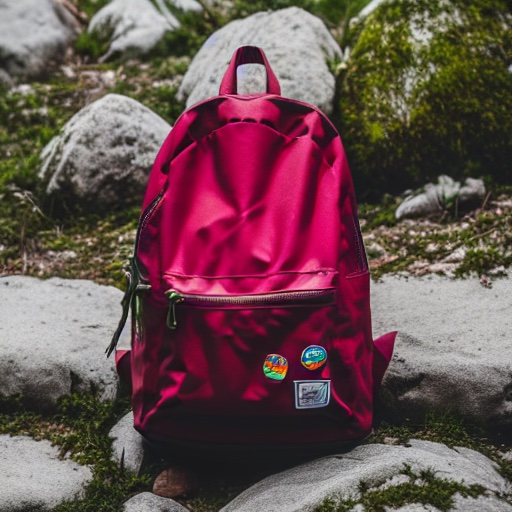} & \dots &
    \includegraphics[width=0.08\textwidth]{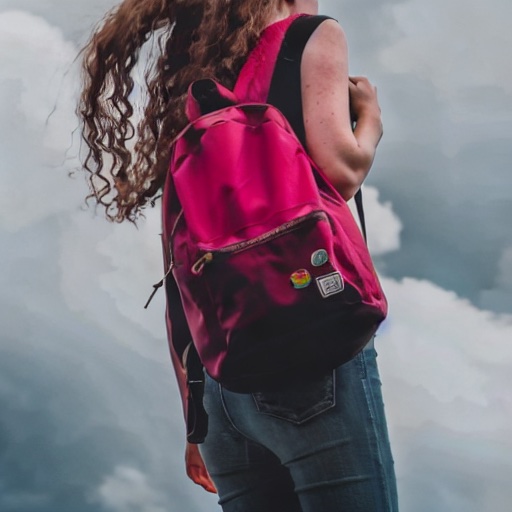} \\

    {5} & {25} & {50} & {75} & {100} & {125} & {150} & {175} & {} & {1000}\\
\end{tabular} 
\end{tabular}

\caption{Images generated by different checkpoints for prompt: `a $backpack^*$ on a cobblestone street' as a SD model is fine-tuned from 5 to 1000 steps. Early checkpoints have higher prompt fidelity and diversity but lower subject fidelity while later checkpoints have higher subject fidelity but lower prompt fidelity and diversity. At step=1000, the model reproduces the input images used for fine-tuning.}
\label{fig:backpacking-overfitting}
\vspace{-10pt}

\end{figure*}

Text-to-Image models like Stable Diffusion\cite{rombach2022high} enable generating images from text prompts with broad capabilities. However, users often seek personalized image generation for specific subjects in diverse contexts. This involves synthesizing novel images conditioned on text prompts using a few subject images. For example, in \cref{fig:dreamblend}, we generate images of subject $teddy^*$ ``with a blue house in the background''. Successful image generation hinges on two criteria: subject fidelity, ensuring the $teddy^*$ in generated images is ``same'' as that in input training images, and prompt fidelity, adhering to the text prompt of a blue house in the background. Additionally, diversity is desired, showcasing $teddy^*$ in varied poses and environments with diverse elements like driveways, lawns and trees.

Prior works have embedded subject identity in an input word embedding\cite{gal2022image}, layers of a pre-trained model through fine-tuning\cite{ruiz2023dreambooth}, or both\cite{kumari2023multi}. Fine-tuning based approaches pose inherent trade-offs between subject fidelity, prompt fidelity and diversity, presented in \cref{fig:dreamblend}. A pre-trained model excels at generating high-fidelity images for various prompts by leveraging its world knowledge. Fine-tuning enhances subject fidelity but reduces prompt fidelity and diversity due to overfitting, language drift and catastrophic forgetting. Over time, generated images closely resemble the training inputs as shown in \cref{fig:backpacking-overfitting}.

Existing methods seek a ``sweet spot" between prompt fidelity, subject fidelity, and diversity using intermediate checkpoints. We introduce \emph{DreamBlend}, which merges the strengths of early and late checkpoints during image generation. This approach provides superior trade-offs by combining early checkpoint prompt fidelity and diversity with late checkpoint subject fidelity. A careful study of image generation in early and later checkpoints reveals the phenomenon of catastrophic attention collapse in later checkpoints. This leads to our key insight of using cross attention guidance to preserve the prompt fidelity in early checkpoint images. By synthesizing images from a later checkpoint guided by an image from an earlier one for the same prompt, we achieve improved subject fidelity, prompt fidelity and diversity, as demonstrated in \cref{fig:dreamblend}.

In summary, our contributions are as follows: first, we study different operating points in current finetuning-based text-to-image personalization methods and observe a catastrophic attention collapse in later checkpoints that diminishes prompt fidelity and diversity; second, we propose a novel approach, \emph{DreamBlend}, of using cross attention guidance to combine the prompt fidelity and diversity of early checkpoints with the subject fidelity of later checkpoints during image generation and find that this regularization on cross attention maps is effective at minimizing the effect of over-fitting; finally, we demonstrate that \emph{DreamBlend} produces superior images with enhanced subject fidelity, prompt fidelity, and diversity on challenging prompts, surpassing existing state-of-the-art fine-tuning based methods.

\section{Related work}
\subsection{Text-to-image diffusion models}
Text-to-image diffusion models are trained to generate samples from a conditional data distribution by the gradual denoising of a variable sampled from a Gaussian distribution. Recent progress in text-to-image synthesis has been fuelled by large models trained on web scale data \cite{rombach2022high, saharia2022photorealistic, ramesh2021zero, sun2023generative, ramesh2022hierarchical, yu2022scaling}. We leverage the capabilities of such models for personalized image generation.

\subsection{Personalized text-to-image diffusion models}
Generating specific subjects with pre-trained text-to-image models via prompt engineering is difficult unless those subjects were well-represented in the training data. Consequently, efforts have emerged to teach specific subjects post-training \cite{gal2022image, gal2023encoder, ruiz2023dreambooth, ruiz2024hyperdreambooth, kumari2023multi, li2024blip, ye2023ip, ma2023unified, wei2023elite, tewel2023key, hao2023vico, alaluf2023neural, voynov2023p+, wang2024instantid, jia2023taming, Shi_2024_CVPR,zhang2024attention}.

New subject identities can be added through input word embeddings \cite{gal2022image, voynov2023p+, alaluf2023neural, wei2023elite}, fine-tuning model weights \cite{ruiz2023dreambooth} or both \cite{kumari2023multi}. While fine-tuning often provides better subject fidelity and photorealism due to greater expressive power, it suffers from loss of prompt fidelity and diversity as fine-tuning continues. This can be attributed to over-fitting and language drift, observed in both language models \cite{lu2020countering, lee2019countering} and text-to-image diffusion models \cite{ruiz2023dreambooth, kumari2023multi, tewel2023key}. To alleviate this, DreamBooth \cite{ruiz2023dreambooth} regularizes the network with its own generated images while Custom Diffusion \cite{kumari2023multi} fine-tunes only text-image cross attention weights and uses class-specific retrieved images for regularization. Perfusion \cite{tewel2023key} performs a gated rank one update inspired by ROME \cite{Meng2022LocatingAE} on key and value projection matrices, locking the key matrices to the subject's super category to reduce over-fitting. Our approach differs by using cross-attention guidance to synthesize images from an overfit checkpoint, guided by an image from an underfit checkpoint for the same prompt.

Encoder-based approaches \cite{Shi_2024_CVPR, gal2023encoder, jia2023taming, chen2024subject, ye2023ip, wang2024instantid, ma2023unified} aim to avoid fine-tuning and storing weights for each subject. Some methods \cite{gal2023encoder, jia2023taming, Shi_2024_CVPR, wang2024instantid} are limited to specific domains like dogs or human faces and may still require some fine-tuning for personalization \cite{gal2023encoder}, while others \cite{li2024blip, chen2024subject} are more generalized with zero-shot capabilities. They often require large scale pre-training with the diffusion model in the loop, with some methods \cite{chen2024subject} harvesting the pre-training data from expert models that are fine-tuned for each subject. Fine-tuning methods provide a cost-effective means to adapt existing text-to-image models for personalized image generation. They also contribute to generating high-quality data for training models with better inference efficiency. Consequently, our focus lies in enhancing the image quality produced by these fine-tuning based approaches.

\subsection{Image editing with diffusion models}
Image editing aims to modify specific regions of an input image while preserving the rest. Approaches like SDEdit \cite{meng2021sdedit} introduce stochastic noise followed by denoising, which can unintentionally alter non-targeted areas. Using a spatial mask from another model \cite{luddecke2022image} or the diffusion model itself \cite{couairon2022diffedit} to alleviate this often leads to content in the mask region being ignored and blending artifacts. In contrast to stochastic methods like DDPM \cite{ho2020denoising} and SDEdit \cite{meng2021sdedit}, deterministic DDIM inversion \cite{song2020denoising} first inverts an image for subsequent editing. Several schemes \cite{mokady2023null, parmar2023zero, wallace2023edict} have been designed to achieve a more editable reconstruction. Recent works \cite{li2023dreamedit, choi2023custom} propose personalized editing of real images using personalized text-to-image models. Choi et al. \cite{choi2023custom} combine methods from \cite{mokady2023null} and \cite{hertz2022prompt}, while Li et al. \cite{li2023dreamedit} iterate image inpainting with \cite{li2023gligen}, guided by spatial segmentation mask. Our work uses DDIM inversion for the initial latent and cross attention guidance for layout control.

Recent research \cite{hertz2022prompt, chefer2023attend, parmar2023zero, epstein2023diffusion, gu2024photoswap} shows that text-image cross attention maps significantly influence generated image layout. Perfusion and Custom Diffusion only update the text-image cross attention weights. While this affects the cross attention maps generated, they do not directly manipulate cross attention. Attend-and-Excite \cite{chefer2023attend} encourages generation of all subjects in the text prompt by enhancing attention values for the most neglected subject token at each time step. Prompt-to-Prompt \cite{hertz2022prompt} directly swaps the attention maps from source image generation into target image generation, for text-driven image editing. Photoswap \cite{gu2024photoswap} advocates swapping the self attention maps in addition to cross attention maps for superior layout control. In our work, we employ a cross attention guidance regularization to align target cross attention maps with reference maps from an underfit model. Compared to image editing, we care less about preserving the exact details in the underfit reference image and this formulation allows slight layout changes for improved subject fidelity. 

\section{Method}
\label{sec:method}

During fine-tuning of a pre-trained text-to-image diffusion model, early checkpoints produce images with high prompt fidelity and diversity but low subject fidelity. In contrast, later checkpoints yield images with high subject fidelity but lower prompt fidelity and diversity. Early checkpoints are under-fitted, lacking sufficient subject learning, while later ones are over-fitted to the subject appearance, pose and environments in the input images. A na\"ive way to combine the best of both would involve overlaying a subject from a later checkpoint onto an image from an earlier checkpoint. This method would likely introduce artifacts due to mismatches in pose, lighting, and other factors. However, this thought experiment inspires new methodologies aimed at integrating the advantages of both under-fitted and over-fitted checkpoints effectively.

    \begin{figure*}
    \centering
    \begin{tabular}{|c|*{5}{>{\centering\arraybackslash}m{2cm}|}}
    \hline
        \diagbox[width=2.1cm,height=1.2cm]{\textbf{tokens}}{\textbf{steps}} &
        \textbf{step 5} &
        \textbf{step 250}&
        \textbf{step 250 + CAG} &
        \textbf{step 1000} &
        \textbf{step 1000 + CAG} \\
        
    \hline
        \textbf{} &
        \vspace{0.1cm}\raisebox{-.025\height}{\includegraphics[width=0.1\textwidth]{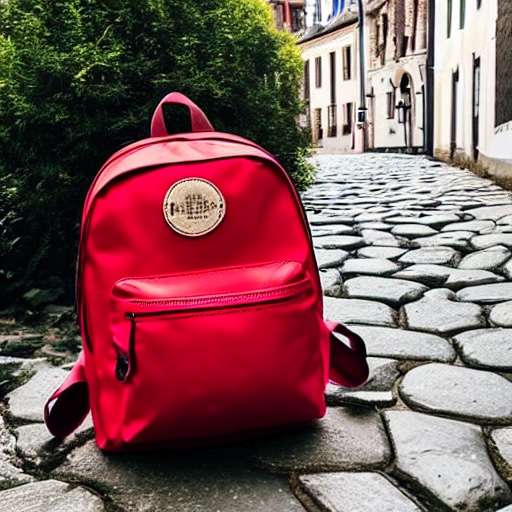}} &
        \raisebox{-.1\height}{\includegraphics[width=0.1\textwidth]{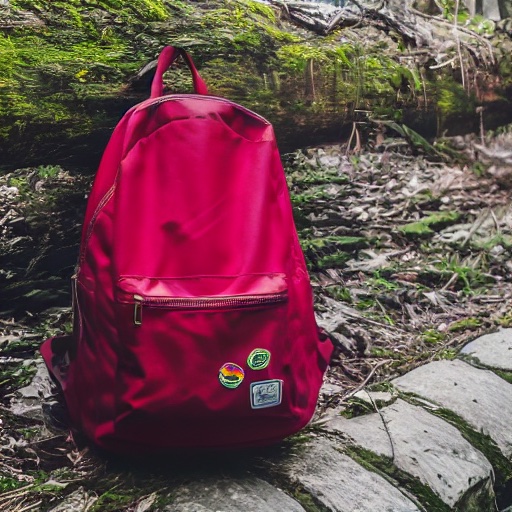}}&
        \raisebox{-.1\height}{\includegraphics[width=0.1\textwidth]{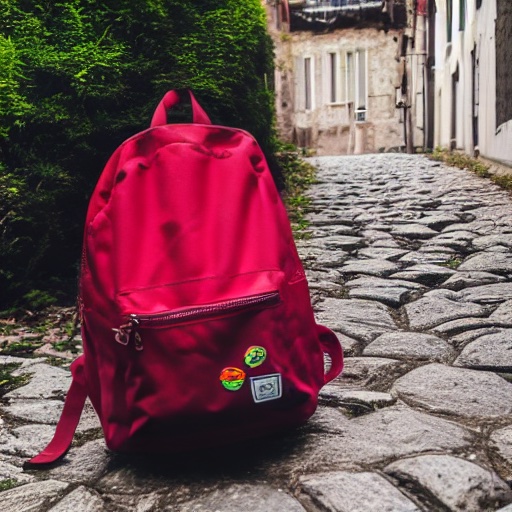}} &
        \raisebox{-.1\height}{\includegraphics[width=0.1\textwidth]{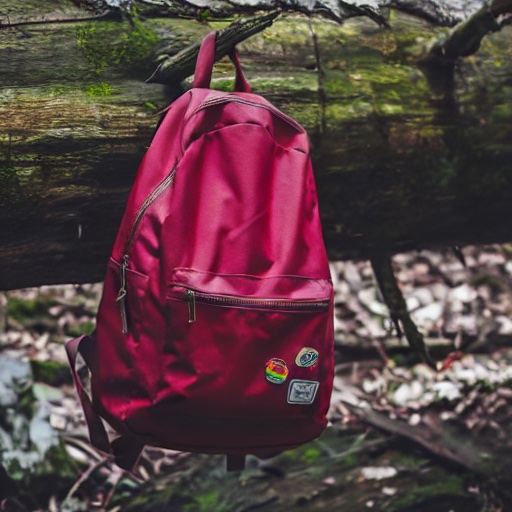}} &
        \raisebox{-.1\height}{\includegraphics[width=0.1\textwidth]{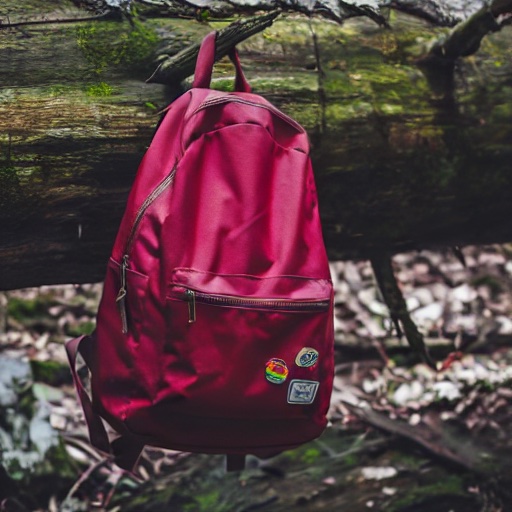}} \\
       
    \hline
        \textbf{sks} &
        \vspace{0.1cm}\raisebox{-.025\height}{\includegraphics[width=0.09\textwidth]{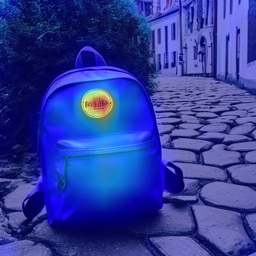}} &
        \raisebox{-.1\height}{\includegraphics[width=0.09\textwidth]{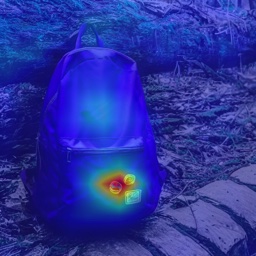}}&
        \raisebox{-.1\height}{\includegraphics[width=0.09\textwidth]{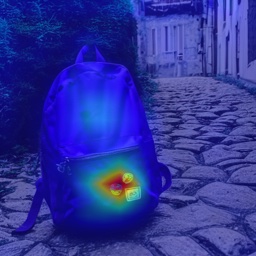}} &
        \raisebox{-.1\height}{\includegraphics[width=0.09\textwidth]{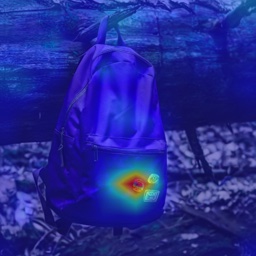}} &
        \raisebox{-.1\height}{\includegraphics[width=0.09\textwidth]{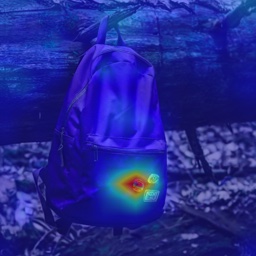}} \\
    \hline  
        \textbf{backpack} &
        \vspace{0.1cm}\raisebox{-.025\height}{\includegraphics[width=0.09\textwidth]{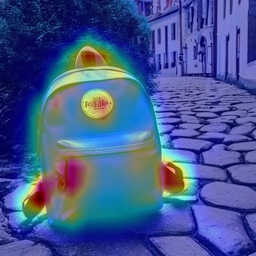}} &
        \raisebox{-.1\height}{\includegraphics[width=0.09\textwidth]{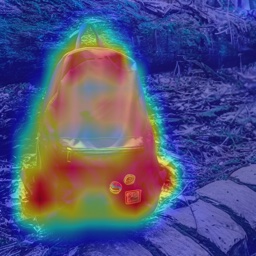}}&
        \raisebox{-.1\height}{\includegraphics[width=0.09\textwidth]{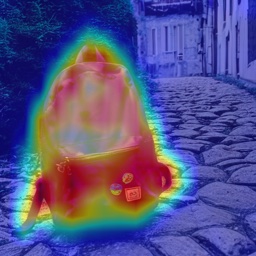}} &
        \raisebox{-.1\height}{\includegraphics[width=0.09\textwidth]{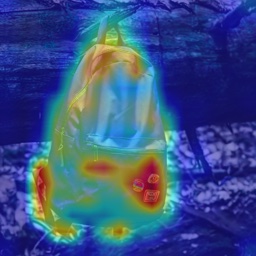}} &
        \raisebox{-.1\height}{\includegraphics[width=0.09\textwidth]{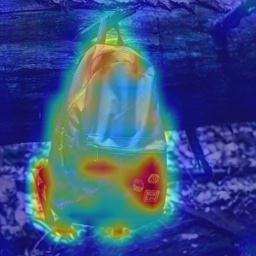}} \\
     \hline  
        \textbf{cobble} &
        \vspace{0.1cm}\raisebox{-.025\height}{\includegraphics[width=0.09\textwidth]{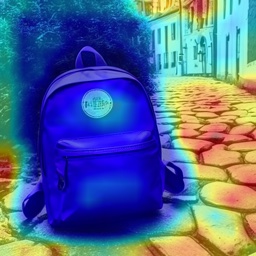}} &
        \raisebox{-.1\height}{\includegraphics[width=0.09\textwidth]{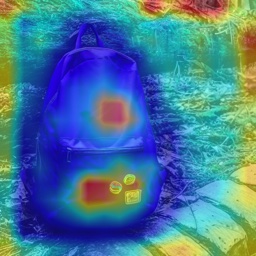}}&
        \raisebox{-.1\height}{\includegraphics[width=0.09\textwidth]{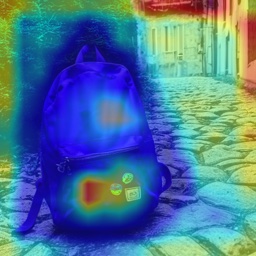}} &
        \raisebox{-.1\height}{\includegraphics[width=0.09\textwidth]{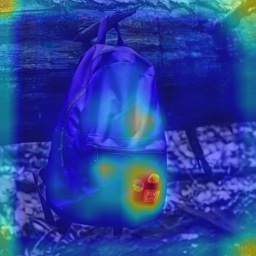}} &
        \raisebox{-.1\height}{\includegraphics[width=0.09\textwidth]{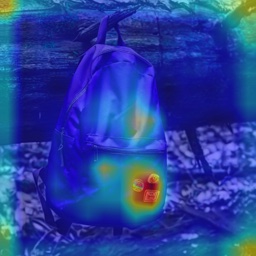}} \\
    \hline  
        \textbf{stone} &
       \vspace{0.1cm}\raisebox{-.025\height}{\includegraphics[width=0.09\textwidth]{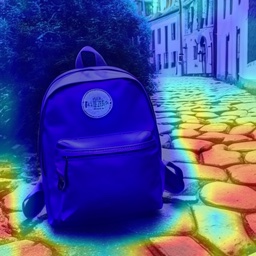}} &
        \raisebox{-.1\height}{\includegraphics[width=0.09\textwidth]{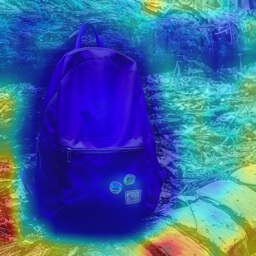}}&
        \raisebox{-.1\height}{\includegraphics[width=0.09\textwidth]{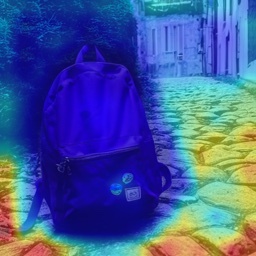}} &
        \raisebox{-.1\height}{\includegraphics[width=0.09\textwidth]{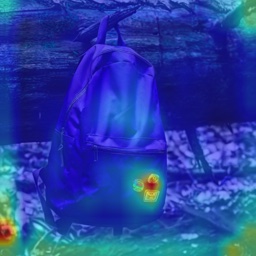}} &
        \raisebox{-.1\height}{\includegraphics[width=0.09\textwidth]{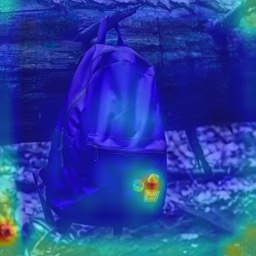}} \\
    \hline  
        \textbf{street} &
        \vspace{0.1cm}\raisebox{-.025\height}{\includegraphics[width=0.09\textwidth]{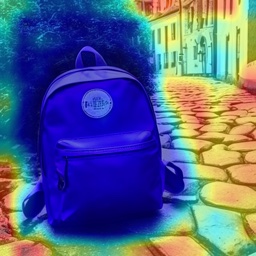}} &
        \raisebox{-.1\height}{\includegraphics[width=0.09\textwidth]{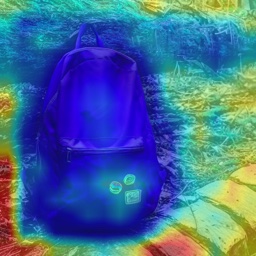}}&
        \raisebox{-.1\height}{\includegraphics[width=0.09\textwidth]{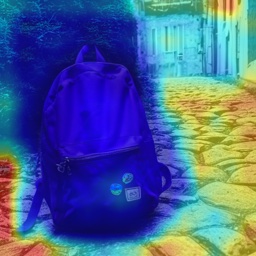}} &
        \raisebox{-.1\height}{\includegraphics[width=0.09\textwidth]{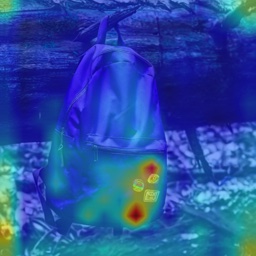}} &
        \raisebox{-.1\height}{\includegraphics[width=0.09\textwidth]{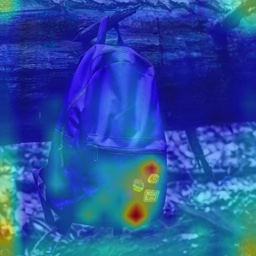}} \\
     \hline  
    \end{tabular}
    \caption{Attention Guidance and Attention Collapse: Images generated at 5, 250, and 1000 steps of DreamBooth fine-tuning, with text-image cross attention maps for $backpack^*$ in \cref{fig:backpacking-overfitting}. The prompt ``a sks backpack on a cobblestone street" features the rare token ``sks" as $backpack^*$. Step 250 + CAG: Cross attention guidance (CAG) from step 5 image is effective. The resulting image maintains layout of step 5 image, while preserving subject fidelity. Step 1000 + CAG: By step 1000, over-fitting leads to catastrophic attention collapse, focusing attention of all tokens mostly on the subject. CAG becomes ineffective as the model maps all latents to one of the input images.}
    \label{fig:cross-attention-maps}
\end{figure*}

In text-image diffusion models, text conditioning is achieved through cross-attention layers spanning various scales of the denoising network, significantly influencing the generated image layout \cite{hertz2022prompt,parmar2023zero,chefer2023attend, zhang2024attention,mou2023dragondiffusion, epstein2023diffusion}. Each word in the input text prompt is tokenized and encoded using a text encoder model like CLIP \cite{radford2021learning}. These text embeddings form the keys $K^{N \times d}$ for cross-attention, where $N$ is the number of tokens and $d$ is the embedding dimension. Queries $Q^{M \times d}$ represent intermediate image features at each cross-attention layer, where $M$ represents image patches treated as tokens, sharing the embedding dimension $d$ with text embeddings. The attention map $A^{M \times N} = {softmax}(QK^T / \sqrt{d})$, tells us how much each text token attends to each image patch. These maps are reshaped to $N \times H \times W$ and visualized in \cref{fig:cross-attention-maps}, where $H \times W = M$ correspond to image spatial dimensions.

 In \cref{fig:cross-attention-maps}, we visualize the text-image cross attention maps during fine-tuning of a StableDiffusion model on the images of $backpack^*$ from \cref{fig:backpacking-overfitting}, represented by the rare text token ``sks'' following DreamBooth \cite{ruiz2023dreambooth, huggingface}. All images stem from same initial latent and prompt ``a sks backpack on a cobblestone street", using same classifier-free guidance and 50 steps of DDIM forward process. At step 5, the model achieves high prompt fidelity and the text-image cross attention for different words in the prompt is focused on relevant parts of the image. By step 250, the model has learnt the subject well and generates images with higher subject fidelity. This is evident as the ``sks" token's attention concentrates on a crucial feature of the backpack: its logo. At the same time, the attentions for other words like ``cobble'', ``stone'' and ``street'' are also beginning to focus on $backpack^*$. By step 1000, over-fitting causes all text tokens to excessively focus on the subject. We call this phenomenon catastrophic attention collapse.

At step 1000, the model becomes highly over-fitted, mapping all latents to one of the input images of $backpack^*$ used for fine-tuning. In contrast, an intermediate model like the one at step 250 retains semantic understanding of concepts like ``cobbles", ``stone" and ``street" from its world knowledge, although their appearances have morphed closer to concepts seen in the input images, as evident from the bush in the step 5 image being replaced by the tree bark of $backpack^*$ input images in the step 250 image. Our key finding is that this step 250 model can be guided to generate an image that follows the layout of the image generated by the early step 5 model, with cross attention guidance. This result is shown in \cref{fig:cross-attention-maps}: observe that step 250 + CAG (Cross Attention Guidance) image follows the layout of the step 5 image. Moreover, the generated backpack is close to our $backpack^*$ as the step 250 model used for image generation has learnt the concept of $backpack^*$ well. This enables us to generate an image with both high prompt fidelity and high subject fidelity.

To achieve this, we implement guided image synthesis where an early checkpoint serves as guidance model $G$ and a later checkpoint as edit model $E$. Starting with a text prompt $P$, we first generate a reference image using $G$, conditioned on text features $c$ derived from $P$. This image prioritizes prompt fidelity but may lack subject fidelity. To utilize the layout of this image as a guide for our final image, we store cross attention maps at each timestep.

Next, we generate the final output image using $E$, starting from the same initial latent used for $G$ and performing cross attention guidance at every step of the diffusion process. At each step, we update the latent in a direction that encourages the current cross attention maps to be close to the reference cross attention maps obtained from $G$. This is achieved by a regularization loss $R$, defined to be the absolute difference of the reference and current cross attention maps and a scalar $\alpha$, that controls the amount of update. These steps are summarized in \cref{alg-guided-image-synthesis}.

\begin{algorithm}
      \caption{Cross Attention Guided Image Synthesis}\label{alg-guided-image-synthesis}
      \textbf{Input:} Text features $c$, Guidance model $G$, Edit model $E$ \\
      \textbf{Output:} Personalized Image $I_{o}$
      \begin{algorithmic}[1]
        \State Step 1: Store reference cross attention maps from $G$
        \State $t \gets 0$
        \State $l \gets l_{i} \in N(0,1)$
        \State $A_{ref} \gets \emptyset$
        \While{$t \leq T$}
        \State $\epsilon,\text{crossAttn} \gets G(l, t, c)$
        \State $l \gets DDIM(l, t, \epsilon)$
        \State $A_{ref}[t] \gets \text{crossAttn}$
        \EndWhile
        \State Step 2: Synthesize final image from $E$, with cross attention guidance
        \State $t \gets 0$
        \State $l \gets l_{i}$
        \While{$t \leq T$}
        \State $ \_,\text{crossAttn} \gets E(l, t, c)$
        \State $R \gets |A_{ref}[t] - \text{crossAttn}|$
        \State $l \gets l - \alpha * \nabla_l(R) $
        \State $\epsilon,\_ \gets E(l, t, c)$
        \State $l \gets DDIM(l, t, \epsilon)$
        \EndWhile
        \State $I_{o} \gets VAE_{decode}(l)$
      \end{algorithmic}
    \end{algorithm}

\section{Experiments and results}
\label{sec:experiments_and_results}
\subsection{Dataset and evaluation metrics}
We use the standard DreamBooth benchmark dataset and evaluation metrics \cite{ruiz2023dreambooth}, used in many works\cite{ruiz2023dreambooth, li2024blip, chen2024subject}. This comprises of 30 subjects from different categories like pets, toys and other objects and 25 prompts for each subject. Subject fidelity and prompt fidelity are important criteria for successful text-to-image personalization. Following prior works \cite{kumari2023multi, ruiz2023dreambooth, li2024blip, chen2024subject, ruiz2024hyperdreambooth}, we use CLIP \cite{radford2021learning} image similarity (CLIP-I) and DINO \cite{caron2021emerging} for subject fidelity and CLIP text similarity (CLIP-T) for prompt fidelity. CLIP-I and DINO are the average pairwise cosine similarities between CLIP and DINO embeddings of input and generated images, respectively. CLIP-T is the average cosine similarity between CLIP embeddings of the text prompt and generated images.

\subsection{Description of methods}
We compare to two methods that fine-tune weights of the text-to-image diffusion model for personalization. DreamBooth \cite{ruiz2023dreambooth} uses a unique rare token to represent a subject, such as ``a sks teddy," and fine-tunes all model weights. Custom Diffusion \cite{kumari2023multi} fine-tunes only text-to-image cross attention weights together with an input word embedding. We apply our approach on models trained with classical DB approach, applying full fine-tune for SDv1.5 and LoRA \cite{hu2021lora} for SDXL. Results for other fine-tuning methods are in \cref{fig:other-backbones}. For comprehensive analysis, we train all models for a large number of 1000 steps, sampling every 5th step up to 50 steps and every 25th step thereafter, resulting in 48 operating points. For our approach, we designate models at steps 100 and 200 as edit models and include all models with lower steps, alongside the pre-trained model, as guidance models, totaling 28 operating points. This dense sampling allows us to study trade offs at different operating points. For each method, we generate 10 images per prompt at each operating point. The best operating point is automatically chosen based on F1 score between CLIP-T and DINO scores.

For completeness, we also compare to non-fine-tuning methods. Textual Inversion \cite{gal2022image} is an inversion-based method that learns an input word embedding to represent a subject. BLIP-Diffusion \cite{li2024blip} and IP-Adapter \cite{ye2023ip} are encoder-based methods offering zero-shot personalized generation. AnyDoor \cite{chen2023anydoor} is a reference-based method that tackles the more challenging problem of placing a specific subject into a specified background image. To evaluate AnyDoor on the DreamBooth benchmark, we used StableDiffusion to generate background images and CLIPSeg \cite{luddecke2022image} for segmentation masks.

\subsection{Qualitative evaluation}
\begin{figure}
\centering

\begin{subfigure}{0.49\textwidth}
    \centering
    \setlength{\tabcolsep}{1pt}
    \begin{tabular}{cccc}
    {Input} & {Overfit} & {Underfit} & {Ours} \\
    \includegraphics[width=0.2\textwidth]{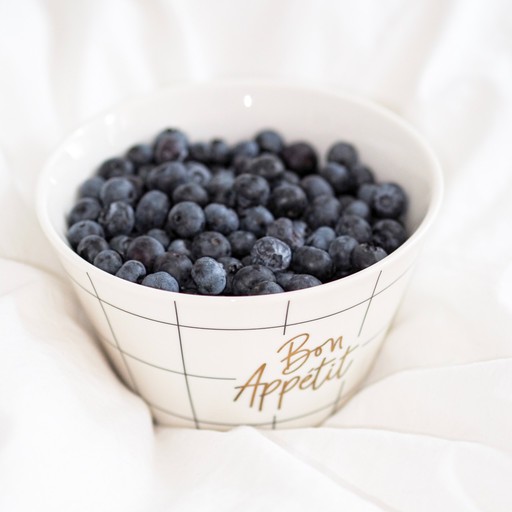} &
    \includegraphics[width=0.2\textwidth]{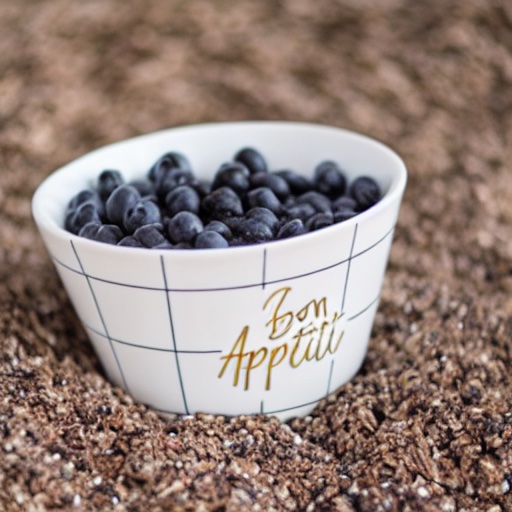} &
    \includegraphics[width=0.2\textwidth]{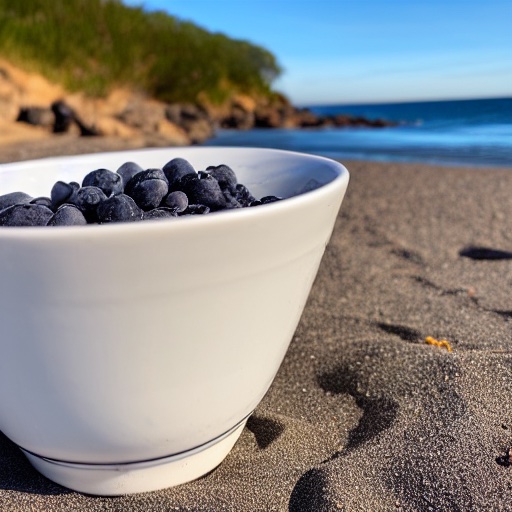} &
    \includegraphics[width=0.2\textwidth]{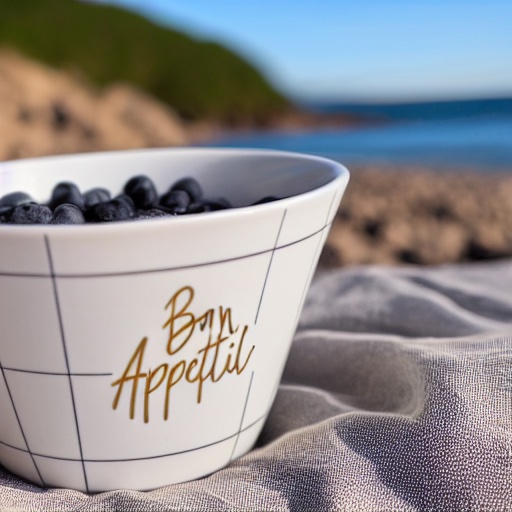} 
    \end{tabular}
    \vspace{-5pt} 
    \caption*{a $bowl^*$ on the beach}
    \label{fig:sub4}
\end{subfigure}

\begin{subfigure}{0.49\textwidth}
    \centering
    \setlength{\tabcolsep}{1pt}
    \begin{tabular}{cccc}
     \includegraphics[width=0.2\textwidth]{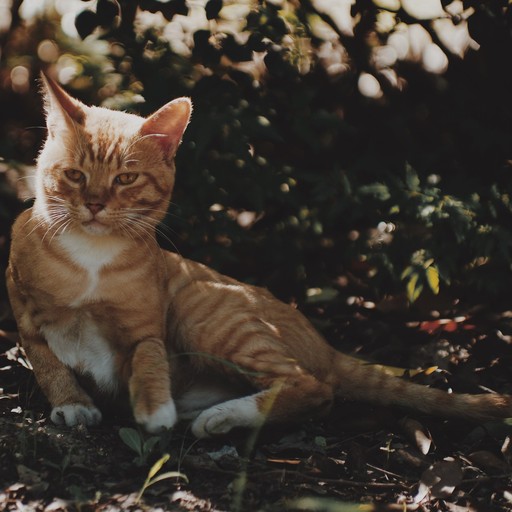} &
    \includegraphics[width=0.2\textwidth]{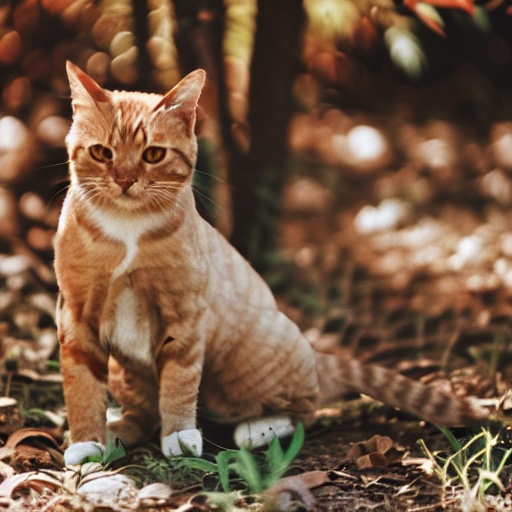} &
    \includegraphics[width=0.2\textwidth]{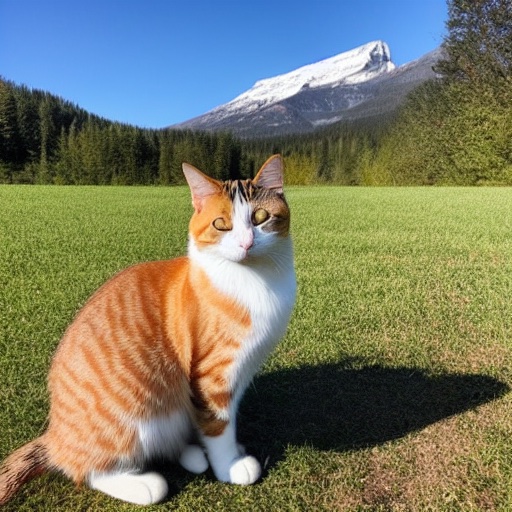} &
    \includegraphics[width=0.2\textwidth]{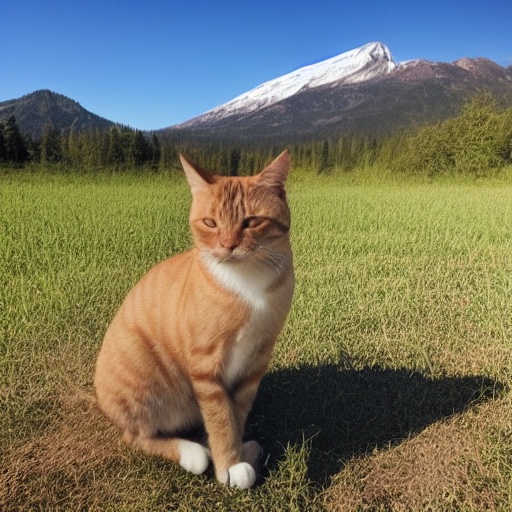} 
    \end{tabular}
    \vspace{-5pt} 
    \caption*{a $cat^*$ with a mountain in the background}
    \label{fig:sub4}
\end{subfigure}

\begin{subfigure}{0.49\textwidth}
    \centering
    \setlength{\tabcolsep}{1pt}
    \begin{tabular}{cccc}
    \includegraphics[width=0.2\textwidth]{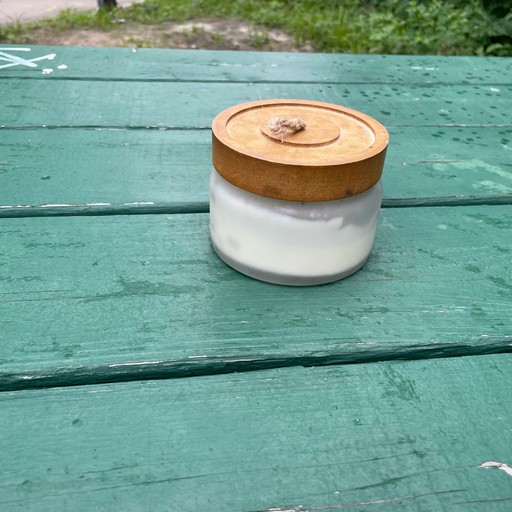} &
    \includegraphics[width=0.2\textwidth]{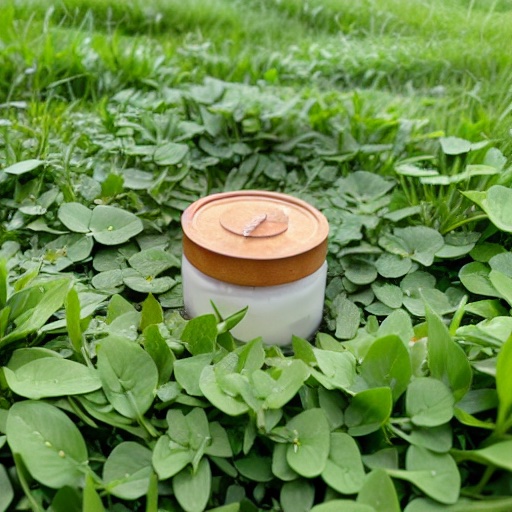} &
    \includegraphics[width=0.2\textwidth]{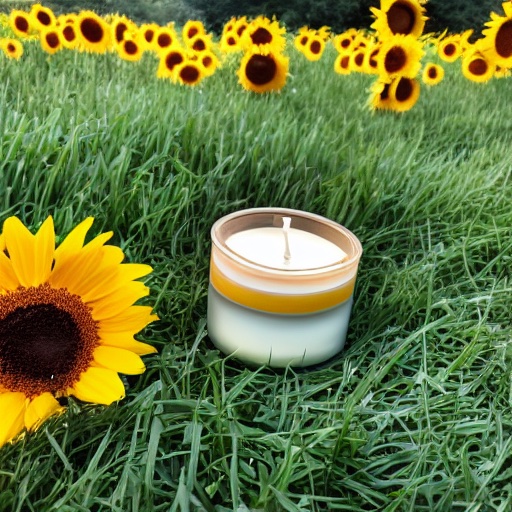} &
    \includegraphics[width=0.2\textwidth]{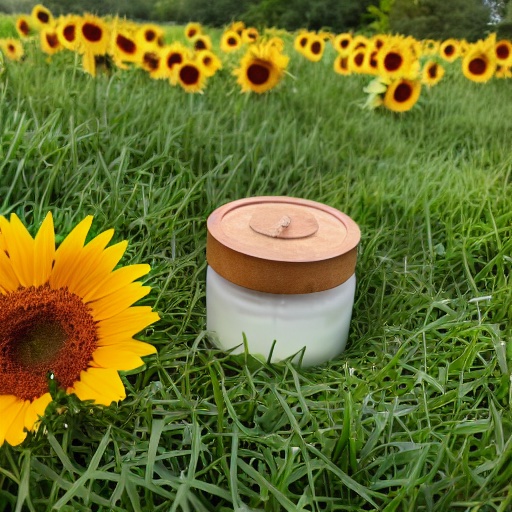} 
    \end{tabular}
    \vspace{-5pt} 
    \caption*{a $candle^*$ on top of green grass with sunflowers around it}
    \label{fig:sub4}
\end{subfigure}

\begin{subfigure}{0.49\textwidth}
    \centering
    \setlength{\tabcolsep}{1pt}
    \begin{tabular}{cccc}
     \includegraphics[width=0.2\textwidth]{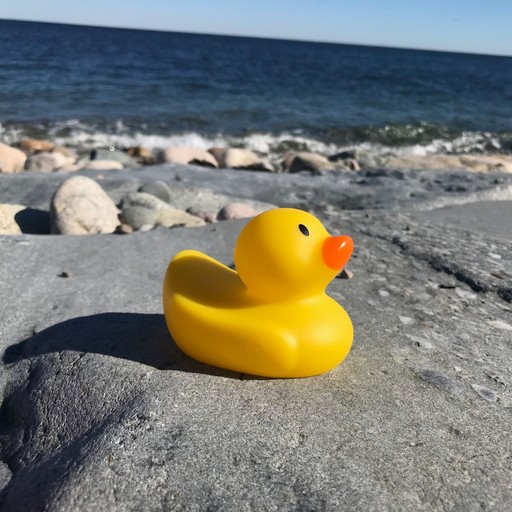} &
    \includegraphics[width=0.2\textwidth]{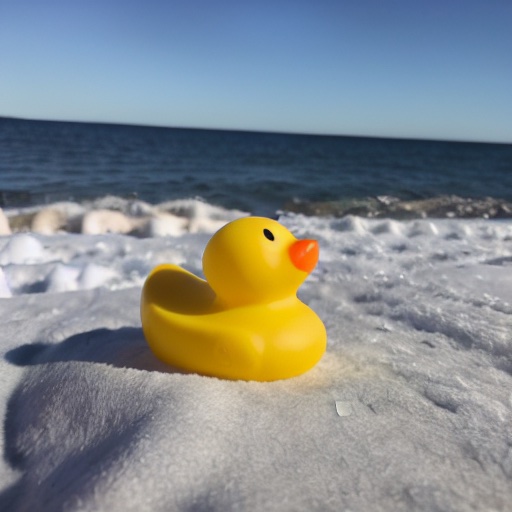} &
    \includegraphics[width=0.2\textwidth]{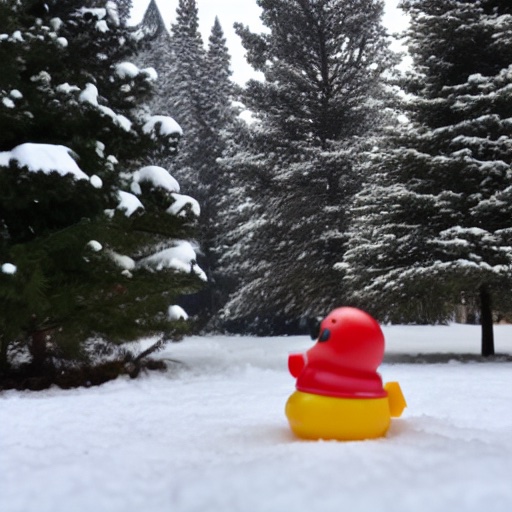} &
    \includegraphics[width=0.2\textwidth]{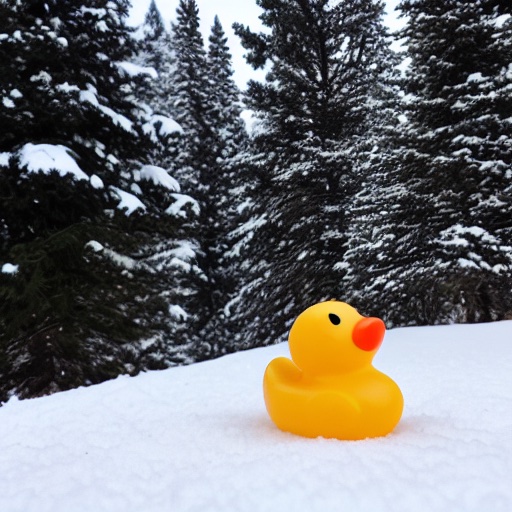} 
    \end{tabular}
    \vspace{-5pt} 
    \caption*{a $toy^*$ in the snow}
    \label{fig:sub4}
\end{subfigure}

\begin{subfigure}{0.49\textwidth}
    \centering
    \setlength{\tabcolsep}{1pt}
    \begin{tabular}{cccc}
    \includegraphics[width=0.2\textwidth]{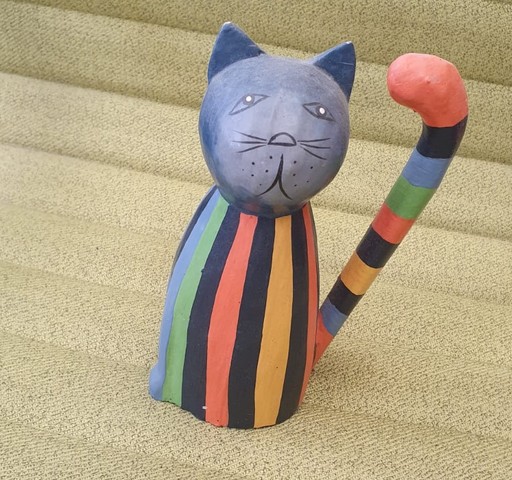} &
    \includegraphics[width=0.2\textwidth]{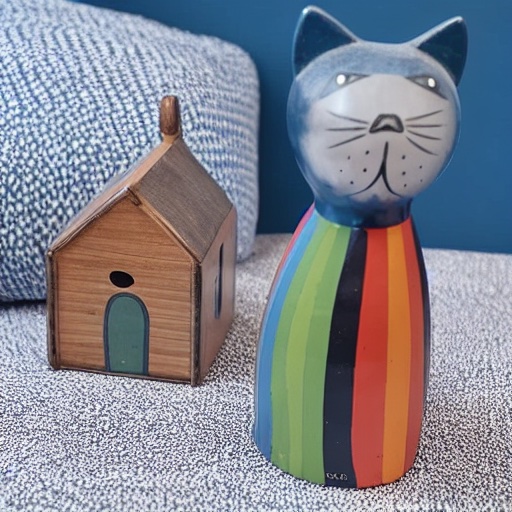} &
    \includegraphics[width=0.2\textwidth]{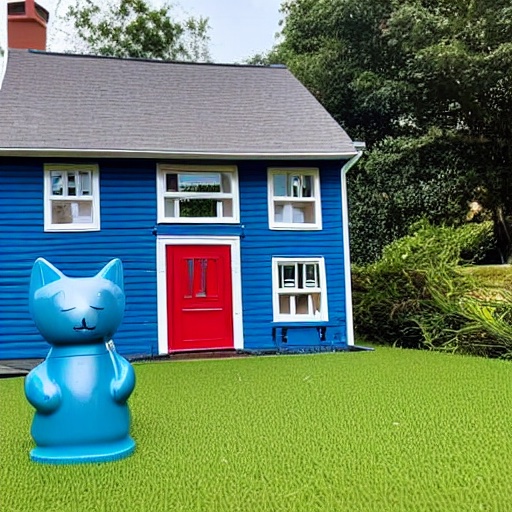} &
    \includegraphics[width=0.2\textwidth]{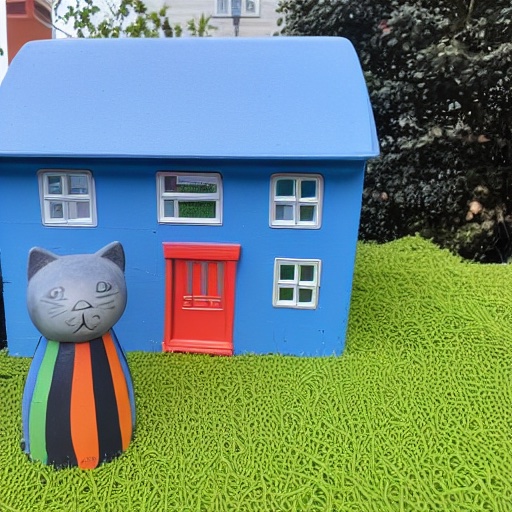} 
    \end{tabular}
    \vspace{-5pt} 
    \caption*{a $toy^*$ with a blue house in the background}
    \label{fig:sub4}
\end{subfigure}

\begin{subfigure}{0.49\textwidth}
    \centering
    \setlength{\tabcolsep}{1pt}
    \begin{tabular}{cccc}
    \includegraphics[width=0.2\textwidth]{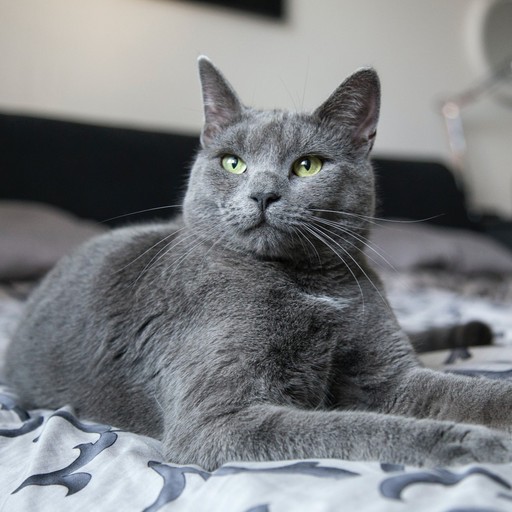} &
    \includegraphics[width=0.2\textwidth]{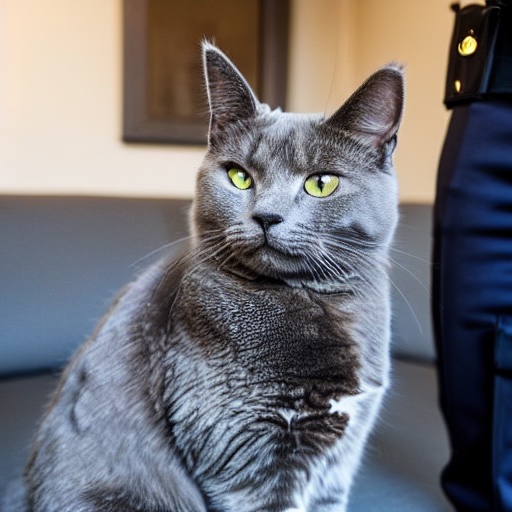} &
    \includegraphics[width=0.2\textwidth]{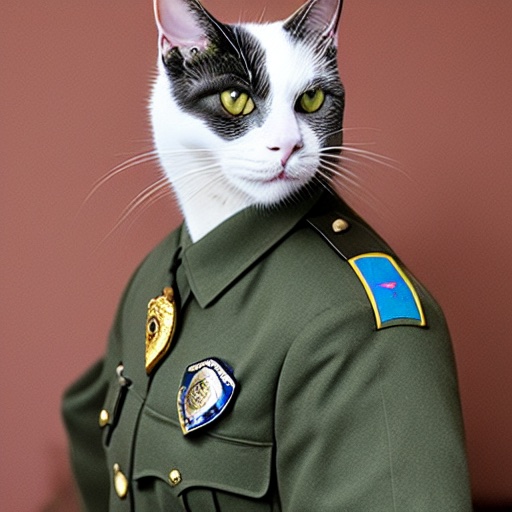} &
    \includegraphics[width=0.2\textwidth]{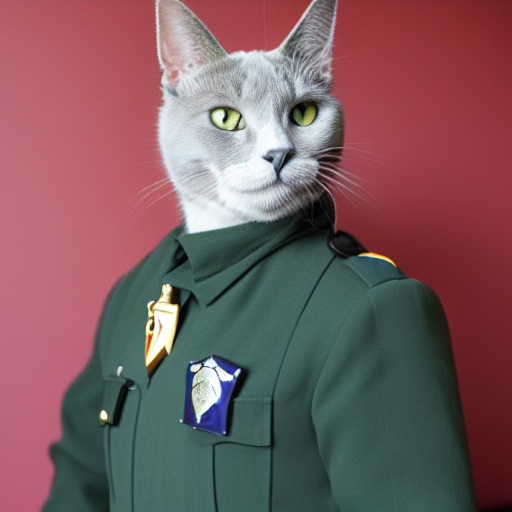} 
    \end{tabular}
    \vspace{-5pt} 
    \caption*{a $cat^*$ in a police outfit}
    \label{fig:sub4}
\end{subfigure}

\begin{subfigure}{0.49\textwidth}
    \centering
    \setlength{\tabcolsep}{1pt}
    \begin{tabular}{cccc}
    \includegraphics[width=0.2\textwidth]{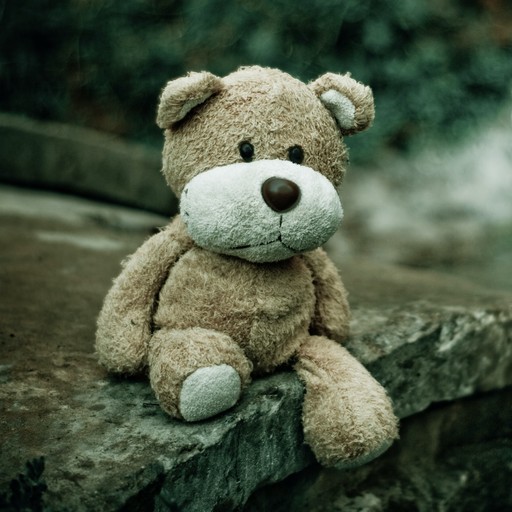} &
    \includegraphics[width=0.2\textwidth]{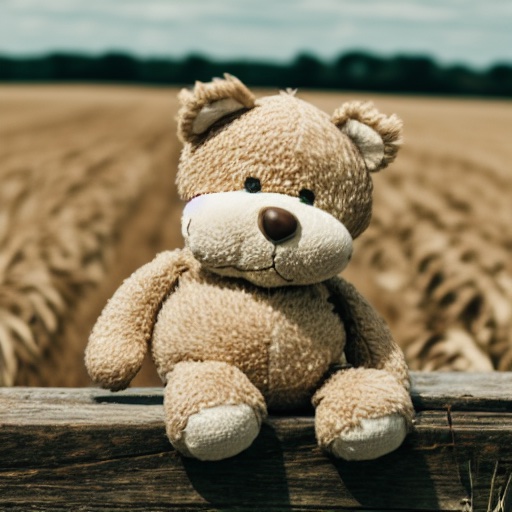} &
    \includegraphics[width=0.2\textwidth]{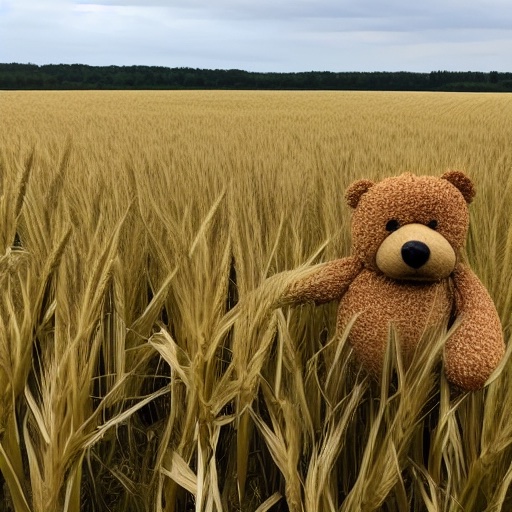} &
    \includegraphics[width=0.2\textwidth]{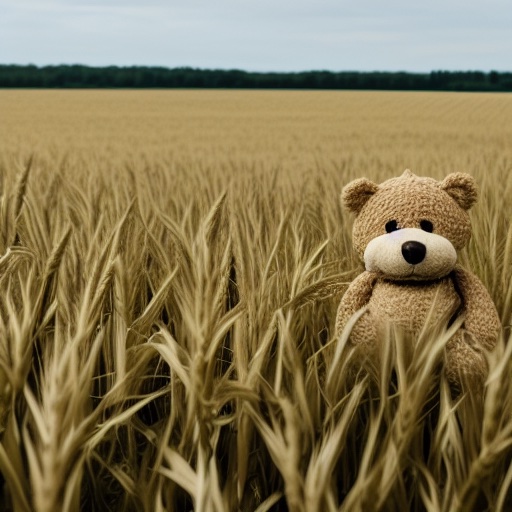} 
    \end{tabular}
    \vspace{-5pt} 
    \caption*{a $teddy^*$ with a wheat field in the background}
    \label{fig:sub4}
\end{subfigure}

\begin{subfigure}{0.49\textwidth}
    \centering
    \setlength{\tabcolsep}{1pt}
    \begin{tabular}{cccc}
     \includegraphics[width=0.2\textwidth]{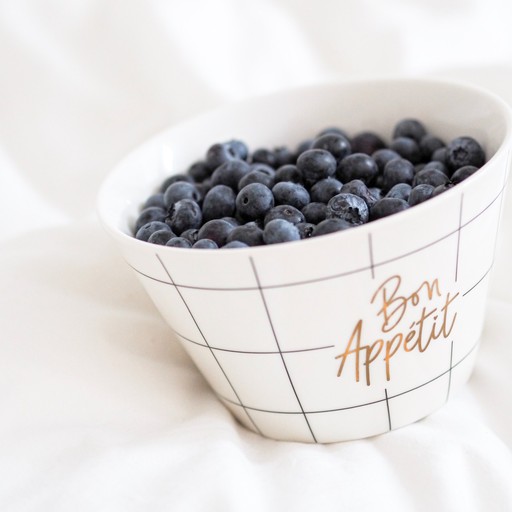} &
    \includegraphics[width=0.2\textwidth]{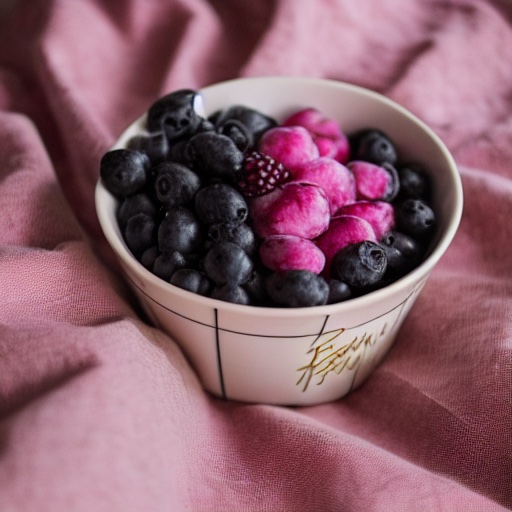} &
    \includegraphics[width=0.2\textwidth]{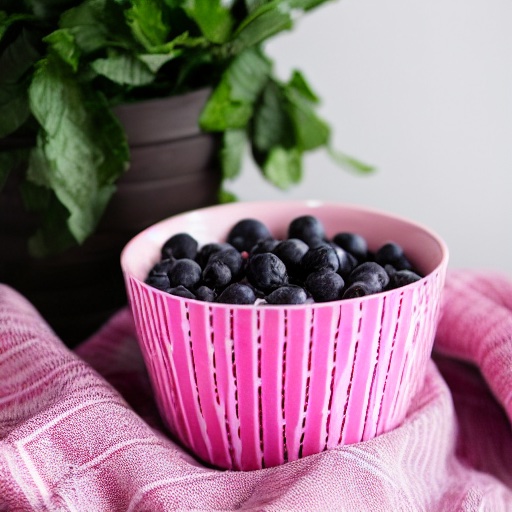} &
    \includegraphics[width=0.2\textwidth]{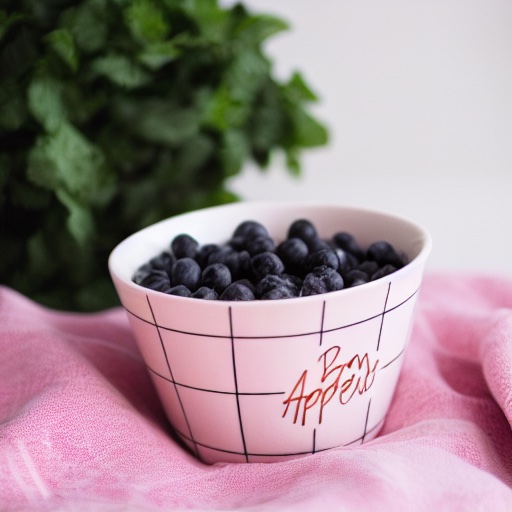} 
    \end{tabular}
    \vspace{-5pt} 
    \caption*{a $bowl^*$ on top of pink fabric}
    \vspace{10pt} 
    \label{fig:sub4}
\end{subfigure}

\vspace{-5pt} 

\caption{Cross Attention Guided Image Synthesis: Across various subjects and prompts, our approach successfully preserves the layout of the reference underfit image as well as the identity of the input subject. Images generated by the Overfit (Edit) and Underfit (Guidance) models used, are shown for reference.}
\label{fig:guided-image-synthesis}

\vspace{-15pt}
\end{figure}

\begin{figure*}
\centering

\begin{subfigure}{0.24\textwidth}
    \centering
    \setlength{\tabcolsep}{2pt}
     \caption*{Input}
    \begin{tabular}{cc}
    \includegraphics[width=0.6\textwidth]{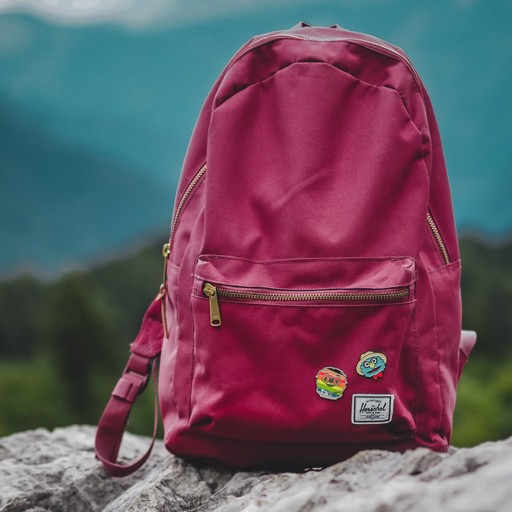} &
    \raisebox{0.32\textwidth} {\begin{tabular}[t]{@{}c@{}} 
    \includegraphics[width=0.28\textwidth]{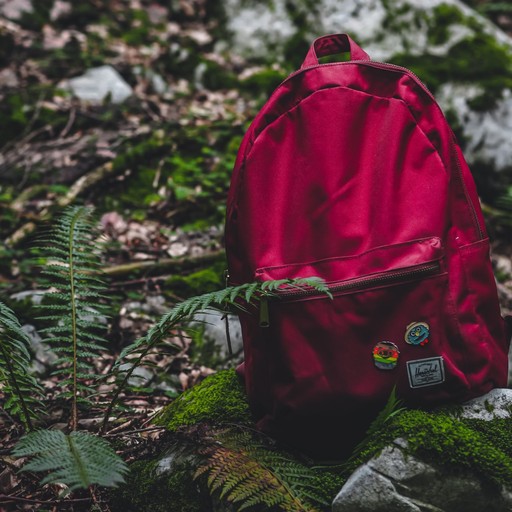} \\
    \includegraphics[width=0.28\textwidth]{images/backpack_input/05.jpg}
    \end{tabular}}
    \end{tabular}
    \label{fig:sub4}
\end{subfigure}
\begin{subfigure}{0.24\textwidth}
    \centering
    \setlength{\tabcolsep}{2pt}
    \caption*{Ours}
    \begin{tabular}{cc}
    \includegraphics[width=0.6\textwidth]{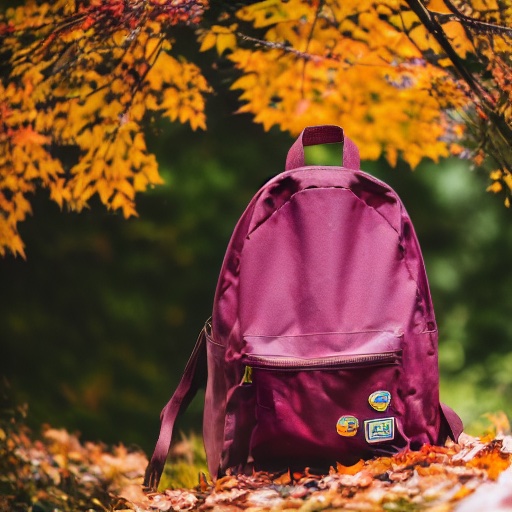} &
    \raisebox{0.32\textwidth} {\begin{tabular}[t]{@{}c@{}} 
    \includegraphics[width=0.28\textwidth]{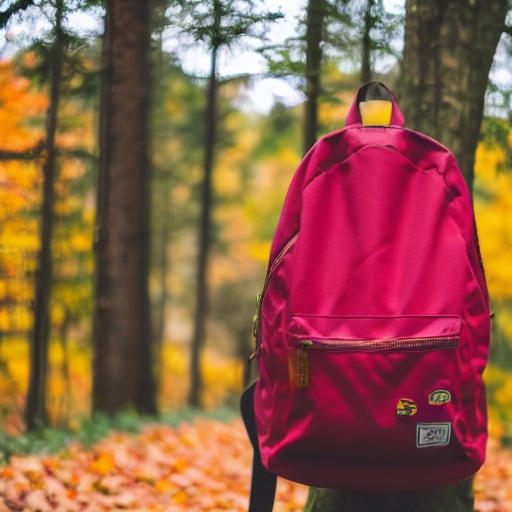} \\
    \includegraphics[width=0.28\textwidth]{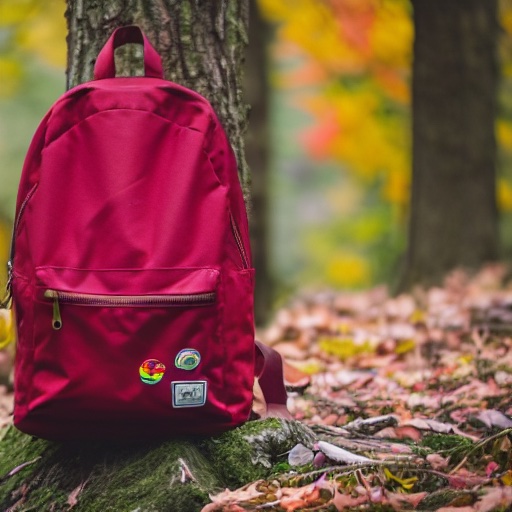}
    \end{tabular}}
    \end{tabular}
    \label{fig:sub4}
\end{subfigure}
\begin{subfigure}{0.24\textwidth}
    \centering
    \setlength{\tabcolsep}{2pt}
    \caption*{DreamBooth}
    \begin{tabular}{cc}
    \includegraphics[width=0.6\textwidth]{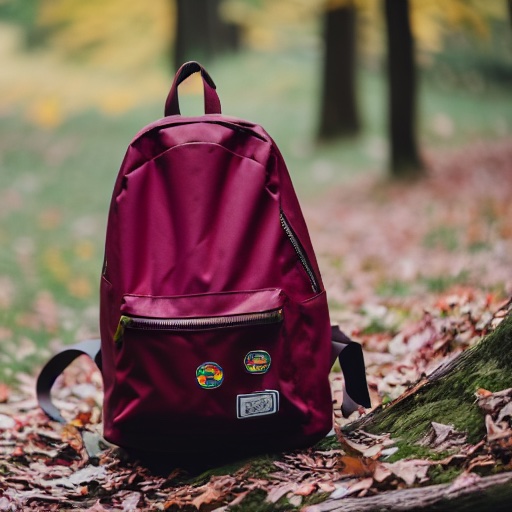} &
    \raisebox{0.32\textwidth} {\begin{tabular}[t]{@{}c@{}} 
    \includegraphics[width=0.28\textwidth]{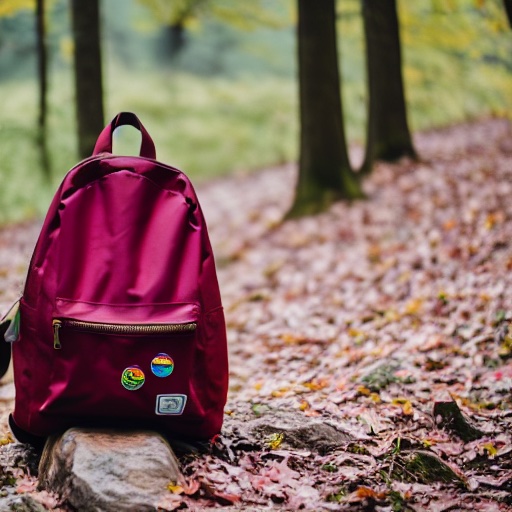} \\
    \includegraphics[width=0.28\textwidth]{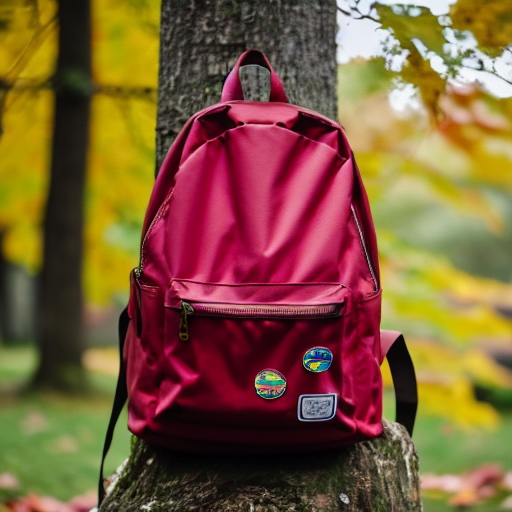}
    \end{tabular}}
    \end{tabular}
    \label{fig:sub4}
\end{subfigure}
\begin{subfigure}{0.24\textwidth}
    \centering
    \setlength{\tabcolsep}{2pt}
    \caption*{CustomDiffusion}
    \begin{tabular}{cc}
    \includegraphics[width=0.6\textwidth]{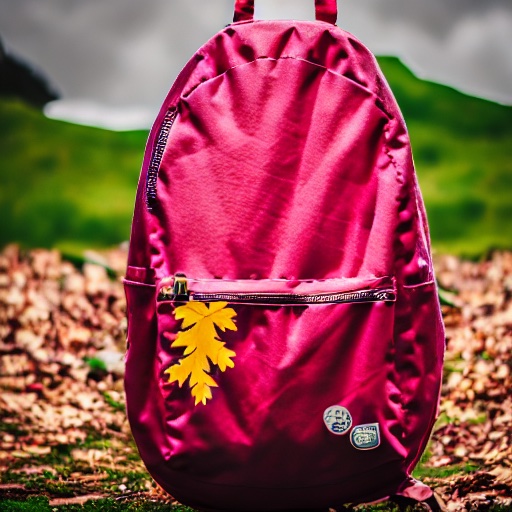} &
    \raisebox{0.32\textwidth} {\begin{tabular}[t]{@{}c@{}} 
    \includegraphics[width=0.28\textwidth]{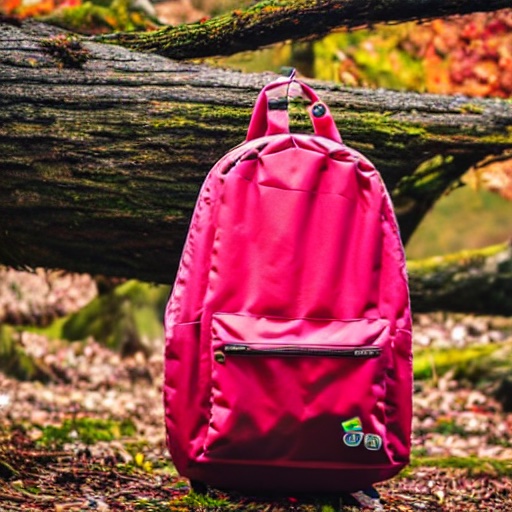} \\
    \includegraphics[width=0.28\textwidth]{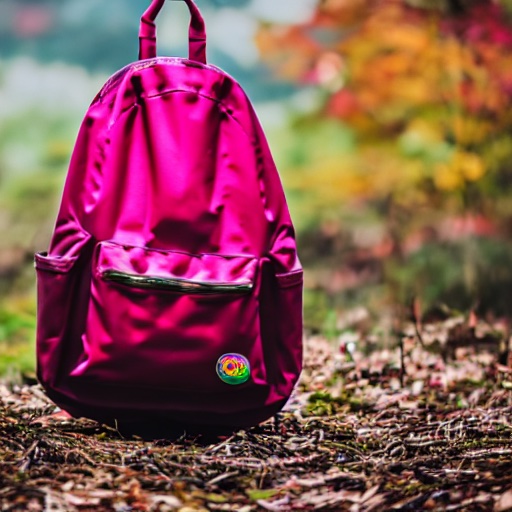}
    \end{tabular}}
    \end{tabular}
    \label{fig:sub4}
\end{subfigure}
\vspace{-12pt} 
\caption*{a $backpack^*$ with a tree and autumn leaves in the background}

\begin{subfigure}{0.24\textwidth}
    \centering
    \setlength{\tabcolsep}{2pt}
    \begin{tabular}{cc}
    \includegraphics[width=0.6\textwidth]{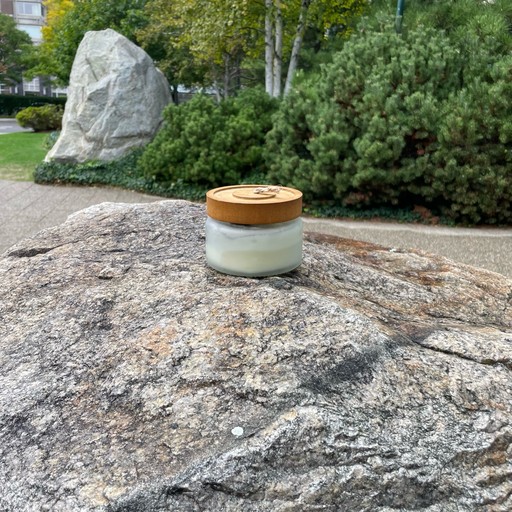} &
    \raisebox{0.32\textwidth} {\begin{tabular}[t]{@{}c@{}} 
    \includegraphics[width=0.28\textwidth]{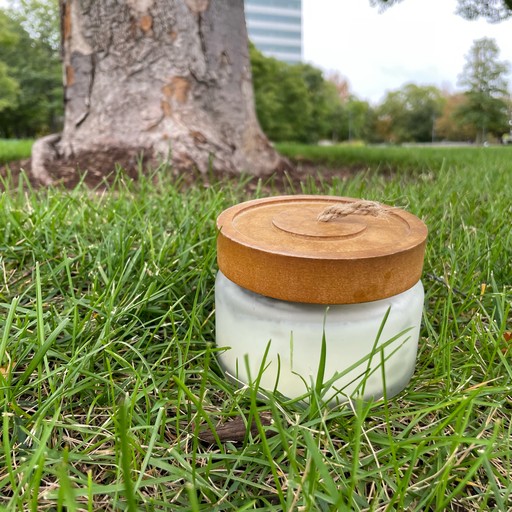} \\
    \includegraphics[width=0.28\textwidth]{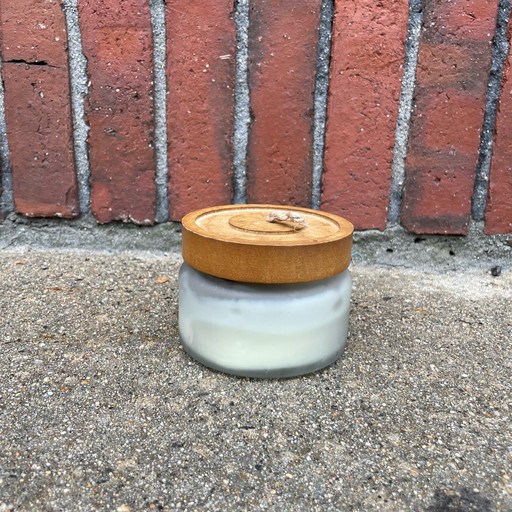}
    \end{tabular}}
    \end{tabular}
    \label{fig:sub4}
\end{subfigure}
\begin{subfigure}{0.24\textwidth}
    \centering
    \setlength{\tabcolsep}{2pt}
    \begin{tabular}{cc}
    \includegraphics[width=0.6\textwidth]{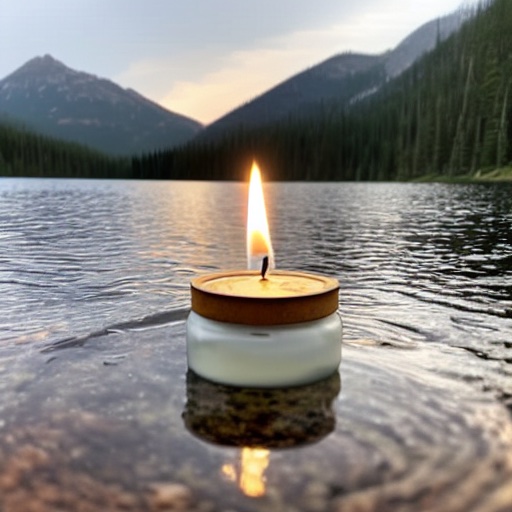} &
    \raisebox{0.32\textwidth} {\begin{tabular}[t]{@{}c@{}} 
    \includegraphics[width=0.28\textwidth]{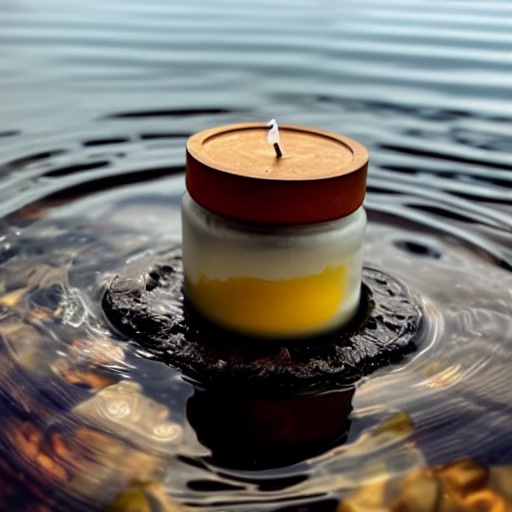} \\
    \includegraphics[width=0.28\textwidth]{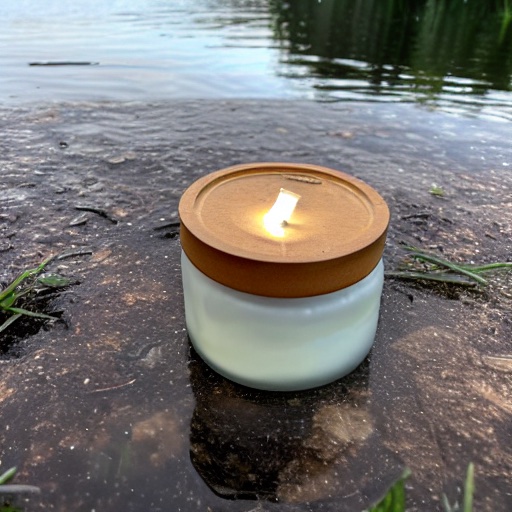}
    \end{tabular}}
    \end{tabular}
    \label{fig:sub4}
\end{subfigure}
\begin{subfigure}{0.24\textwidth}
    \centering
    \setlength{\tabcolsep}{2pt}
    \begin{tabular}{cc}
    \includegraphics[width=0.6\textwidth]{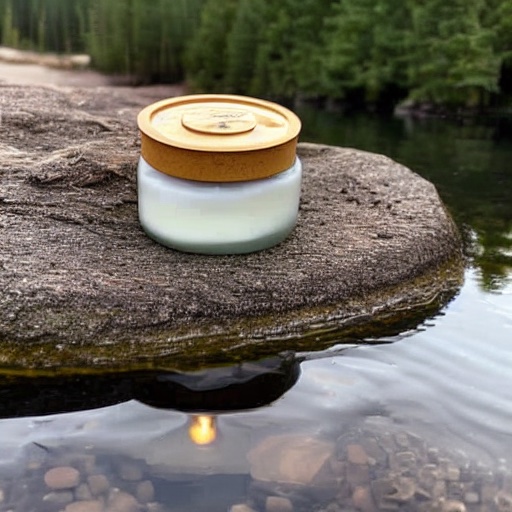} &
    \raisebox{0.32\textwidth} {\begin{tabular}[t]{@{}c@{}} 
    \includegraphics[width=0.28\textwidth]{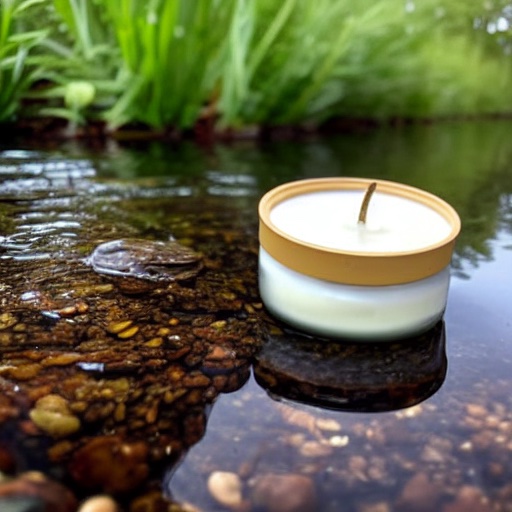} \\
    \includegraphics[width=0.28\textwidth]{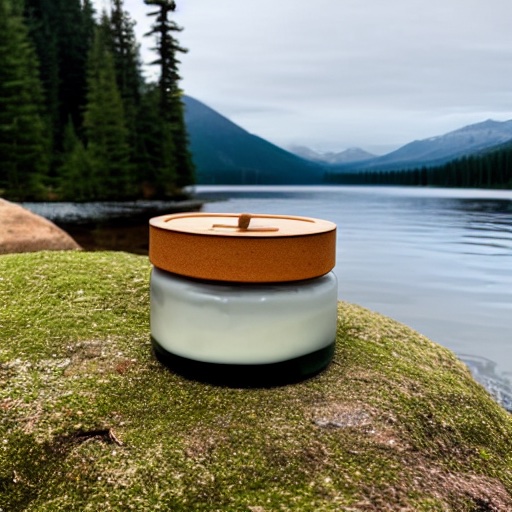}
    \end{tabular}}
    \end{tabular}
    \label{fig:sub4}
\end{subfigure}
\begin{subfigure}{0.24\textwidth}
    \centering
    \setlength{\tabcolsep}{2pt}
    \begin{tabular}{cc}
    \includegraphics[width=0.6\textwidth]{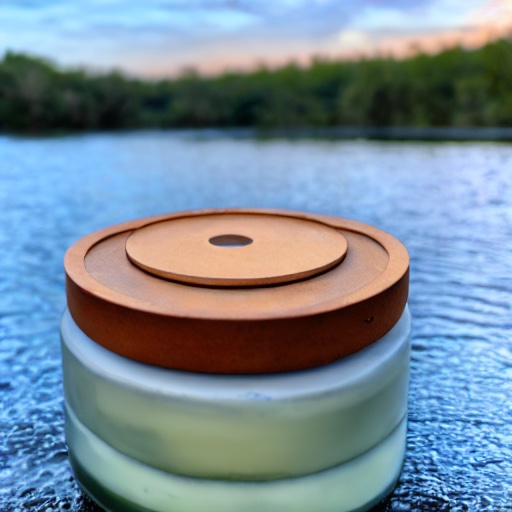} &
    \raisebox{0.32\textwidth} {\begin{tabular}[t]{@{}c@{}} 
    \includegraphics[width=0.28\textwidth]{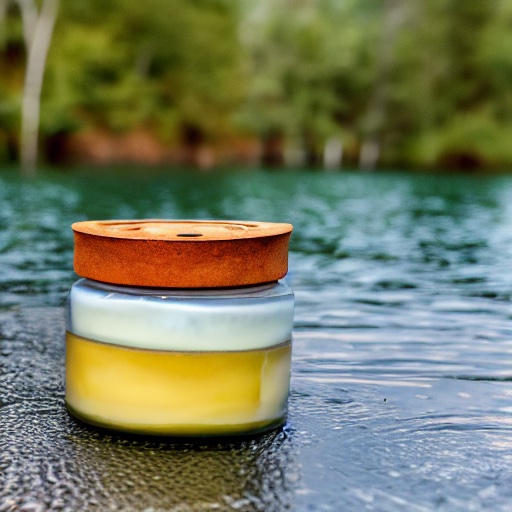} \\
    \includegraphics[width=0.28\textwidth]{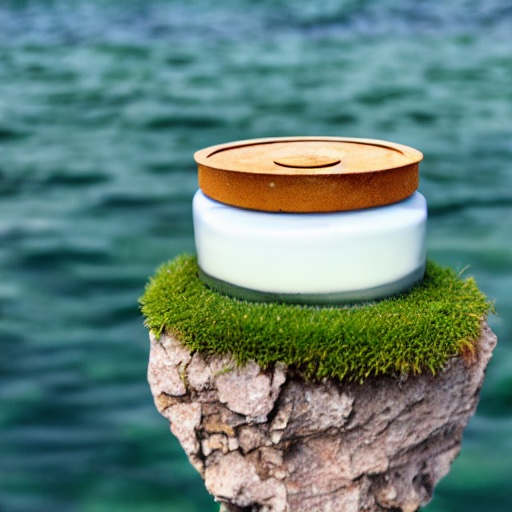}
    \end{tabular}}
    \end{tabular}
    \label{fig:sub4}
\end{subfigure}
\vspace{-12pt} 
\caption*{a $candle^*$ floating on top of water}

\begin{subfigure}{0.24\textwidth}
    \centering
    \setlength{\tabcolsep}{2pt}
    \begin{tabular}{cc}
    \includegraphics[width=0.6\textwidth]{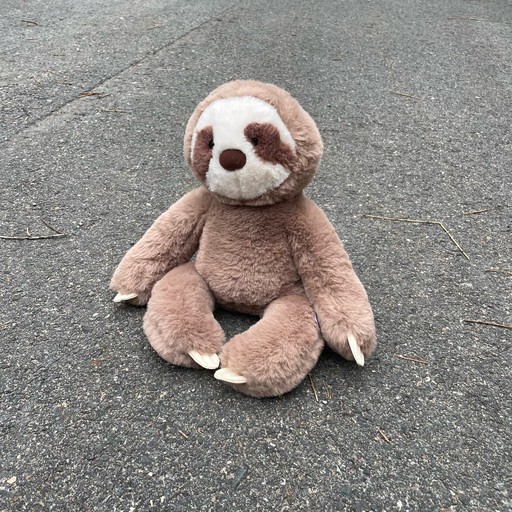} &
    \raisebox{0.32\textwidth} {\begin{tabular}[t]{@{}c@{}} 
    \includegraphics[width=0.28\textwidth]{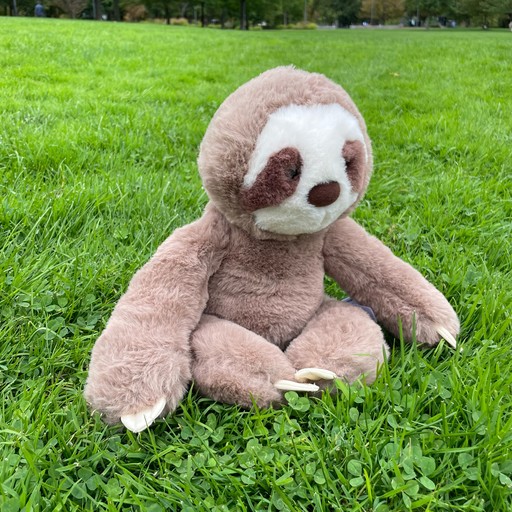} \\
    \includegraphics[width=0.28\textwidth]{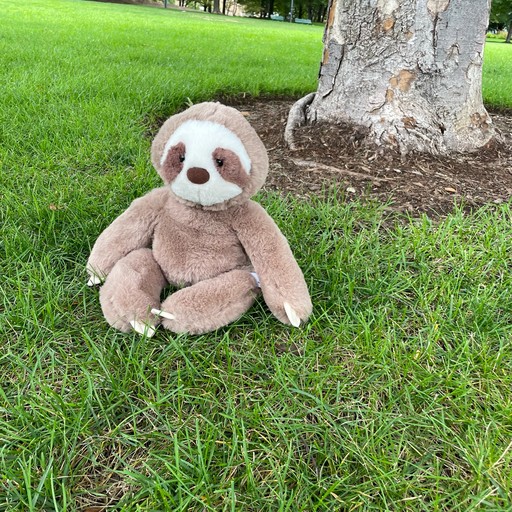}
    \end{tabular}}
    \end{tabular}
    \label{fig:sub4}
\end{subfigure}
\begin{subfigure}{0.24\textwidth}
    \centering
    \setlength{\tabcolsep}{2pt}
    \begin{tabular}{cc}
    \includegraphics[width=0.6\textwidth]{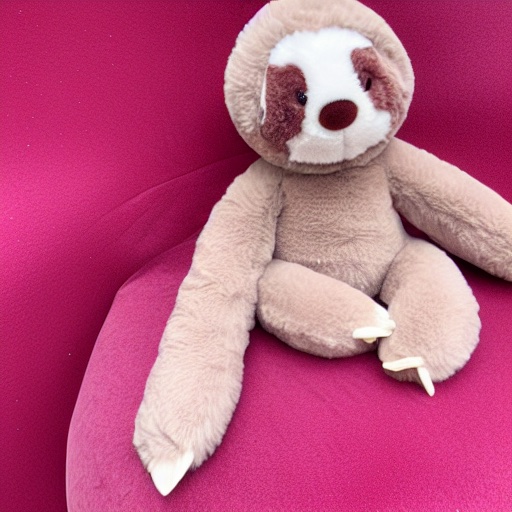} &
    \raisebox{0.32\textwidth} {\begin{tabular}[t]{@{}c@{}} 
    \includegraphics[width=0.28\textwidth]{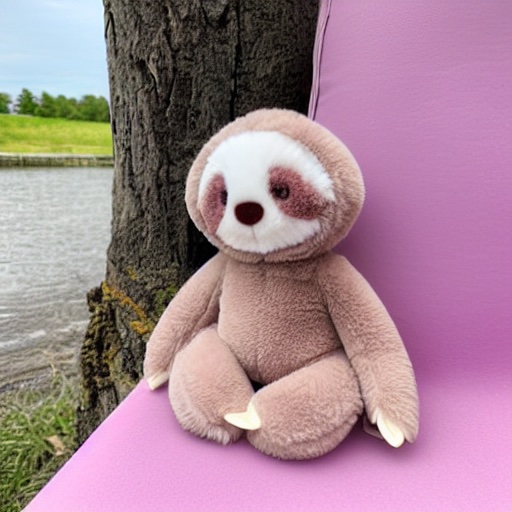} \\
    \includegraphics[width=0.28\textwidth]{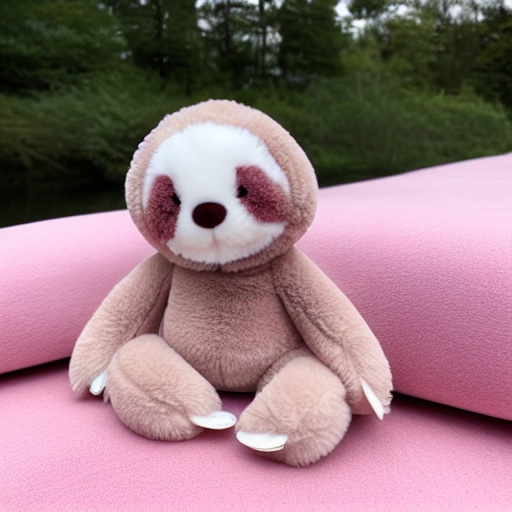}
    \end{tabular}}
    \end{tabular}
    \label{fig:sub4}
\end{subfigure}
\begin{subfigure}{0.24\textwidth}
    \centering
    \setlength{\tabcolsep}{2pt}
    \begin{tabular}{cc}
    \includegraphics[width=0.6\textwidth]{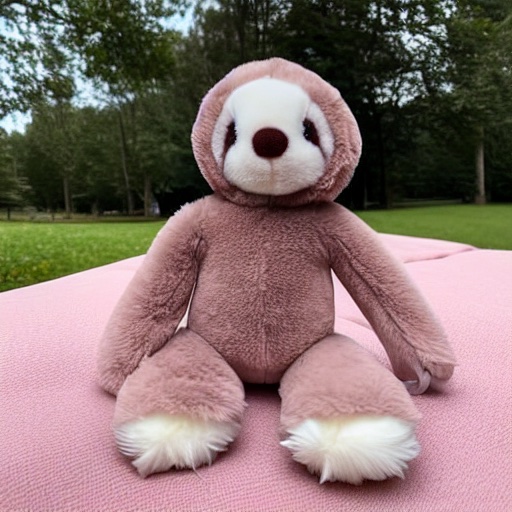} &
    \raisebox{0.32\textwidth} {\begin{tabular}[t]{@{}c@{}} 
    \includegraphics[width=0.28\textwidth]{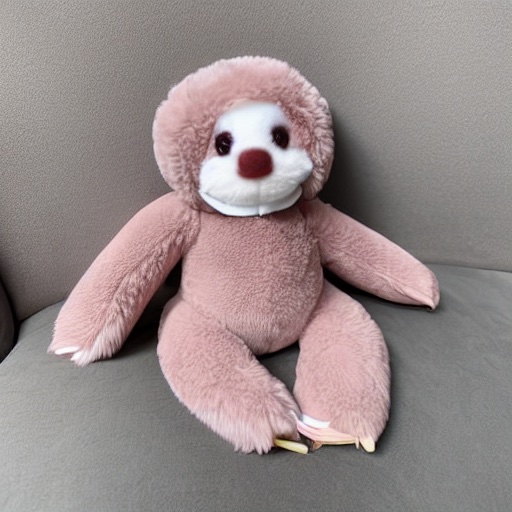} \\
    \includegraphics[width=0.28\textwidth]{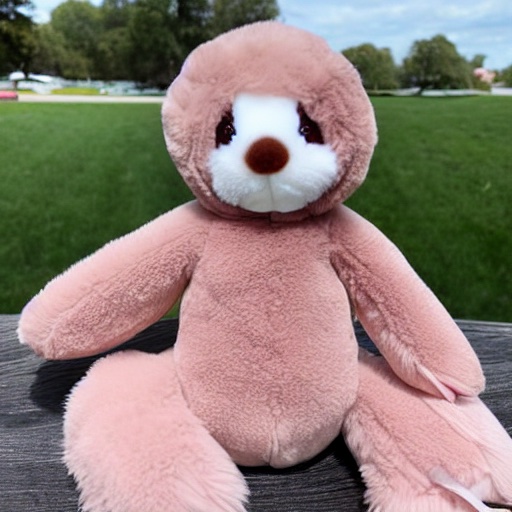}
    \end{tabular}}
    \end{tabular}
    \label{fig:sub4}
\end{subfigure}
\begin{subfigure}{0.24\textwidth}
    \centering
    \setlength{\tabcolsep}{2pt}
    \begin{tabular}{cc}
    \includegraphics[width=0.6\textwidth]{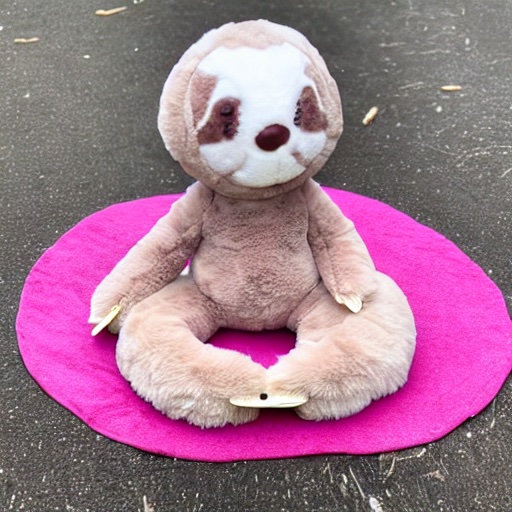} &
    \raisebox{0.32\textwidth} {\begin{tabular}[t]{@{}c@{}} 
    \includegraphics[width=0.28\textwidth]{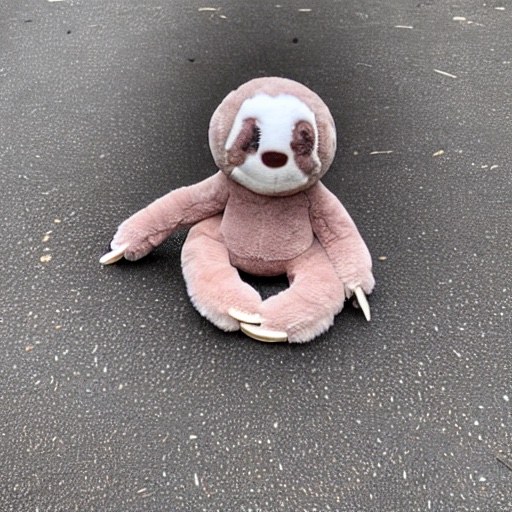} \\
    \includegraphics[width=0.28\textwidth]{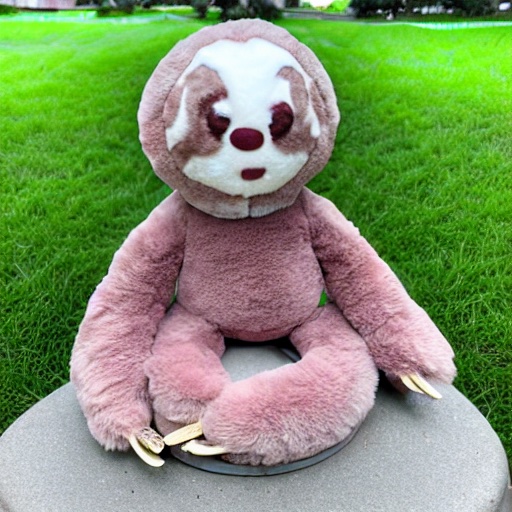}
    \end{tabular}}
    \end{tabular}
    \label{fig:sub4}
\end{subfigure}
\vspace{-12pt} 
\caption*{a \textit{stuffed animal$^*$} on top of pink fabric}

\caption{Comparison with fine-tuning methods: Our approach successfully generates images with better subject fidelity, prompt fidelity and diversity on challenging prompts. In our results, autumn leaves in $backpack^*$ images are more visible, $candle^*$ is floating on water in all three images while maintaining subject fidelity and \textit{stuffed animal$^*$} is on a pink fabric in all three images, exhibiting different poses.}
\label{fig:qualitative-comparison}
\vspace{-5pt}
\end{figure*}

\begin{figure}
    \centering
        \begin{subfigure}{1.0\linewidth}
            \centering
            \setlength{\tabcolsep}{1pt}
            \begin{tabular}{cccccc}
            Input &  TI & BLIP-D & IP-A & AnyDoor & Ours \\
            \includegraphics[width=0.16\linewidth]{images/backpack_input/05.jpg} &
            \includegraphics[width=0.16\linewidth]{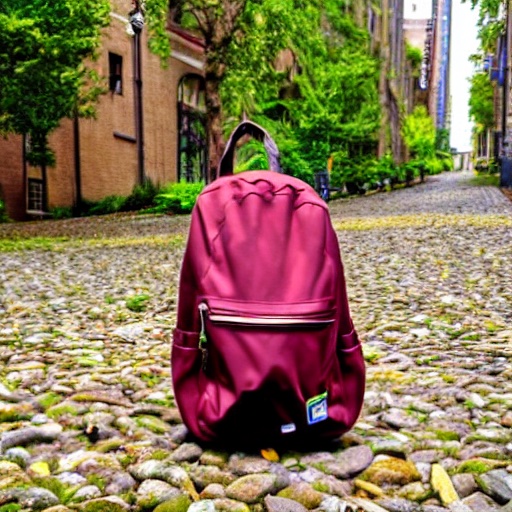} &
            \includegraphics[width=0.16\linewidth]{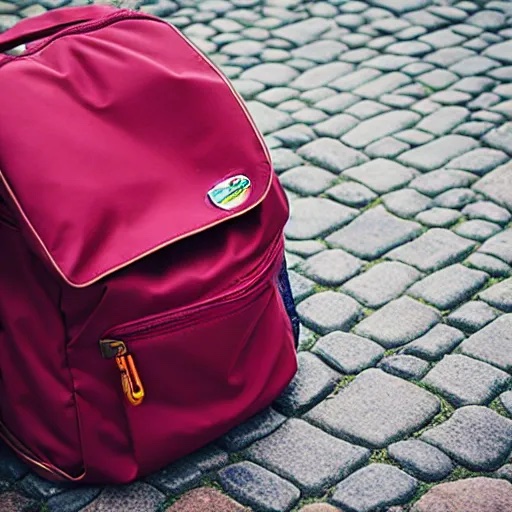} &
            \includegraphics[width=0.16\linewidth]{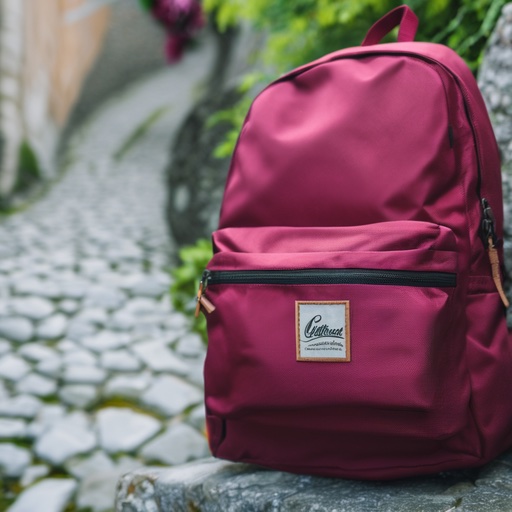} &
            \includegraphics[width=0.16\linewidth]{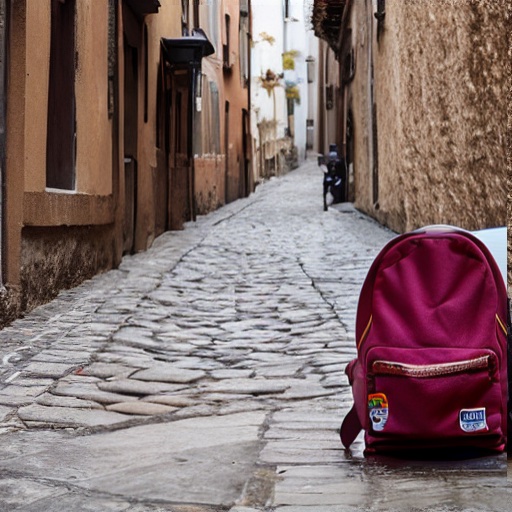} &
            \includegraphics[width=0.16\linewidth]{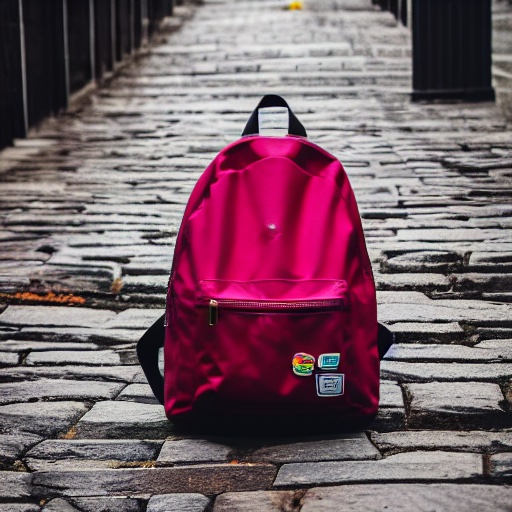} 
            \end{tabular}
            \vspace{-5pt} 
            \caption*{a $backpack^*$ on a cobblestone street}
        \end{subfigure}

        \begin{subfigure}{1.0\linewidth}
            \centering
            \setlength{\tabcolsep}{1pt}
            \begin{tabular}{cccccc}
            \includegraphics[width=0.16\textwidth]{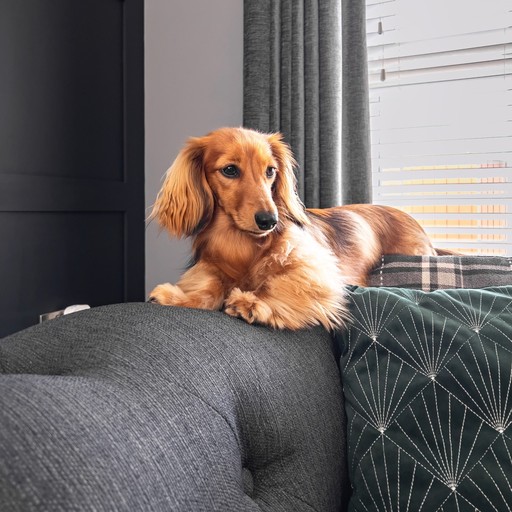} &
            \includegraphics[width=0.16\textwidth]{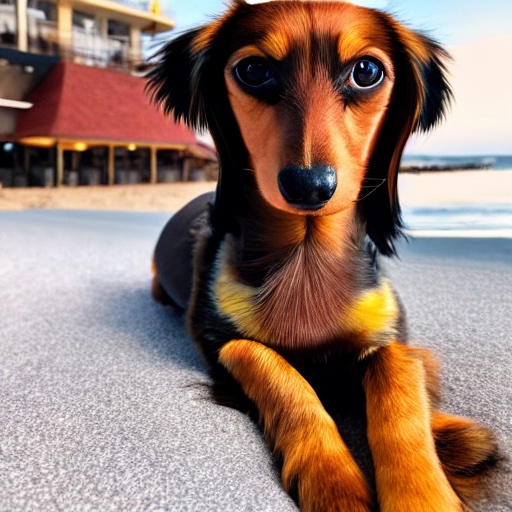} &
            \includegraphics[width=0.16\textwidth]{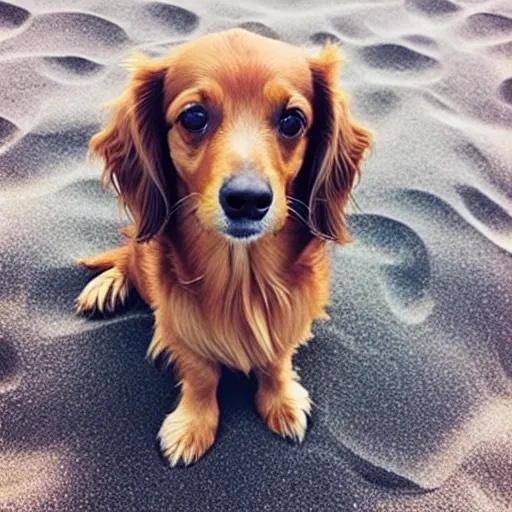} &
            \includegraphics[width=0.16\textwidth]{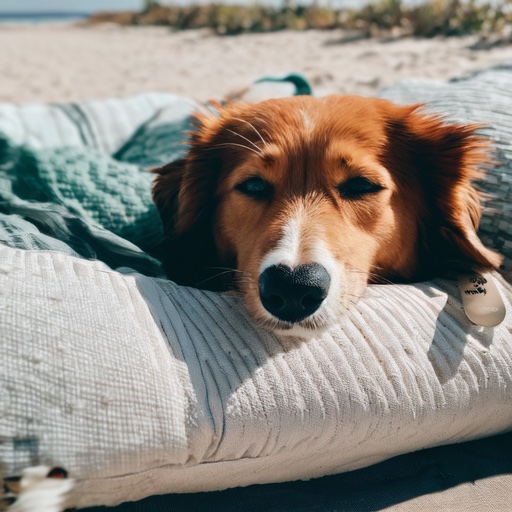} &
            \includegraphics[width=0.16\textwidth]{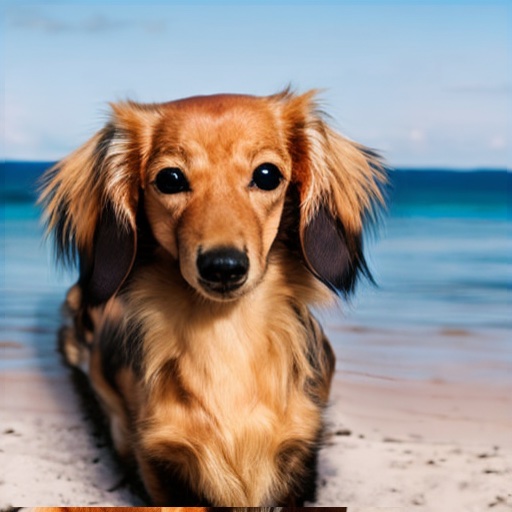} &
            \includegraphics[width=0.16\textwidth]{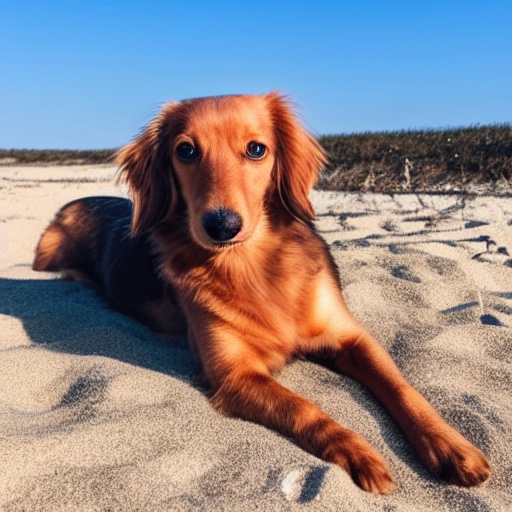} \\
            \end{tabular}
            \vspace{-5pt} 
            \caption*{a $dog^*$ on a beach}
        \end{subfigure}
        
        \begin{subfigure}{1.0\linewidth}
            \centering
            \setlength{\tabcolsep}{1pt}
            \begin{tabular}{cccccc}
            \includegraphics[width=0.16\textwidth]{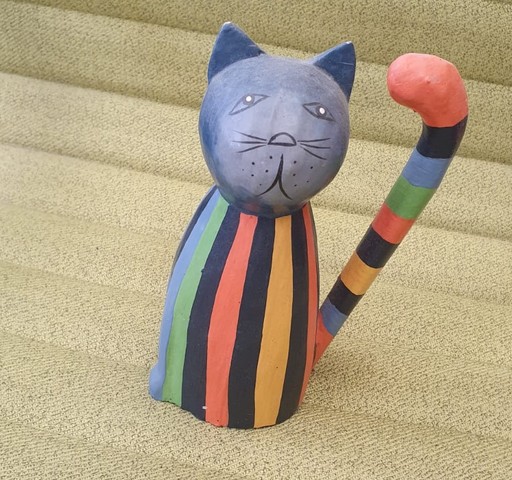} &
            \includegraphics[width=0.16\textwidth]{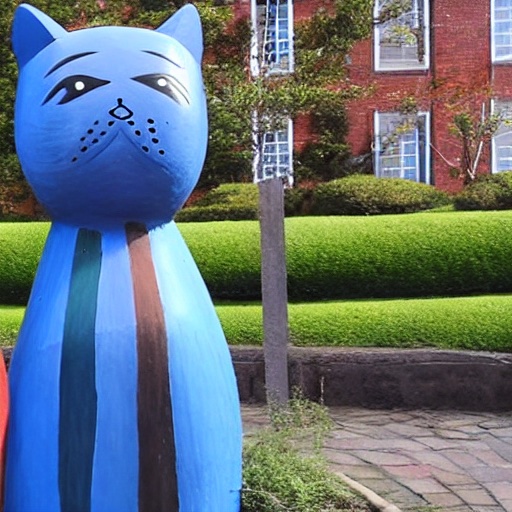} &
            \includegraphics[width=0.16\textwidth]{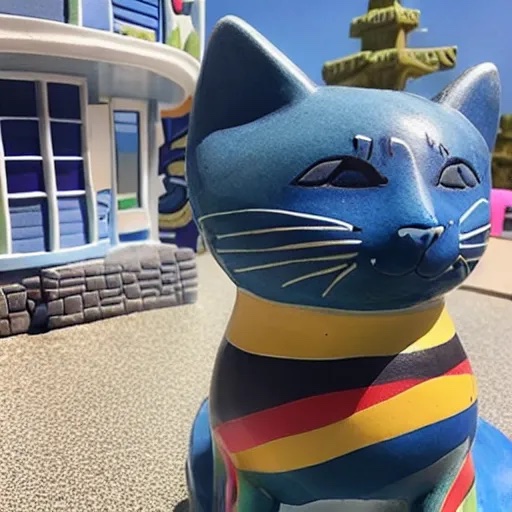} &
            \includegraphics[width=0.16\textwidth]{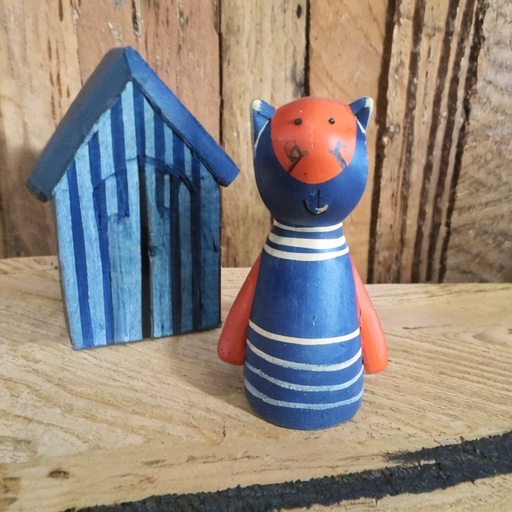} &
            \includegraphics[width=0.16\textwidth]{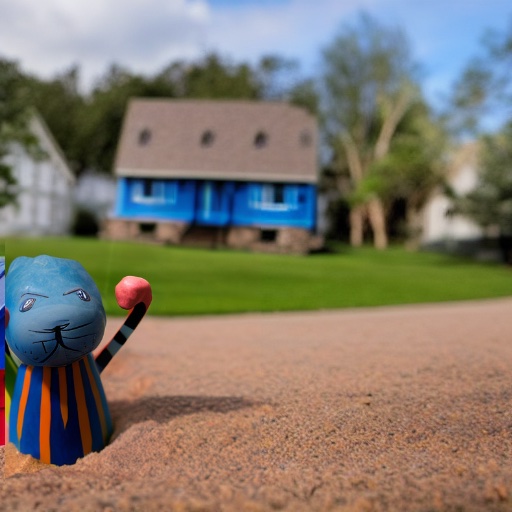} &
            \includegraphics[width=0.16\textwidth]{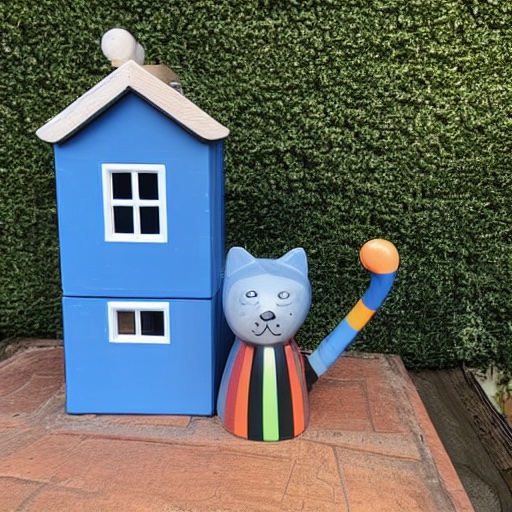} \\
            \end{tabular}
            \vspace{-5pt} 
            \caption*{a $toy^*$ with a blue house in the background}
        \end{subfigure}
        \caption{Comparison to non-fine-tuning methods TI (Textual Inversion), BLIP-D (BLIP-Diffusion), IP-A (IP-Adapter) and AnyDoor.}
        \label{fig:non-finetune-comparison}
        \vspace{-10pt}
\end{figure}

In \cref{fig:guided-image-synthesis}, we present qualitative results. Across different subjects and prompts, DreamBlend is able to generate images preserving the layout of the underfit reference image as well as the identity of the subjects. In \cref{fig:qualitative-comparison}, we present a qualitative comparison with DreamBooth and Custom Diffusion on some challenging prompts. In each case, we manually select the best results for each method through visual inspection. Our method is able to generate images that achieve higher subject fidelity, prompt fidelity and diversity compared to the baselines. In \cref{fig:non-finetune-comparison}, we present a comparison with non-fine-tuning methods, revealing that these approaches fall short in subject fidelity and photorealism, particularly for complex subjects. Additional results are in \cref{sec:more-qualitative-results}.

\subsection{Quantitative evaluation} 

\begin{figure}
    \centering
    \begin{subfigure}{0.49\linewidth}
        \centering
        \includegraphics[width=\linewidth]{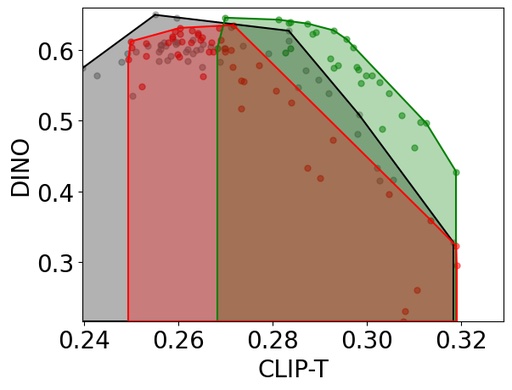}
        \caption{candle}
        \label{fig:plot1}
    \end{subfigure}
    \begin{subfigure}{0.49\linewidth}
        \centering
        \includegraphics[width=\linewidth]{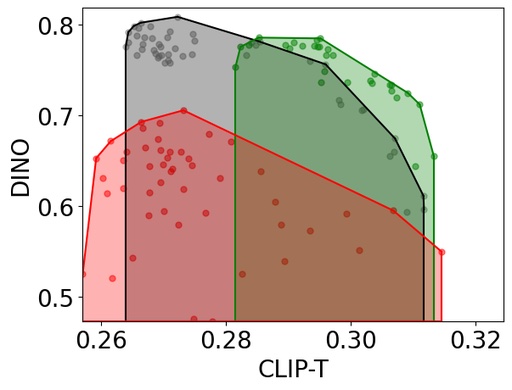}
        \caption{bear plushie}
        \label{fig:pareto}
    \end{subfigure}
    \caption{Image alignment (DINO) - text alignment (CLIP-T) space spanned by densely sampled operating points of DreamBooth (gray), Custom Diffusion (red) and our method (green) for two example subjects. Our method advances the pareto front and enables generation of images closer to top right corner [1,1] of the image-text alignment space, inaccessible to existing methods.}
    \label{fig:pareto}
    \vspace{-15pt}
\end{figure}

\cref{tab:quant} shows the metrics averaged over all subjects and prompts of DB benchmark. Our approach achieves the best DINO, CLIP-I and CLIP-T scores, showing notable improvement in DINO and CLIP-T. As DINO is not trained to ignore differences between images that might have similar text descriptions, it is better at capturing subtle differences in subject fidelity, as also noted in DreamBooth \cite{ruiz2023dreambooth}.

In \cref{fig:pareto}, we visualize how DreamBlend advances the pareto front for two example subjects, averaged over all 25 prompts. As each subject has a different fine-tuning trajectory, it is not possible to average metrics for different subjects after a certain number of fine-tuning steps and we present the result for each subject separately. Compared to all densely sampled operating points of both DB and CD, DreamBlend achieves better trade-offs, enabling better subject fidelity and prompt fidelity. Results for more subjects are in \cref{sec:more-quantitative-results}.

\begin{table}
  \centering
  {\small{
  \begin{tabular}{@{}lcccc@{}}
    \toprule
    Method &  Backbone & DINO & CLIP-I & CLIP-T\\
    \midrule
    Real Images  &  - & 0.774 & 0.885 & -\\
    \midrule
    Textual Inversion \cite{gal2022image} & SD1.5 & 0.569 & 0.780 & 0.255 \\
    BLIP-Diffusion \cite{li2024blip} (ZS) & SD1.5 & 0.594 & 0.779 & 0.300 \\

    DreamBooth \cite{ruiz2023dreambooth} & SD1.5 & 0.659 & 0.805& 0.296 \\
    CustomDiffusion \cite{kumari2023multi} & SD1.5 & 0.627 & 0.789 & 0.286 \\
    DreamBlend (Ours)  & SD1.5 & \textbf{0.675} & \textbf{0.808} & \textbf{0.308} \\
    \midrule
    AnyDoor \cite{chen2023anydoor} & SD2.1 & 0.588 & 0.753 & 0.289 \\
    \midrule
    IP-Adapter \cite{ye2023ip} & SDXL & 0.584 & 0.786 & 0.303 \\     
    DB-LoRA &  SDXL & 0.669 & 0.794 & 0.312 \\
    DreamBlend (Ours) & SDXL &  \textbf{0.678} & \textbf{0.795} & \textbf{0.313} \\
    \bottomrule
  \end{tabular}
  }}
  \caption{Quantitative evaluation on CLIP-I, CLIP-T and DINO}
  \label{tab:quant}
\end{table}

\begin{table}
  \centering
  {\small{
  \begin{tabular}{@{}lcc@{}}
    \toprule
     Study & Overall & Diversity\\
    \midrule
    over DB & 61.11 ($3e^{-12}$, [58.1, 64.1]) & 61.82 ($3e^{-09}$, [58.1, 65.6]) \\
    over CD & 70.16 (2$e^{-20}$, [66.2, 74.1]) & 72.70 ($2e^{-35}$, [69.5, 75.9]) \\
    \bottomrule
  \end{tabular}
  }}
  \caption{Human preference study (in \% of preference), with the Exact Binomial p-values and 95\% Confidence Intervals.}
  \label{tab:human-preference}
  \vspace{-10pt}
\end{table}

\subsection{Human preference study}
We conducted human preference studies comparing our approach to DB and CD baselines. Two studies were performed, assessing overall preference and diversity, for each baseline. In the overall preference study, users chose between an image generated by our method and a baseline method for the same text prompt, considering both subject and prompt fidelity. In the diversity study, users selected the more diverse collection of four images between our method and a baseline. The results of human preference study are presented in \cref{tab:human-preference}, where our method is preferred over baselines. We perform one-sample binomial test and results are statistically significant with very low p-values and lower bound of 95\% confidence intervals always greater than 50\%. More details in \cref{sec:human-preference-study}.

\section{Discussion}
\label{sec:discussion}
\begin{figure}

\begin{subfigure}{1.0\linewidth}
    \centering
    \setlength{\tabcolsep}{1pt}
    \begin{tabular}{cccc}
    {Input} & {Overfit} & {Underfit} & {Ours} \\
    \includegraphics[width=0.2\textwidth]{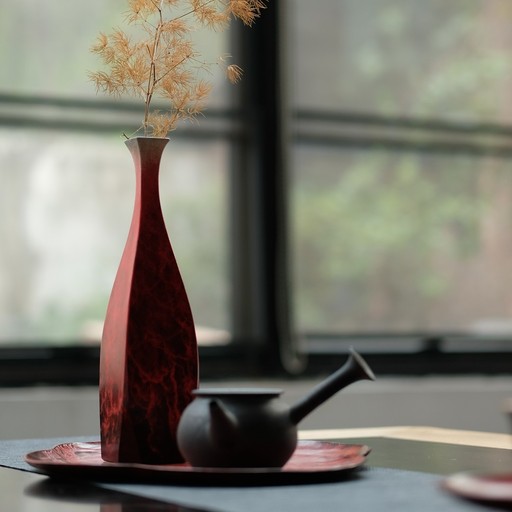} &
    \includegraphics[width=0.2\textwidth]{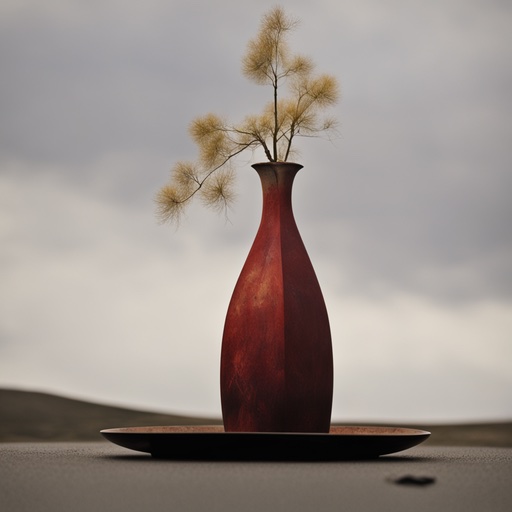} &
    \includegraphics[width=0.2\textwidth]{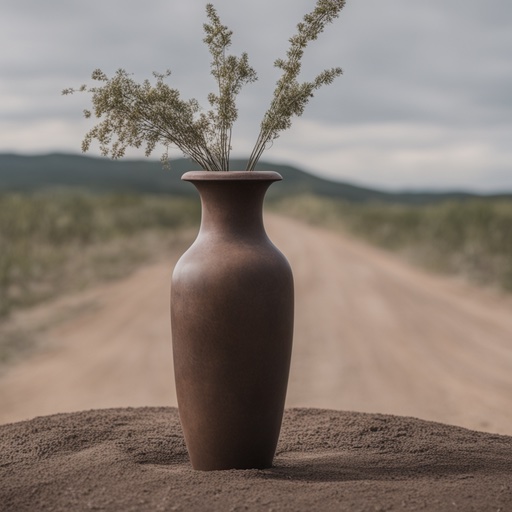} &
    \includegraphics[width=0.2\textwidth]{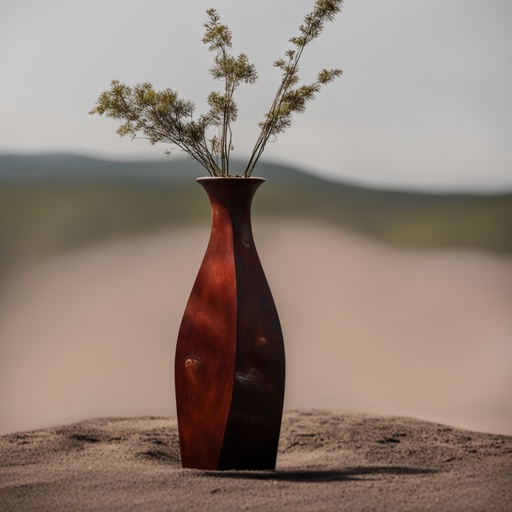} 
    \end{tabular}
    \vspace{-5pt} 
    \caption*{Backbone: SDXL. Prompt: a $vase^*$ on top of a dirt road}
\end{subfigure}
\begin{subfigure}{1.0\linewidth}
    \centering
    \setlength{\tabcolsep}{1pt}
    
    \begin{tabular}{cccc}
    \raisebox{0.1\linewidth} {\begin{tabular}[t]{@{}c@{}} 
    \hspace*{10pt} 
    \includegraphics[width=0.1\linewidth]{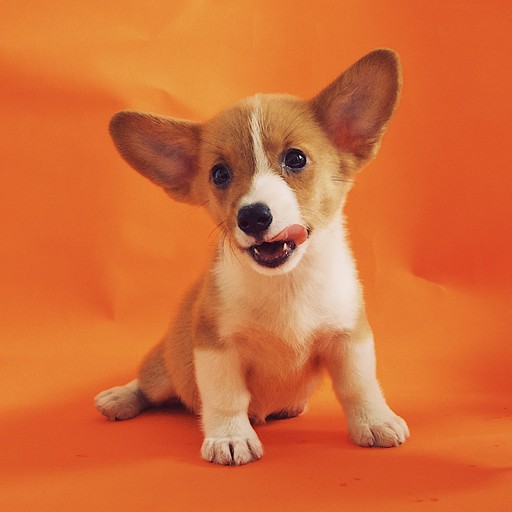} 
    \hspace*{10pt} 
    \\
    \includegraphics[width=0.1\linewidth]{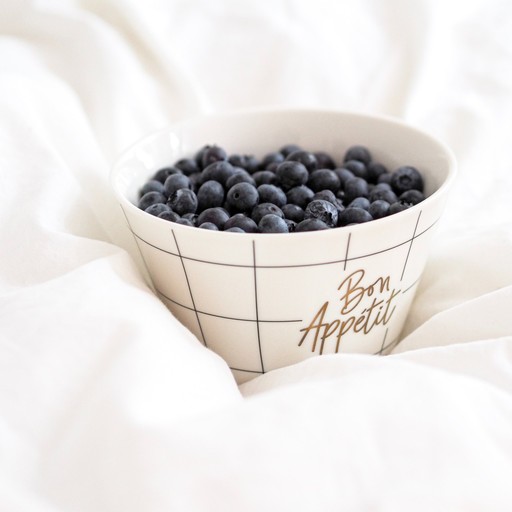}
    \end{tabular}}

     &
    \includegraphics[width=0.2\textwidth]{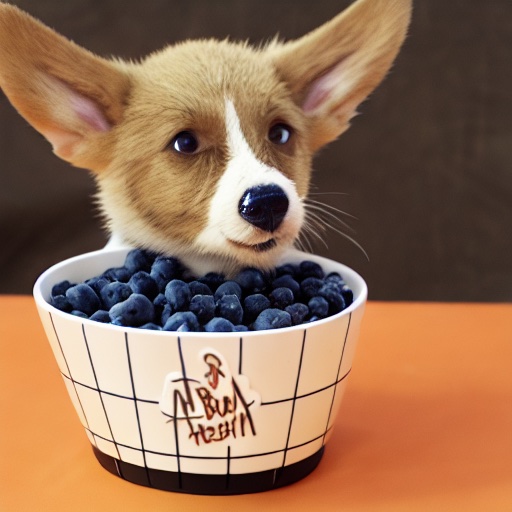} &
    \includegraphics[width=0.2\textwidth]{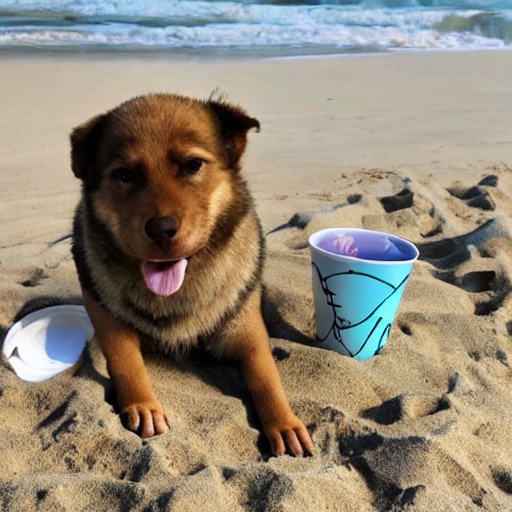} &
    \includegraphics[width=0.2\textwidth]{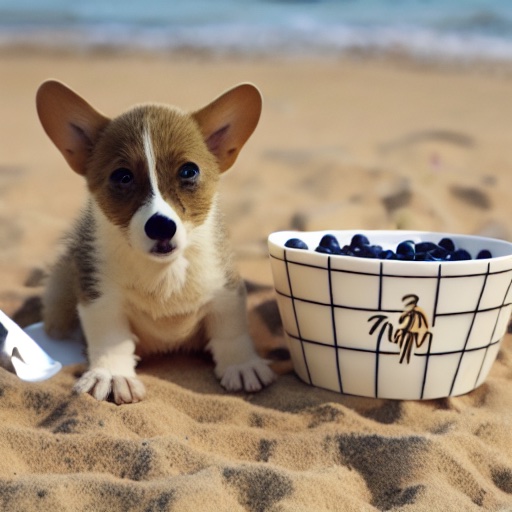} 
    \end{tabular}
    \vspace{-5pt} 
    \caption*{Backbone: SD1.4. Prompt: a $dog^*$ with a $bowl^*$ on a beach}
\end{subfigure}

\begin{subfigure}{1.0\linewidth}
    \centering
    \setlength{\tabcolsep}{1pt}
    \begin{tabular}{cccc}
    \includegraphics[width=0.2\textwidth]{images/candle_input/00.jpg} &
    \includegraphics[width=0.2\textwidth]{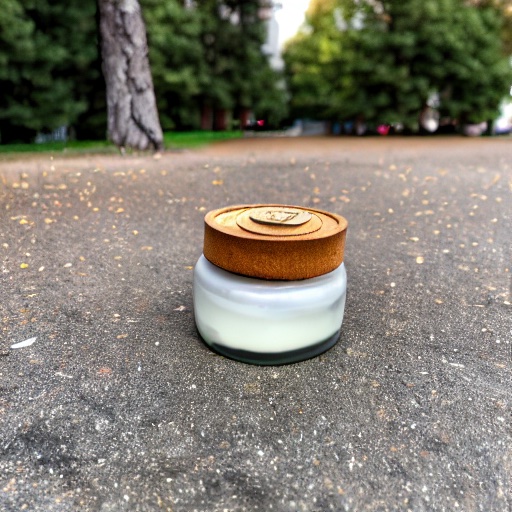} &
    \includegraphics[width=0.2\textwidth]{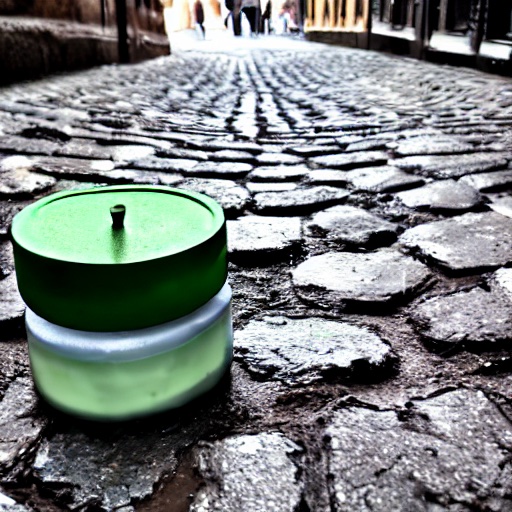} &
    \includegraphics[width=0.2\textwidth]{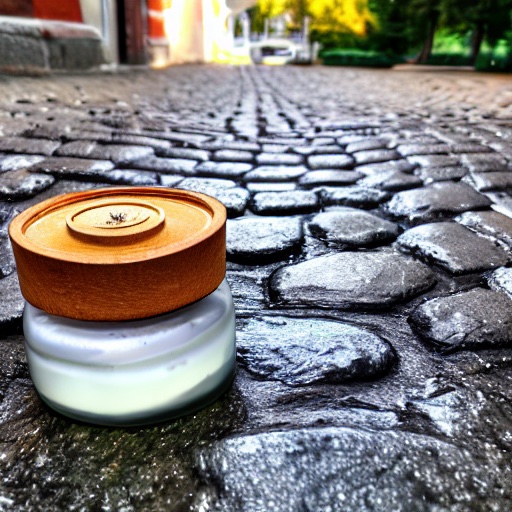} 
    \end{tabular}
    \vspace{-5pt} 
    \caption*{Backbone: SD1.5. Prompt: a $candle^*$ on a cobblestone street}
    \vspace{5pt} 
\end{subfigure} 
\hrule height 0.4pt
\begin{subfigure}{1.0\linewidth}
    \vspace{5pt}
    \centering
    \setlength{\tabcolsep}{1pt}
    \begin{tabular}{ccc}
    {$cat^*$} & {Real} & {Edited} \\
    \includegraphics[width=0.2\textwidth]{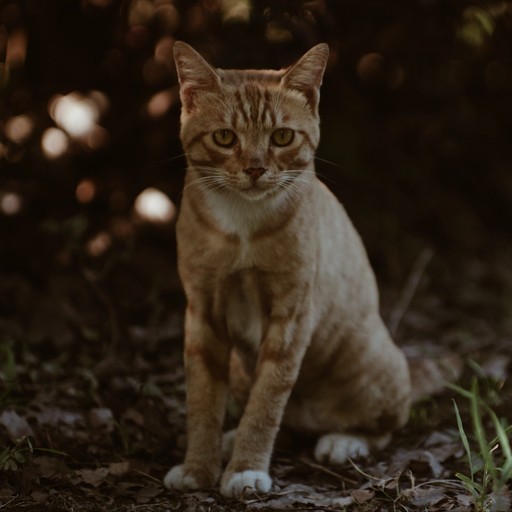} &
    \includegraphics[width=0.2\textwidth]{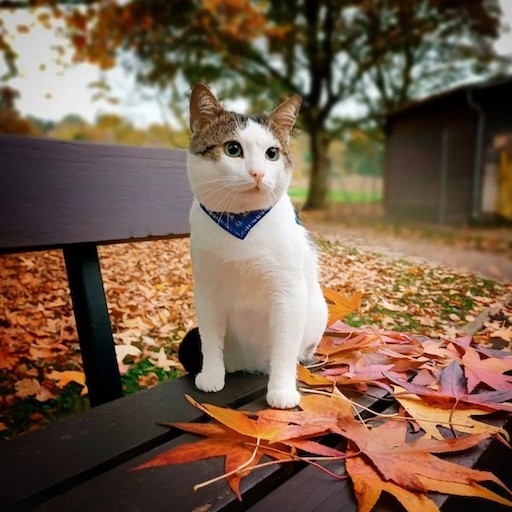} &
    \includegraphics[width=0.2\textwidth]{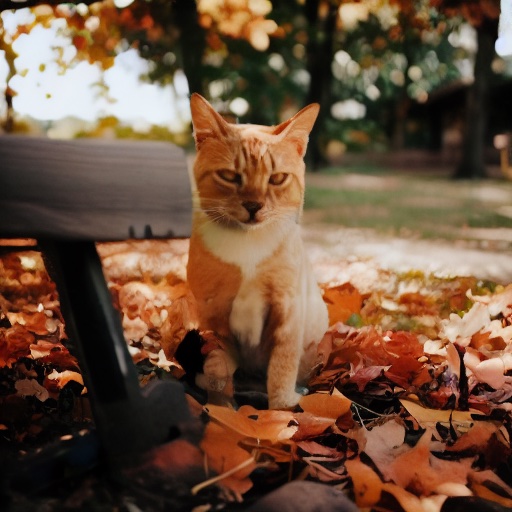} 
    \end{tabular}
    \vspace{-5pt} 
   \caption*{Personalized editing of real image: cat to $cat^*$}
    \label{fig:sub4}
\end{subfigure}

\caption{DreamBlend applied on different backbones, different fine-tuning techniques, real image editing. Top to bottom: SDXL fine-tuned with LoRA for $vase^*$, SDv1.4 fine-tuned with Custom Diffusion for multiple concepts $dog^*$ and $bowl^*$, SDv1.5 fine-tuned with Custom Diffusion for $candle^*$, personalized editing of real image with DDIM inversion.}
\label{fig:other-backbones}
\end{figure}

\textbf{Generalization:} Our approach is applicable to different fine-tuning methods like DreamBooth, Custom Diffusion and LoRA, different text-to-image diffusion models that feature a cross attention mechanism, as well as personalized editing of real images, as shown in \cref{fig:other-backbones}.

\begin{figure}
    \centering
     \begin{subfigure}{0.18\linewidth}
        \centering
        \includegraphics[width=\linewidth]{images/backpack_input/00.jpg}
        \caption*{$backpack^*$}
        \label{fig:plot1}
    \end{subfigure}
    \begin{subfigure}{0.18\linewidth}
        \centering
        \includegraphics[width=\linewidth]{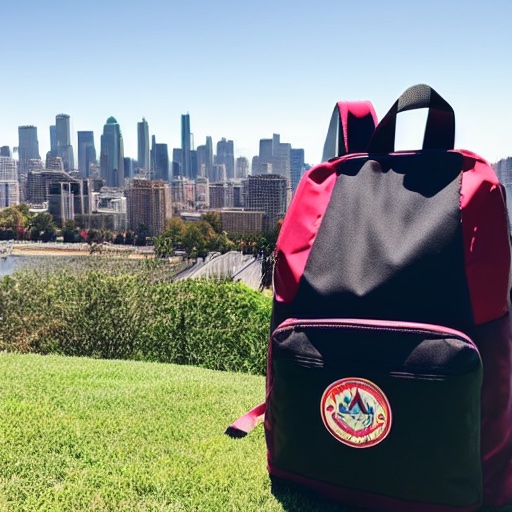}
        \caption*{Underfit}
        \label{fig:plot1}
    \end{subfigure}
    \begin{subfigure}{0.18\linewidth}
        \centering
        \includegraphics[width=\linewidth]{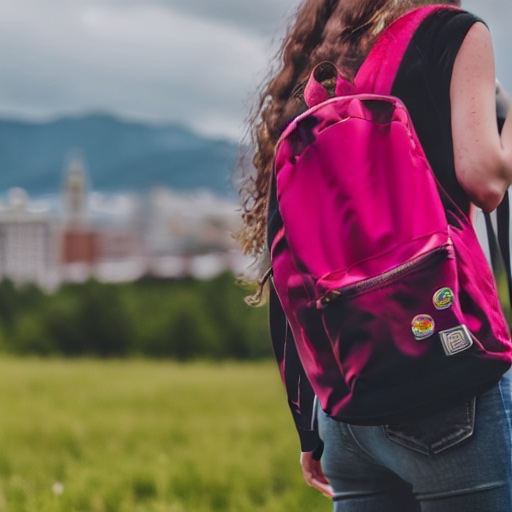}
        \caption*{$\alpha$=0}
        \label{fig:plot1}
    \end{subfigure}
    \begin{subfigure}{0.18\linewidth}
        \centering
        \includegraphics[width=\linewidth]{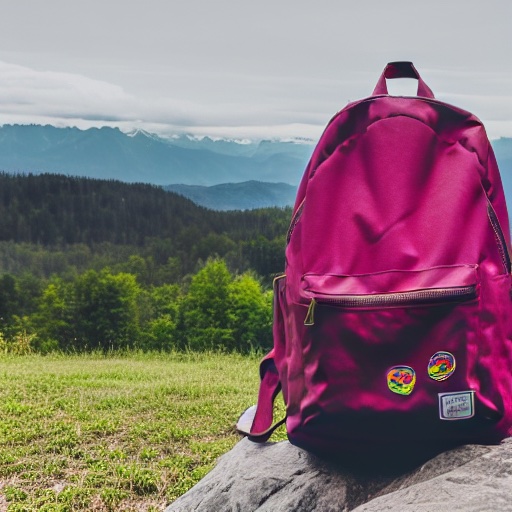}
        \caption*{$\alpha$=0.1}
        \label{fig:plot2}
    \end{subfigure}
    \begin{subfigure}{0.18\linewidth}
        \centering
        \includegraphics[width=\linewidth]{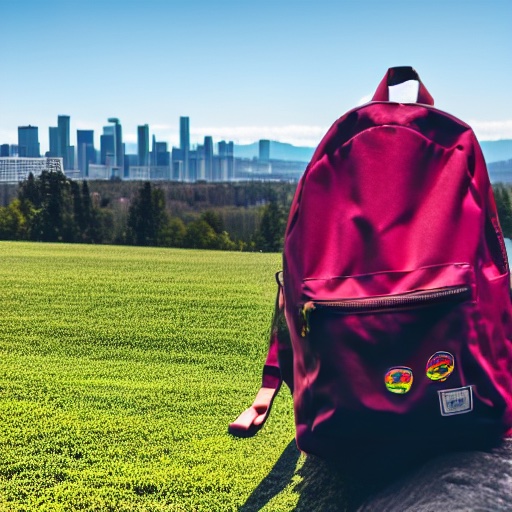}
        \caption*{$\alpha$=0.2}
        \label{fig:plot3}
    \end{subfigure}
    \caption{Effect of Cross Attention Guidance Scale $\alpha$}
    \label{fig:cag_effect}
\end{figure}

\textbf{Effect of cross attention guidance:} An over-fitted model tends to generate concepts it has seen in input images of the subject, reducing prompt fidelity and diversity. Cross attention guidance encourages it to follow the layout of the reference underfit image instead. \cref{fig:cag_effect} shows the effect of varying cross attention guidance scale $\alpha$ in \cref{alg-guided-image-synthesis}.

\begin{figure}
    \centering
     \begin{subfigure}{0.23\textwidth}
        \centering
        \includegraphics[width=\linewidth]{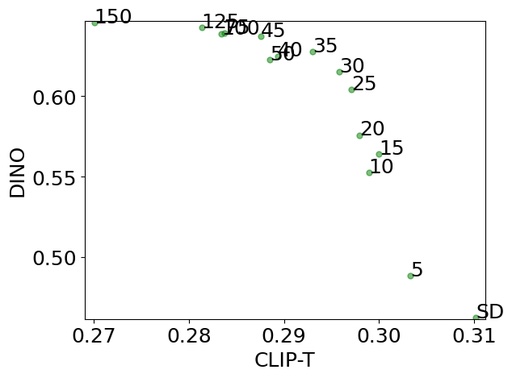}
        \captionsetup{justification=centering}
        \caption{Different guidance models with step 200 edit model}
        \label{fig:guidance}
    \end{subfigure}
    \begin{subfigure}{0.23\textwidth}
        \centering
        \includegraphics[width=\linewidth]{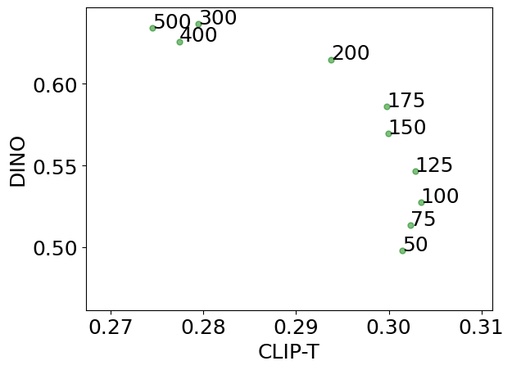}
        \captionsetup{justification=centering}
        \caption{Different edit models with step 25 guidance model}
        \label{fig:edit}
    \end{subfigure}
    \caption{Effect of using different checkpoints as guidance and edit models for $candle^*$, numbers denote fine-tuning steps.}
    \label{fig:choice_g_e}
    \vspace{-10pt}
\end{figure}

\textbf{Choice of guidance and edit models:} Optimal performance in the guidance model is achieved by selecting an underfit checkpoint with some subject resemblance, ensuring successful edits while maintaining prompt fidelity. Conversely, the edit model benefits from choosing a checkpoint that has learnt the subject without experiencing catastrophic attention collapse [\ref{fig:cross-attention-maps}]. \cref{fig:choice_g_e} presents these trade-offs quantitatively for an example subject $candle^*$.

\begin{figure}

\begin{subfigure}{1.0\linewidth}
    \centering
    \setlength{\tabcolsep}{1pt}
    \begin{tabular}{ccccc}
    {Input} & {Overfit} & {SD} & {Underfit} & {Ours} \\
    \includegraphics[width=0.2\textwidth]{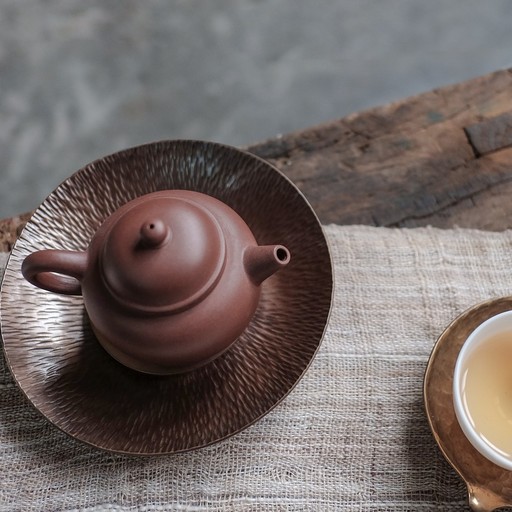} &
    \includegraphics[width=0.2\textwidth]{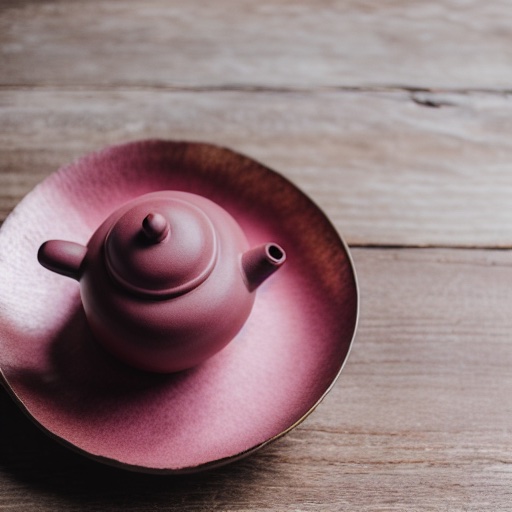} &
    \includegraphics[width=0.2\textwidth]{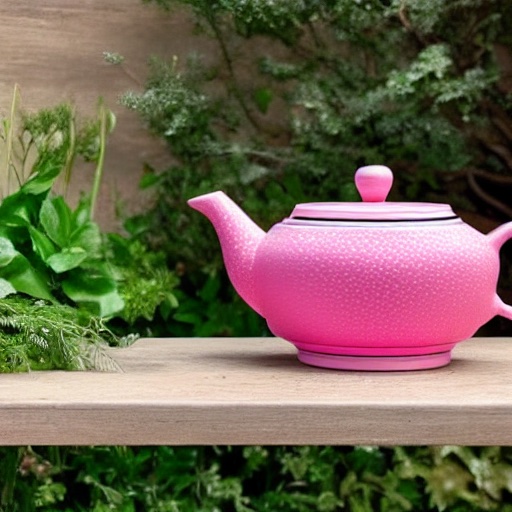} &
    \includegraphics[width=0.2\textwidth]{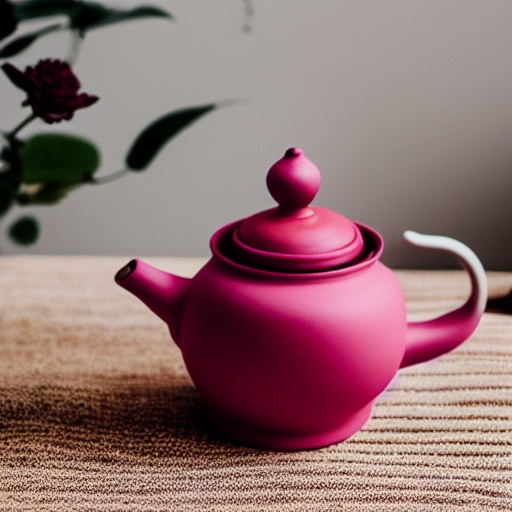} &
    \includegraphics[width=0.2\textwidth]{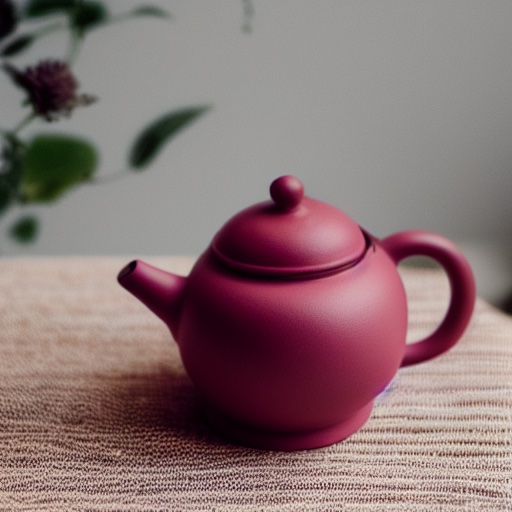} 
    \end{tabular}
    \vspace{-5pt} 
    \caption*{a $teapot^*$ on top of pink fabric}
   \vspace{5pt} 
\end{subfigure}
\hrule height 0.4pt
\begin{subfigure}{1.0\linewidth}
\vspace{5pt}
    \centering
    \setlength{\tabcolsep}{1pt}
    \begin{tabular}{cccc}
    {Input} & {Overfit} & {Underfit} & {Ours} \\
    \includegraphics[width=0.2\textwidth]{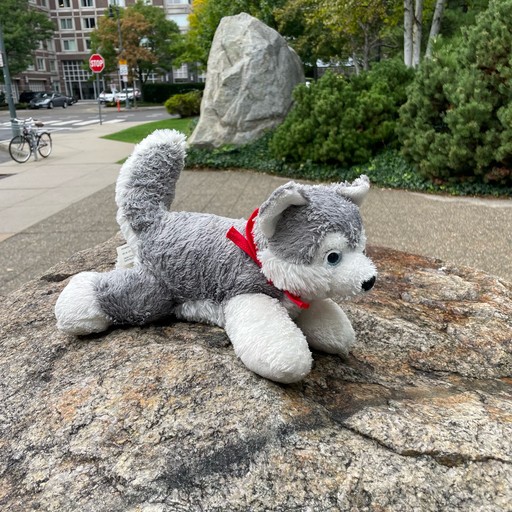} &
    \includegraphics[width=0.2\textwidth]{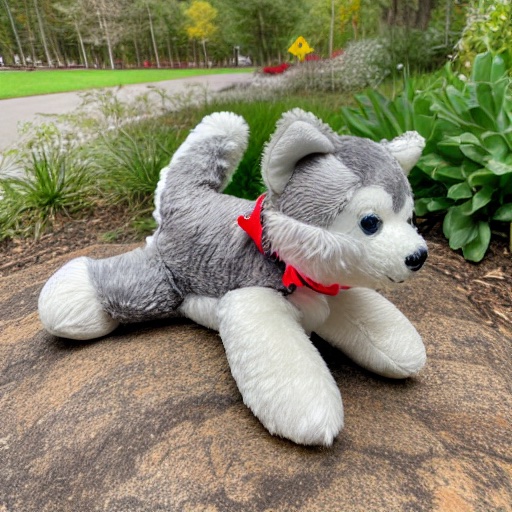} &
    \includegraphics[width=0.2\textwidth]{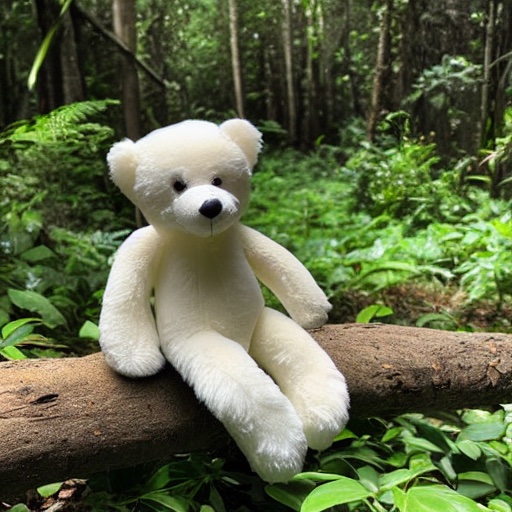} &
    \includegraphics[width=0.2\textwidth]{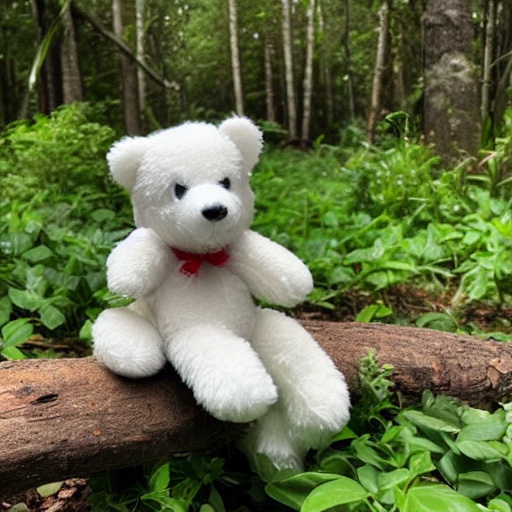} 
    \end{tabular}
    \vspace{-5pt} 
    \caption*{a \textit{stuffed animal$^*$} in the jungle}
\end{subfigure}

\caption{Failure Cases. Top: SD and underfit model fail to generate images that follow the prompt. Bottom: Shape of the subject in underfit image is too different from the actual subject for a successful edit.}
\label{fig:failure_cases}
 \vspace{-5pt}
\end{figure}

\textbf{Limitations:} Like other text-to-image personalization methods, we share the challenge of inheriting failures when the pre-trained model fails to generate an image with high prompt fidelity for a text prompt as shown in the top row of \cref{fig:failure_cases}. Yet, in cases where the model initially succeeds but that knowledge is gradually lost in the fine-tuning, our approach combines benefits at different operating points for effective image synthesis. Similar to other fine-tuning based approaches, our success hinges on optimal selection of operating points. If the edit model is too over-fit, cross attention guidance becomes ineffective, as shown in \cref{fig:cross-attention-maps}. If the subject in the guidance image is too different from the actual subject, DreamBlend may fail to perform a successful edit, as shown in bottom row of \cref{fig:failure_cases}. By leveraging two checkpoints for inference, our method may appear to have higher storage requirements. However, given that current fine-tuning methods also necessitate storing checkpoints for sampling and selection, our storage needs are comparable in practice. 

\section{Conclusions}

We presented DreamBlend, an approach that combines prompt fidelity and diversity from earlier checkpoints and subject fidelity from later checkpoints during image generation. It is straightforward and efficient, requiring only inference-time adjustments to existing techniques. It generates images with better subject fidelity, prompt fidelity and diversity, advancing the pareto front and surpassing state-of-the-art fine-tuning methods. Importantly, it successfully generates high fidelity images for challenging prompts, on which existing approaches struggled. For future work, we will explore utilizing this idea of  a regularization on cross attention maps to combat over-fitting and reconstructability-editability trade-offs in other scenarios.

\section{Societal impact}
Fine-tuning text-to-image diffusion models has democratized the creation of personalized visuals such as pets, furniture, or self-portraits, making it accessible compared to training large models from scratch. Technologies like DreamBlend improve image fidelity across diverse contexts, enhancing creative potential in various fields. However, this advancement also brings associated risks, including concerns about copyright, privacy, authenticity, and the potential for fake media to propagate misinformation. To responsibly deploy these technologies in production settings, it is essential to implement safeguards and mitigation strategies, such as detecting AI-generated content. Moreover, disseminating knowledge about these technologies' inner workings is critical for promoting responsible innovation.

{\small
\bibliographystyle{ieee_fullname}
\bibliography{main}
}

\newpage 

\appendix
\section*{Appendix}

In \cref{sec:more-qualitative-results}, we present more qualitative results in addition to \cref{fig:guided-image-synthesis},  \cref{fig:qualitative-comparison} and \cref{fig:other-backbones} of the main text. In \cref{sec:more-quantitative-results}, we visualize how DreamBlend advances the pareto front for more example subjects from DreamBooth benchmark, in addition to \cref{fig:pareto} of the main text. In \cref{sec:human-preference-study}, we explain details of the human preference studies conducted, present examples of the user interface used and validate statistical significance. In \cref{sec:gs_ags}, we present the effects of varying the cross attention guidance and classifier-free guidance. In \cref{sec:implementation-details}, we present some implementation details. In \cref{sec:non-fine-tuning-methods}, we present comparisons to non-fine-tuning based text-to-image personalization methods.

\section{More qualitative results}
\label{sec:more-qualitative-results}
\begin{figure*}
\centering



\begin{subfigure}{0.49\textwidth}
    \centering
    \setlength{\tabcolsep}{1pt}
    \begin{tabular}{cccc}
    {Input} & {Overfit} & {Underfit} & {Ours} \\
     \includegraphics[width=0.2\textwidth]{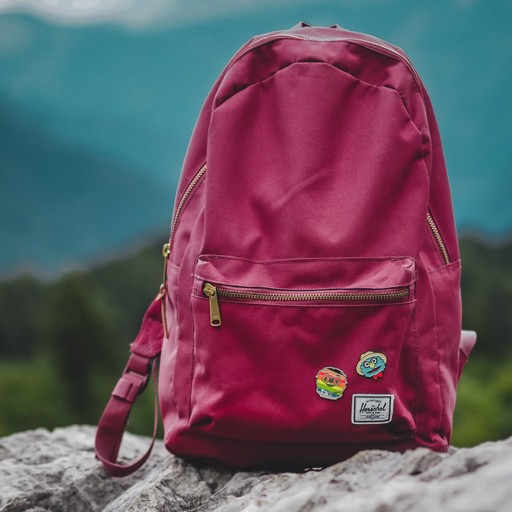} &
    \includegraphics[width=0.2\textwidth]{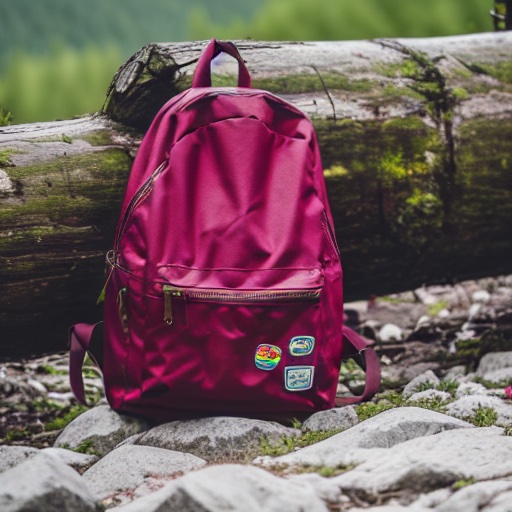} &
    \includegraphics[width=0.2\textwidth]{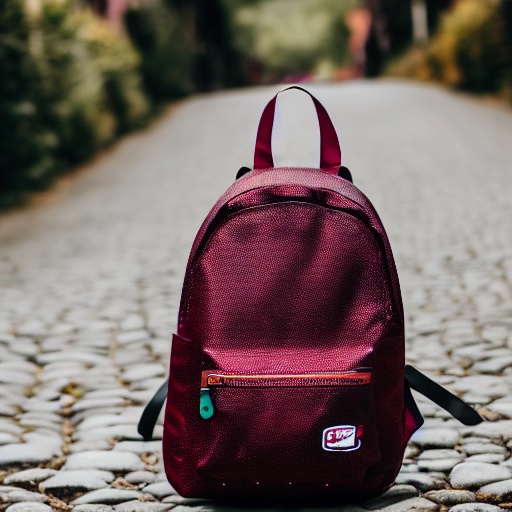} &
    \includegraphics[width=0.2\textwidth]{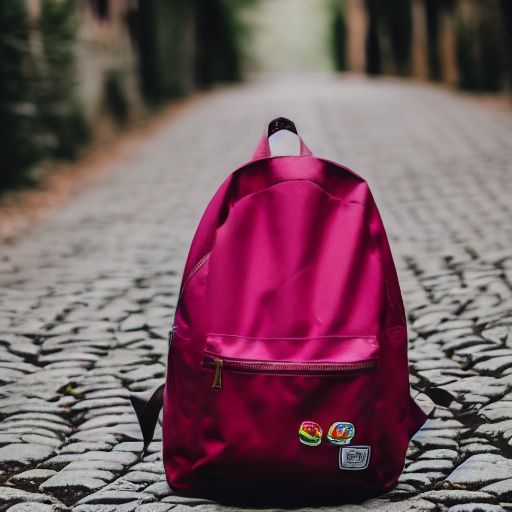} 
    \end{tabular}
    \vspace{-5pt} 
    \caption*{a $backpack^*$ on a cobblestone street}
    \vspace{10pt} 
    \label{fig:sub4}
\end{subfigure}
\begin{subfigure}{0.49\textwidth}
    \centering
    \setlength{\tabcolsep}{1pt}
    \begin{tabular}{cccc}
     {Input} & {Overfit} & {Underfit} & {Ours} \\
     \includegraphics[width=0.2\textwidth]{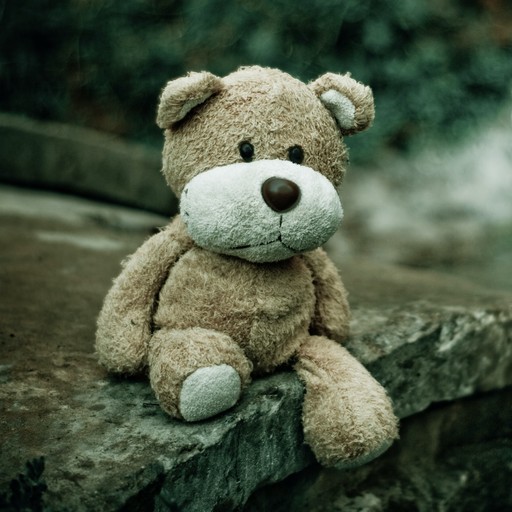} &
    \includegraphics[width=0.2\textwidth]{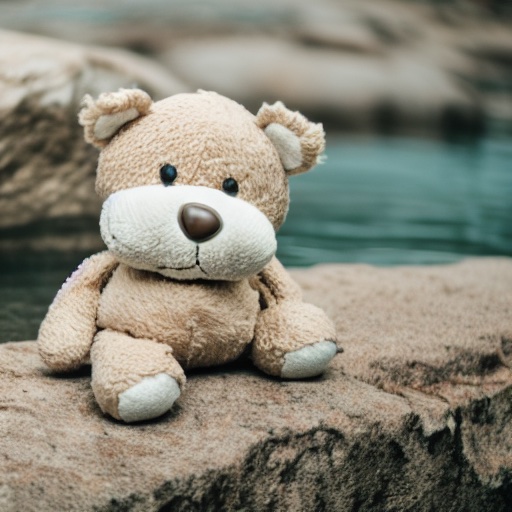} &
    \includegraphics[width=0.2\textwidth]{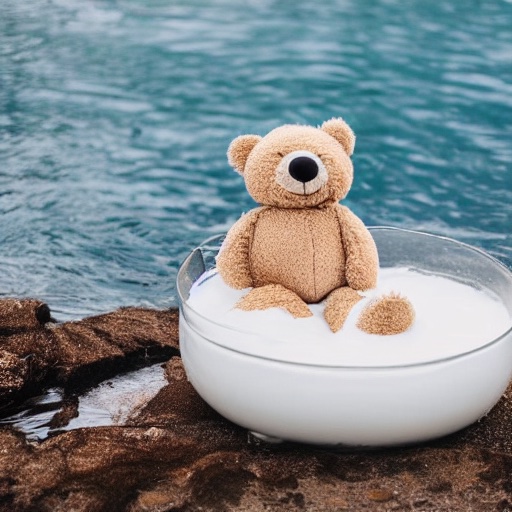} &
    \includegraphics[width=0.2\textwidth]{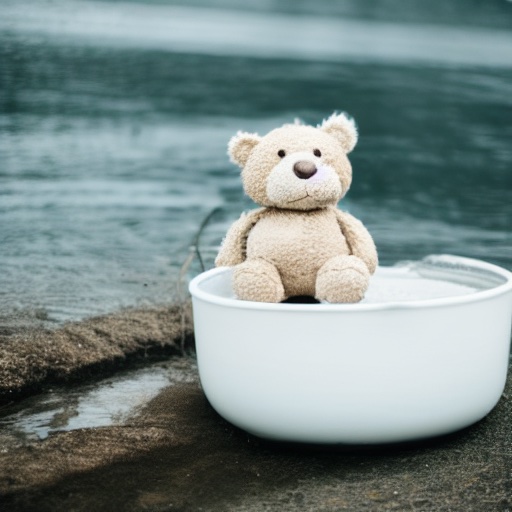} 
    \end{tabular}
    \vspace{-5pt} 
    \caption*{a $teddy^*$ floating in an ocean of milk}
    \vspace{10pt} 
    \label{fig:sub4}
\end{subfigure}

\begin{subfigure}{0.49\textwidth}
    \centering
    \setlength{\tabcolsep}{1pt}
    \begin{tabular}{cccc}
     \includegraphics[width=0.2\textwidth]{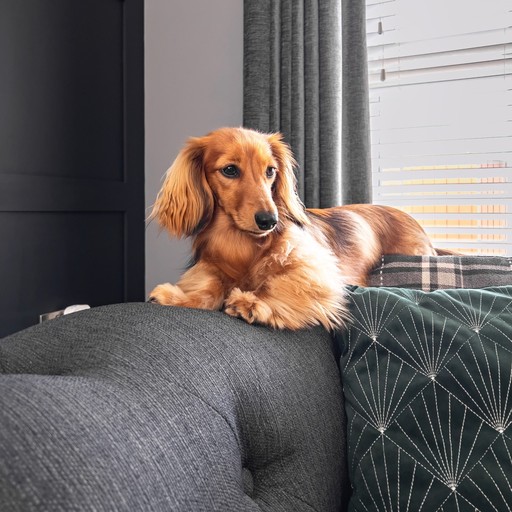} &
    \includegraphics[width=0.2\textwidth]{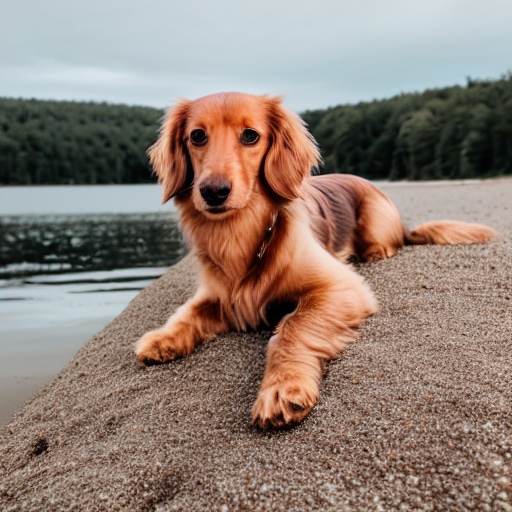} &
    \includegraphics[width=0.2\textwidth]{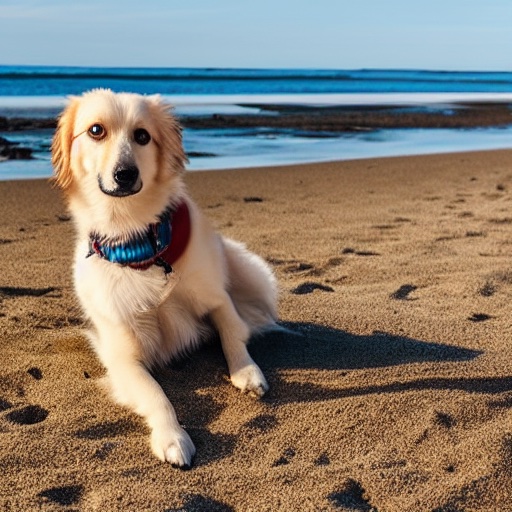} &
    \includegraphics[width=0.2\textwidth]{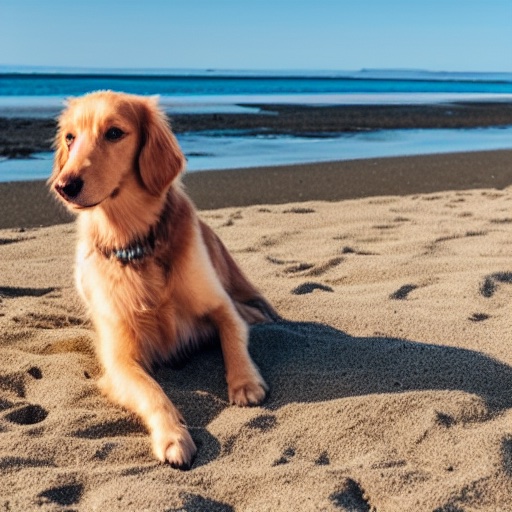} 
    \end{tabular}
    \vspace{-5pt} 
    \caption*{a $dog^*$ on the beach}
    \vspace{10pt} 
    \label{fig:sub4}
\end{subfigure}
\begin{subfigure}{0.49\textwidth}
    \centering
    \setlength{\tabcolsep}{1pt}
    \begin{tabular}{cccc}
    \includegraphics[width=0.2\textwidth]{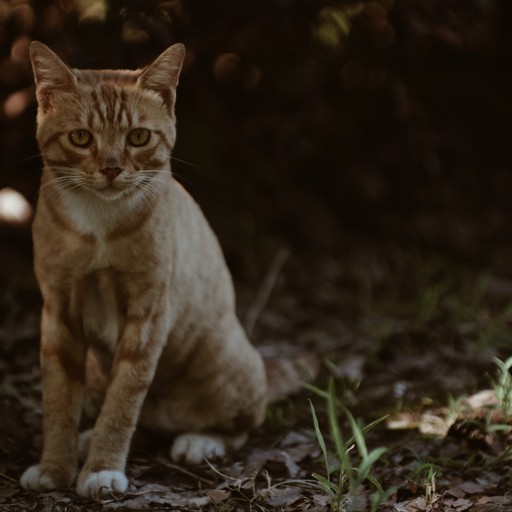} &
    \includegraphics[width=0.2\textwidth]{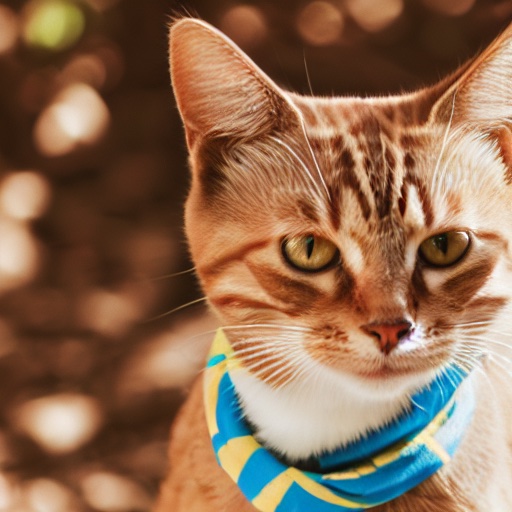} &
    \includegraphics[width=0.2\textwidth]{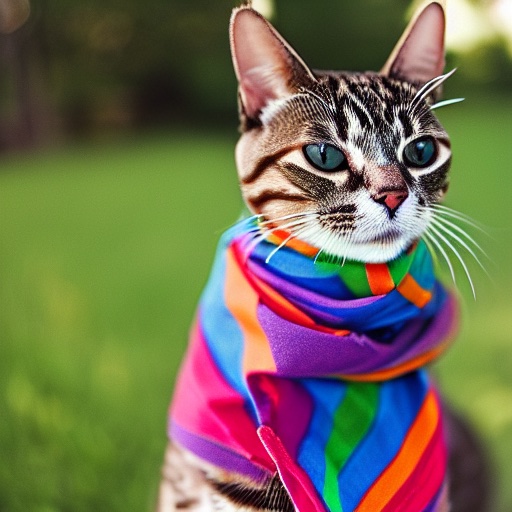} &
    \includegraphics[width=0.2\textwidth]{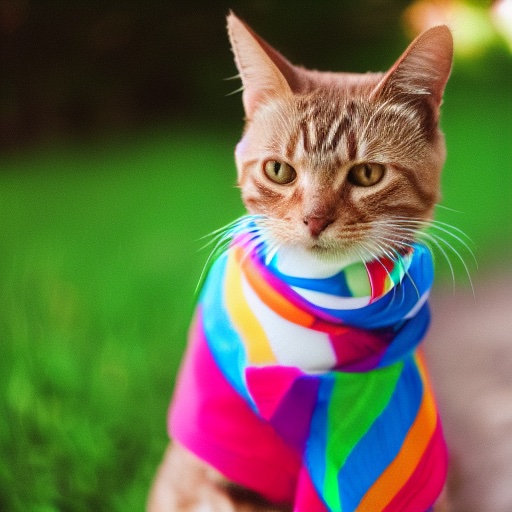} 
    \end{tabular}
    \vspace{-5pt} 
    \caption*{a $cat^*$ wearing a rainbow scarf}
    \vspace{10pt} 
    \label{fig:sub4}
\end{subfigure}

\begin{subfigure}{0.49\textwidth}
    \centering
    \setlength{\tabcolsep}{1pt}
    \begin{tabular}{cccc}
    \includegraphics[width=0.2\textwidth]{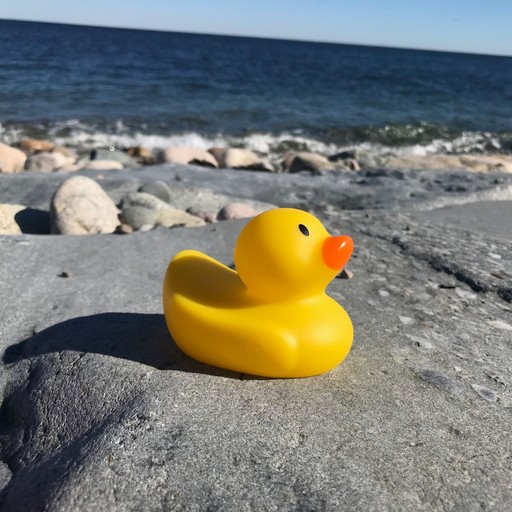} &
    \includegraphics[width=0.2\textwidth]{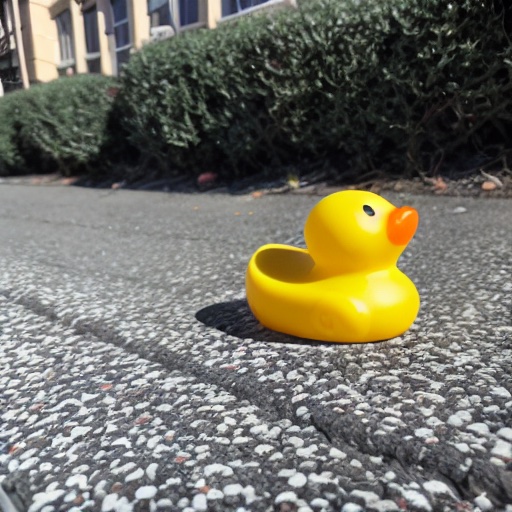} &
    \includegraphics[width=0.2\textwidth]{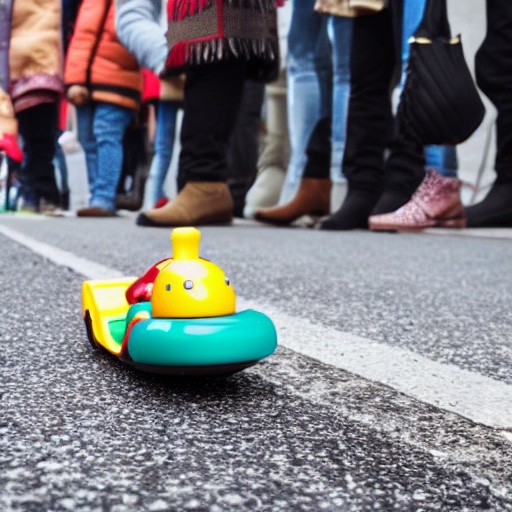} &
    \includegraphics[width=0.2\textwidth]{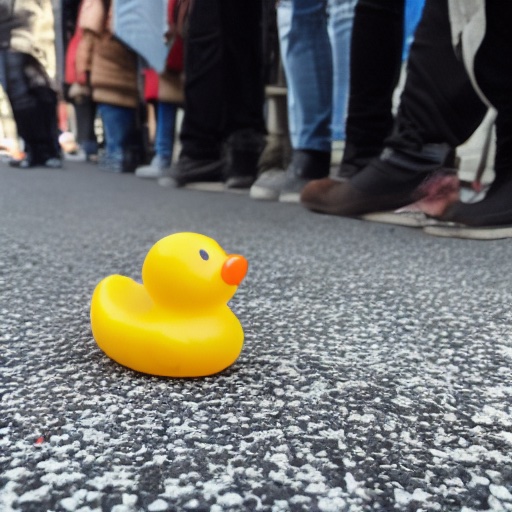} 
    \end{tabular}
    \vspace{-5pt} 
    \caption*{a $toy^*$ on top of the sidewalk in a crowded street}
    \vspace{10pt} 
    \label{fig:sub4}
\end{subfigure}
\begin{subfigure}{0.49\textwidth}
    \centering
    \setlength{\tabcolsep}{1pt}
    \begin{tabular}{cccc}
     \includegraphics[width=0.2\textwidth]{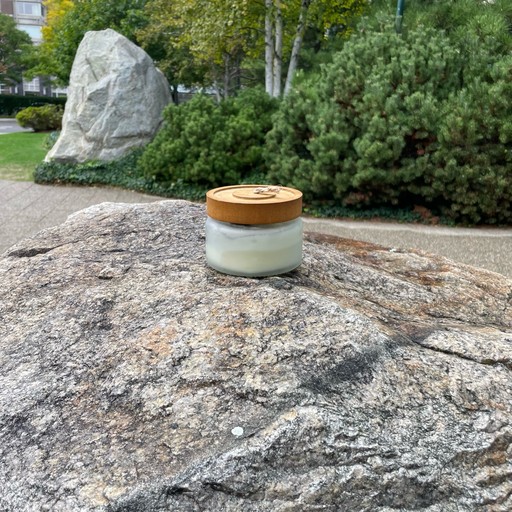} &
    \includegraphics[width=0.2\textwidth]{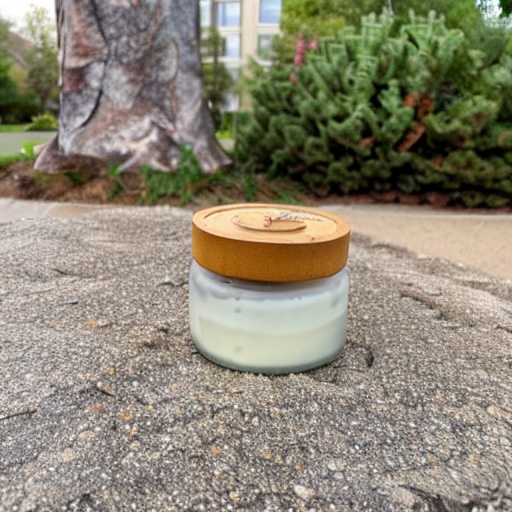} &
    \includegraphics[width=0.2\textwidth]{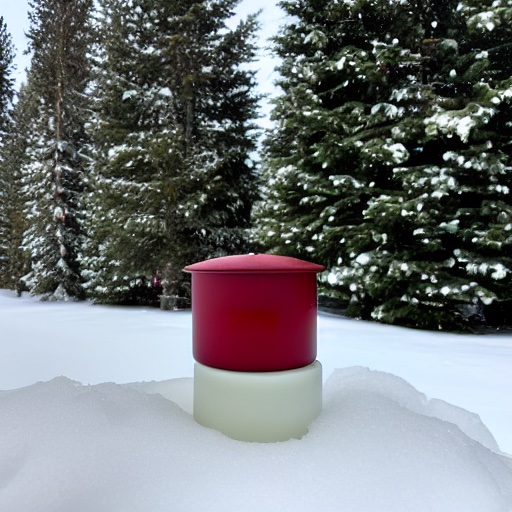} &
    \includegraphics[width=0.2\textwidth]{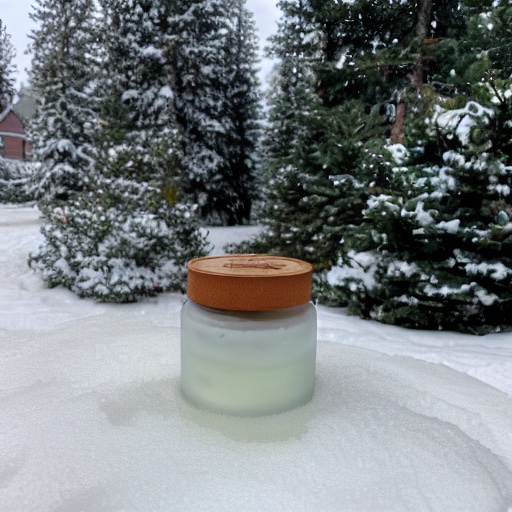} 
    \end{tabular}
    \vspace{-5pt} 
    \caption*{a $candle^*$ in the snow}
    \vspace{10pt} 
    \label{fig:sub4}
\end{subfigure}

\begin{subfigure}{0.49\textwidth}
    \centering
    \setlength{\tabcolsep}{1pt}
    \begin{tabular}{cccc}
     \includegraphics[width=0.2\textwidth]{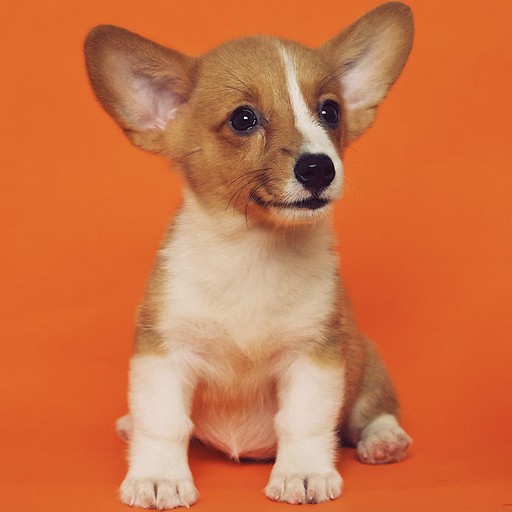} &
    \includegraphics[width=0.2\textwidth]{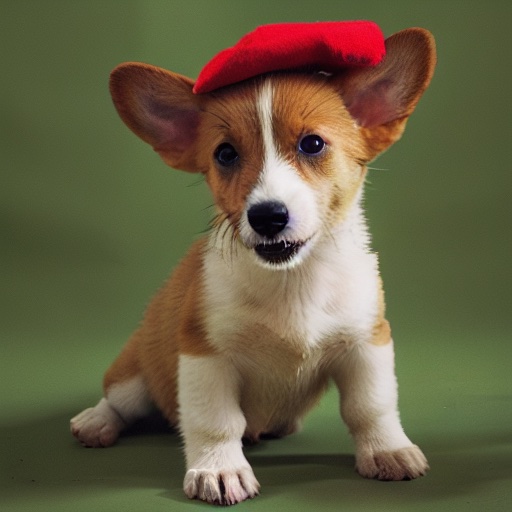} &
    \includegraphics[width=0.2\textwidth]{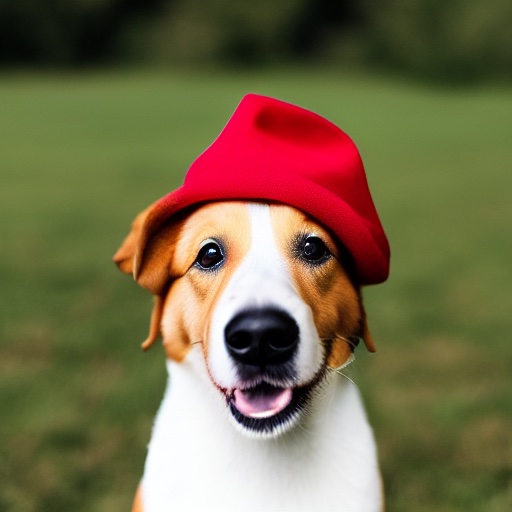} &
    \includegraphics[width=0.2\textwidth]{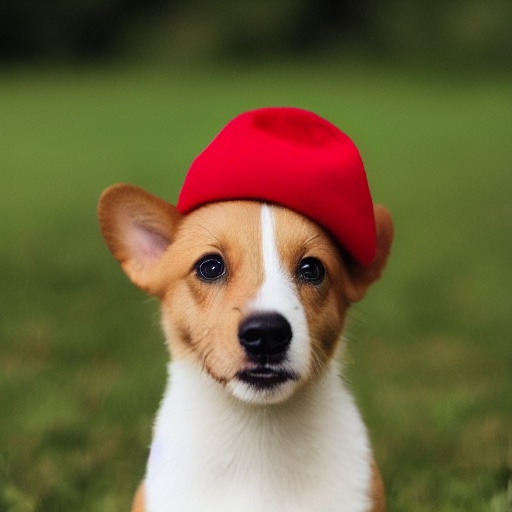} 
    \end{tabular}
    \vspace{-5pt} 
    \caption*{a $dog^*$ wearing a red hat}
    \vspace{10pt} 
    \label{fig:sub4}
\end{subfigure}
\begin{subfigure}{0.49\textwidth}
    \centering
    \setlength{\tabcolsep}{1pt}
    \begin{tabular}{cccc}
     \includegraphics[width=0.2\textwidth]{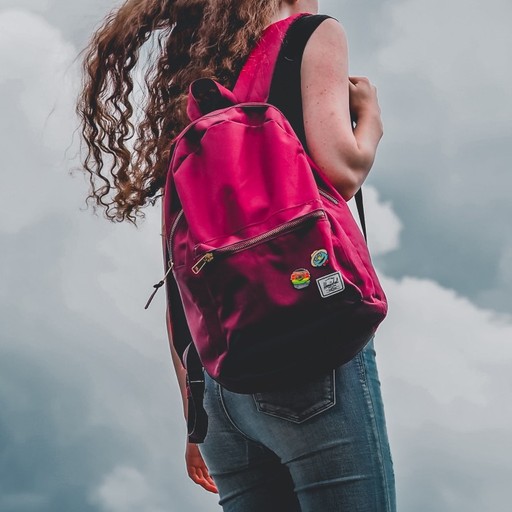} &
    \includegraphics[width=0.2\textwidth]{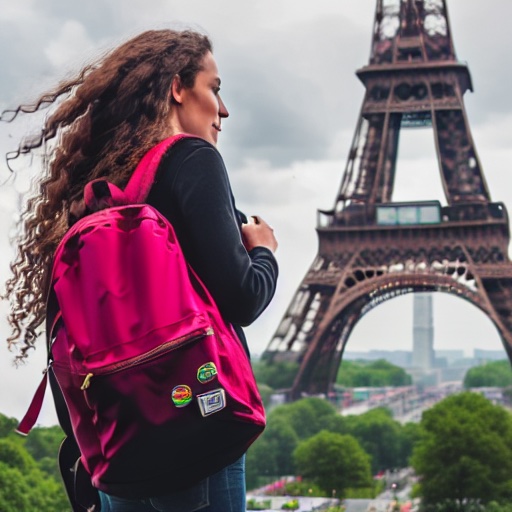} &
    \includegraphics[width=0.2\textwidth]{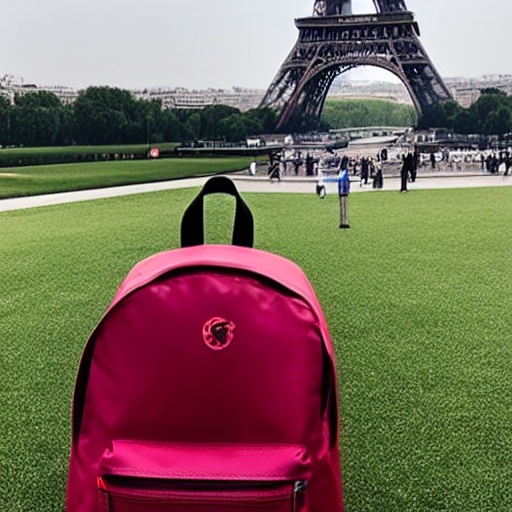} &
    \includegraphics[width=0.2\textwidth]{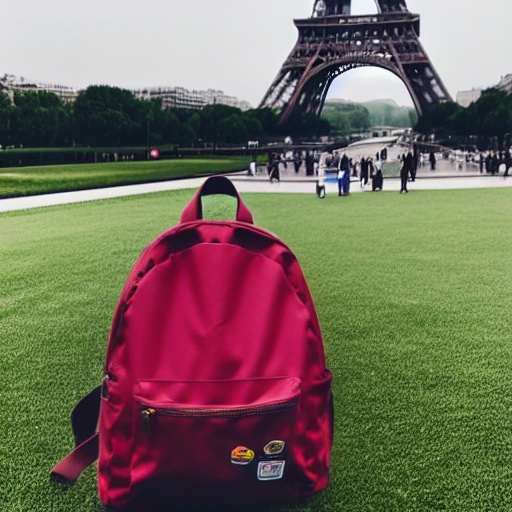} 
    \end{tabular}
    \vspace{-5pt} 
    \caption*{a $backpack^*$ with the Eiffel Tower in the background}
    \vspace{10pt} 
    \label{fig:sub4}
\end{subfigure}

\begin{subfigure}{0.49\textwidth}
    \centering
    \setlength{\tabcolsep}{1pt}
    \begin{tabular}{cccc}
     \includegraphics[width=0.2\textwidth]{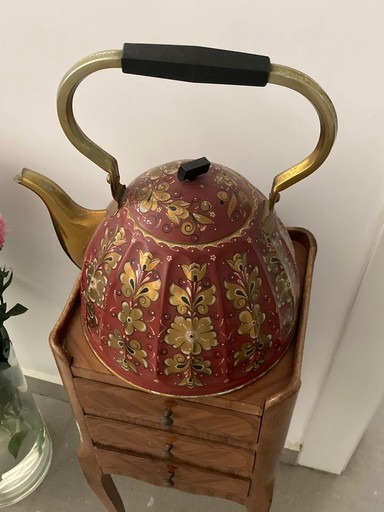} &
    \includegraphics[width=0.2\textwidth]{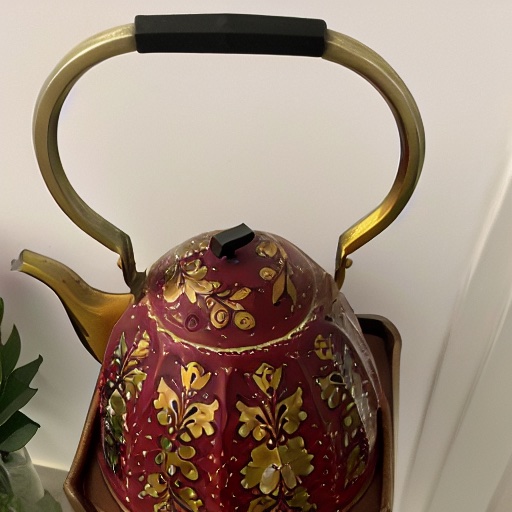} &
    \includegraphics[width=0.2\textwidth]{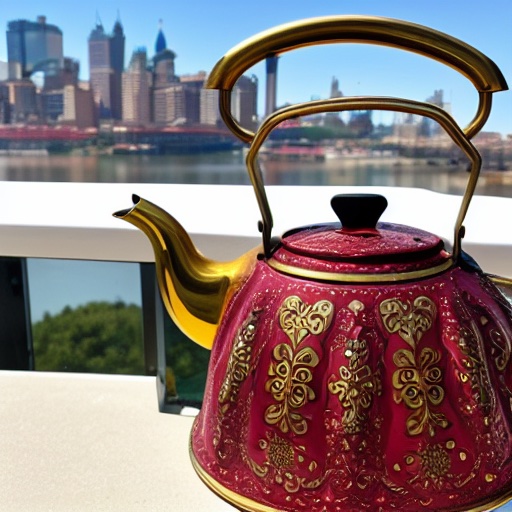} &
    \includegraphics[width=0.2\textwidth]{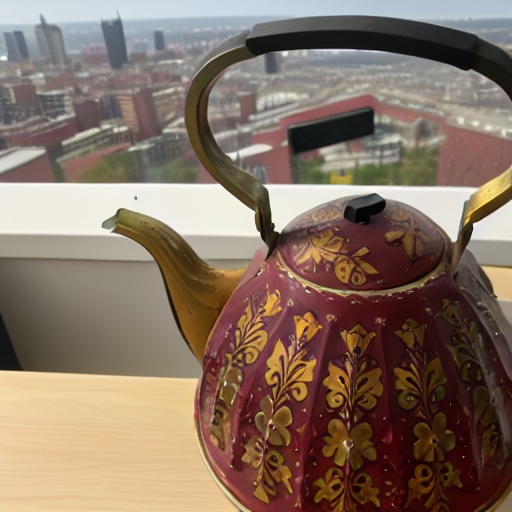} 
    \end{tabular}
    \vspace{-5pt} 
    \caption*{a $teapot^*$ with a city in the background}
    \vspace{10pt} 
    \label{fig:sub4}
\end{subfigure}
\begin{subfigure}{0.49\textwidth}
    \centering
    \setlength{\tabcolsep}{1pt}
    \begin{tabular}{cccc}
     \includegraphics[width=0.2\textwidth]{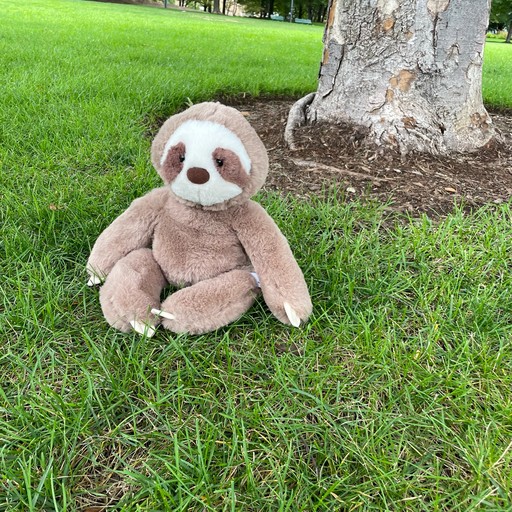} &
    \includegraphics[width=0.2\textwidth]{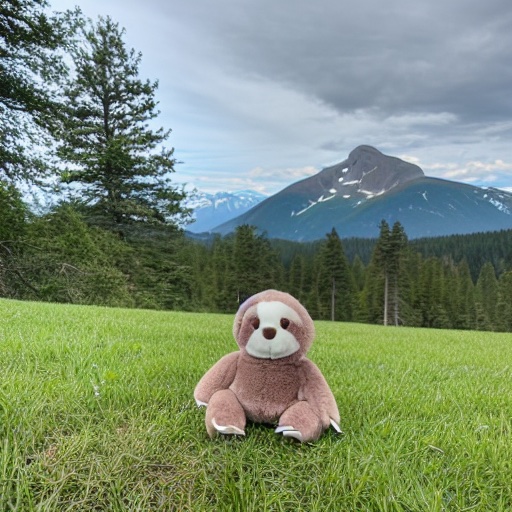} &
    \includegraphics[width=0.2\textwidth]{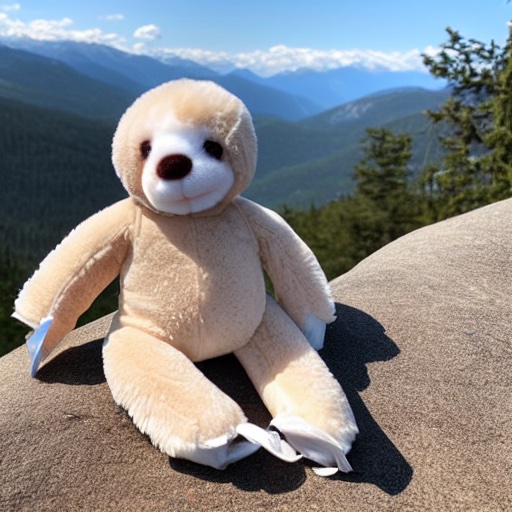} &
    \includegraphics[width=0.2\textwidth]{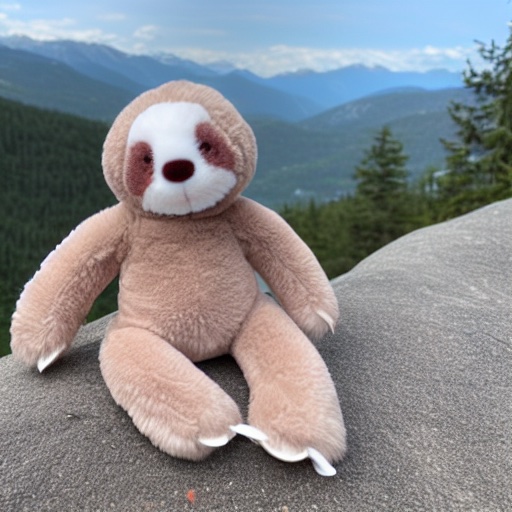} 
    \end{tabular}
    \vspace{-5pt} 
    \caption*{a \textit{stuffed animal$^*$} with a mountain in the background}
    \vspace{10pt} 
    \label{fig:sub4}
\end{subfigure}

\vspace{-5pt} 

\caption{Guided Image Synthesis: Across various subjects and prompts, our approach successfully preserves the layout of the reference underfit image as well as the identity of the input subject. Images generated by the Overfit (Edit) and Underfit (Guidance) models used in our approach are shown for reference.}
\label{fig:guided-image-synthesis-2}
\vspace{15pt}
\end{figure*}

\begin{figure*}
\centering

\begin{subfigure}{0.24\textwidth}
    \centering
    \setlength{\tabcolsep}{2pt}
     \caption*{Input}
    \begin{tabular}{cc}
    \includegraphics[width=0.6\textwidth]{images/cat_statue_input/2.jpeg} &
    \raisebox{0.32\textwidth} {\begin{tabular}[t]{@{}c@{}} 
    \includegraphics[width=0.28\textwidth]{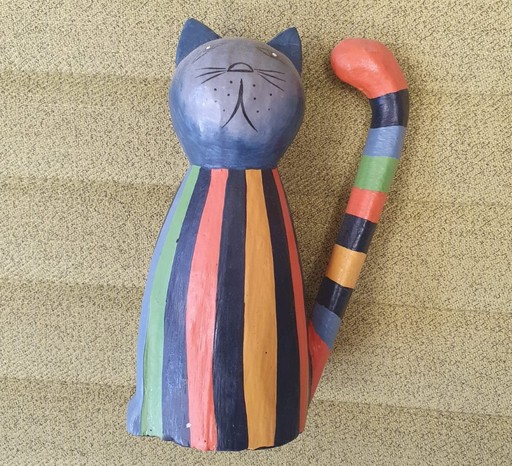} \\
    \includegraphics[width=0.28\textwidth]{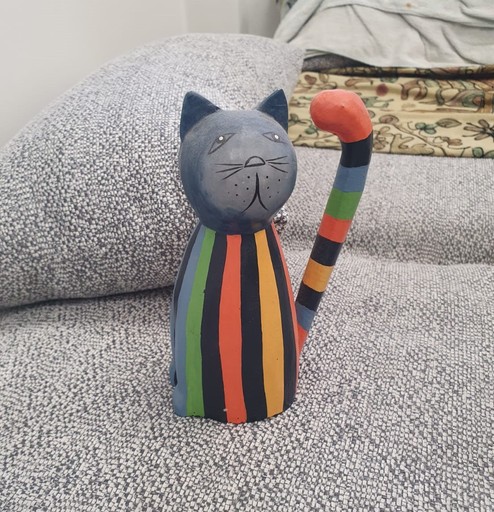}
    \end{tabular}}
    \end{tabular}
    \label{fig:sub4}
\end{subfigure}
\begin{subfigure}{0.24\textwidth}
    \centering
    \setlength{\tabcolsep}{2pt}
    \caption*{Ours}
    \begin{tabular}{cc}
    \includegraphics[width=0.6\textwidth]{images/comparisons/supplementary/c3/ours1.jpg} &
    \raisebox{0.32\textwidth} {\begin{tabular}[t]{@{}c@{}} 
    \includegraphics[width=0.28\textwidth]{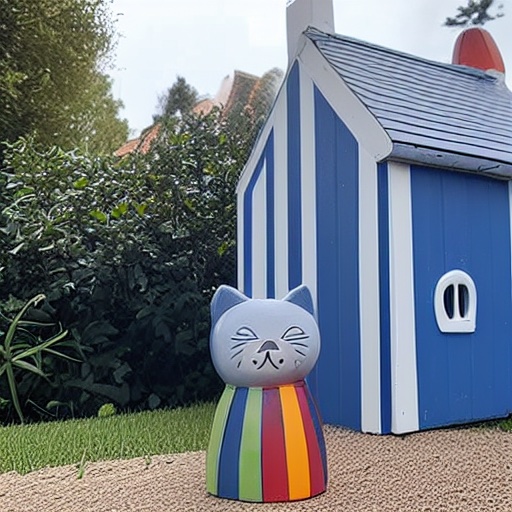} \\
    \includegraphics[width=0.28\textwidth]{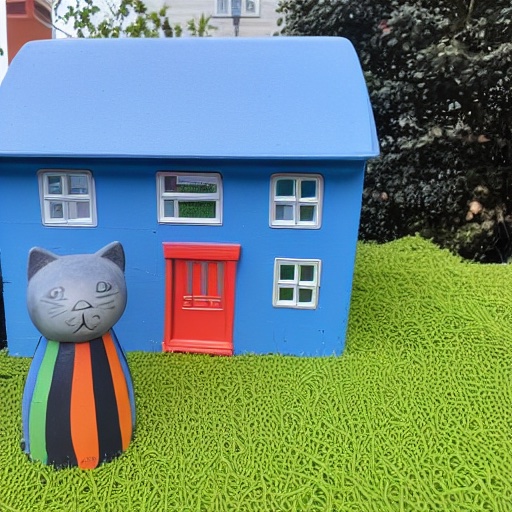}
    \end{tabular}}
    \end{tabular}
    \label{fig:sub4}
\end{subfigure}
\begin{subfigure}{0.24\textwidth}
    \centering
    \setlength{\tabcolsep}{2pt}
    \caption*{DreamBooth}
    \begin{tabular}{cc}
    \includegraphics[width=0.6\textwidth]{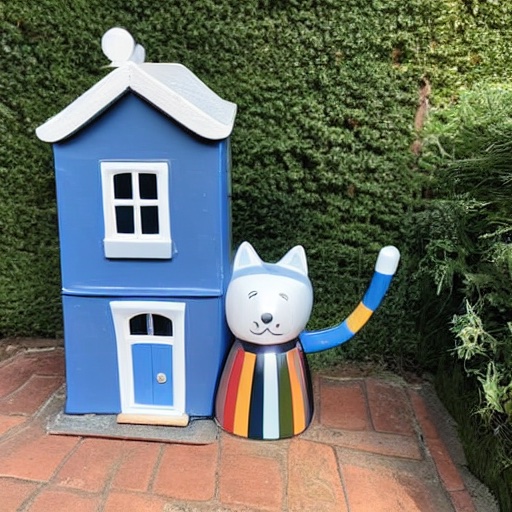} &
    \raisebox{0.32\textwidth} {\begin{tabular}[t]{@{}c@{}} 
    \includegraphics[width=0.28\textwidth]{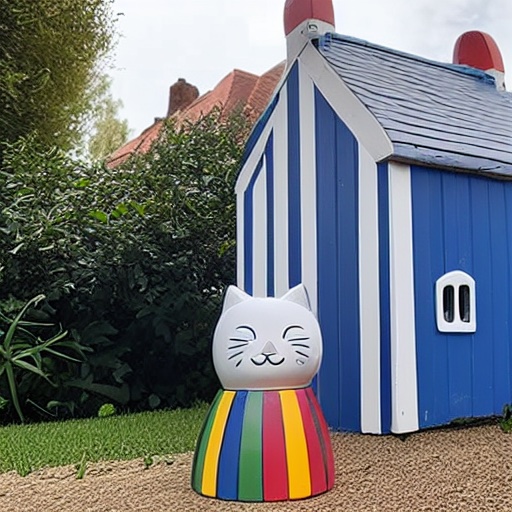} \\
    \includegraphics[width=0.28\textwidth]{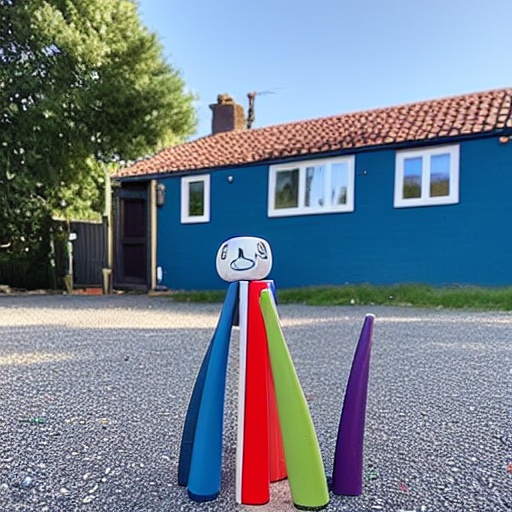}
    \end{tabular}}
    \end{tabular}
    \label{fig:sub4}
\end{subfigure}
\begin{subfigure}{0.24\textwidth}
    \centering
    \setlength{\tabcolsep}{2pt}
    \caption*{CustomDiffusion}
    \begin{tabular}{cc}
    \includegraphics[width=0.6\textwidth]{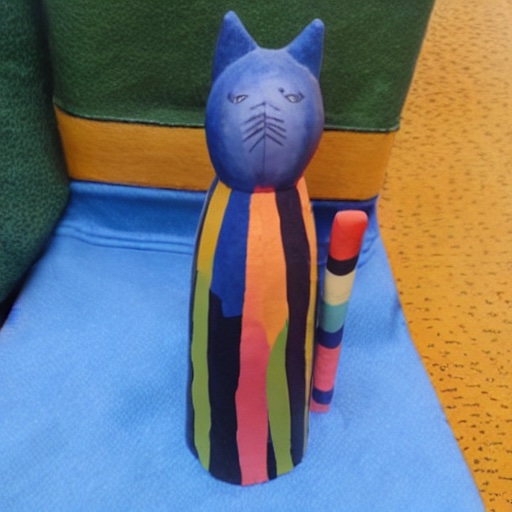} &
    \raisebox{0.32\textwidth} {\begin{tabular}[t]{@{}c@{}} 
    \includegraphics[width=0.28\textwidth]{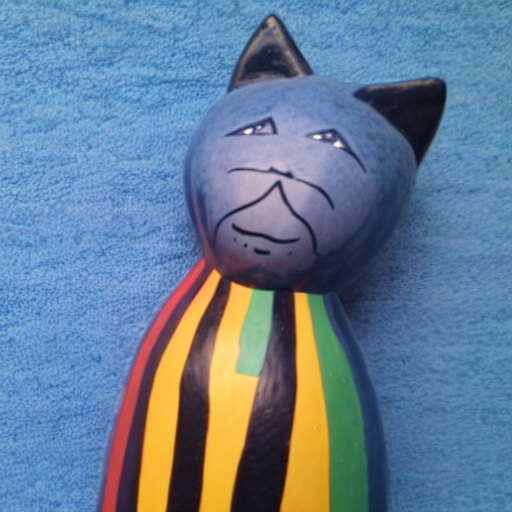} \\
    \includegraphics[width=0.28\textwidth]{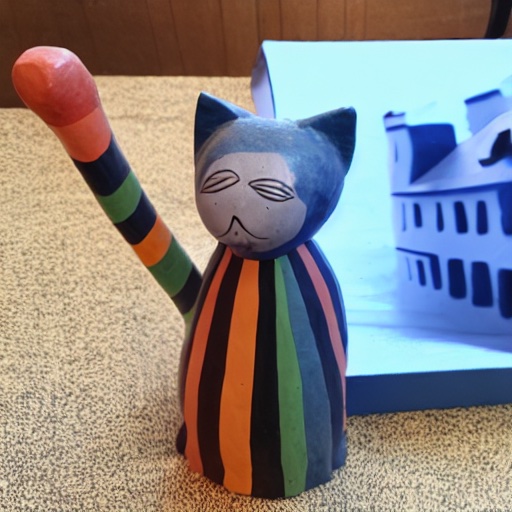}
    \end{tabular}}
    \end{tabular}
    \label{fig:sub4}
\end{subfigure}
\vspace{-12pt} 
\caption*{a $toy^*$ with a blue house in the background}

\vspace{12pt} 
\begin{subfigure}{0.24\textwidth}
    \centering
    \setlength{\tabcolsep}{2pt}
    \begin{tabular}{cc}
    \includegraphics[width=0.6\textwidth]{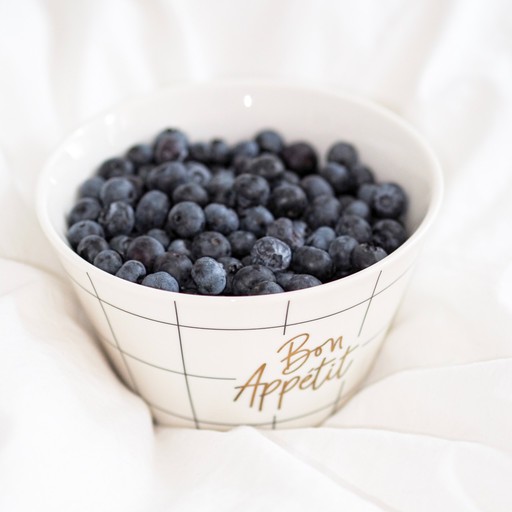} &
    \raisebox{0.32\textwidth} {\begin{tabular}[t]{@{}c@{}} 
    \includegraphics[width=0.28\textwidth]{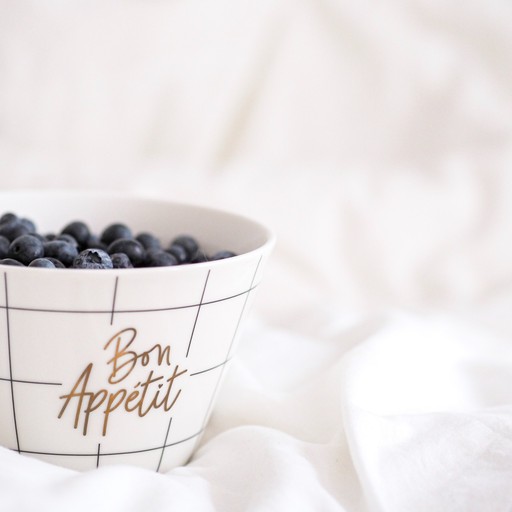} \\
    \includegraphics[width=0.28\textwidth]{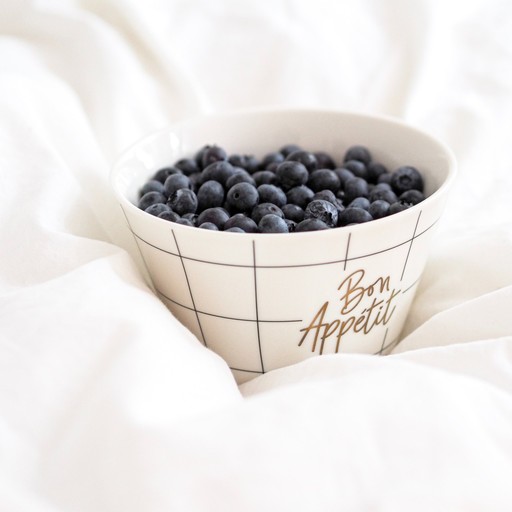}
    \end{tabular}}
    \end{tabular}
    \label{fig:sub4}
\end{subfigure}
\begin{subfigure}{0.24\textwidth}
    \centering
    \setlength{\tabcolsep}{2pt}
    \begin{tabular}{cc}
    \includegraphics[width=0.6\textwidth]{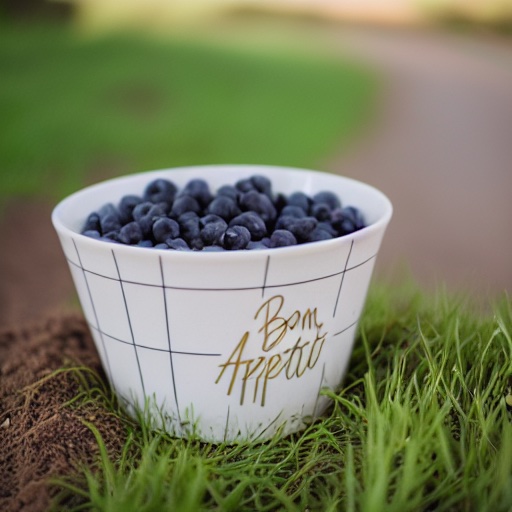} &
    \raisebox{0.32\textwidth} {\begin{tabular}[t]{@{}c@{}} 
    \includegraphics[width=0.28\textwidth]{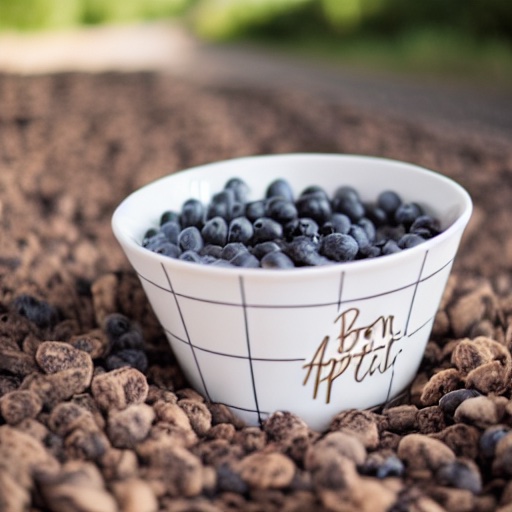} \\
    \includegraphics[width=0.28\textwidth]{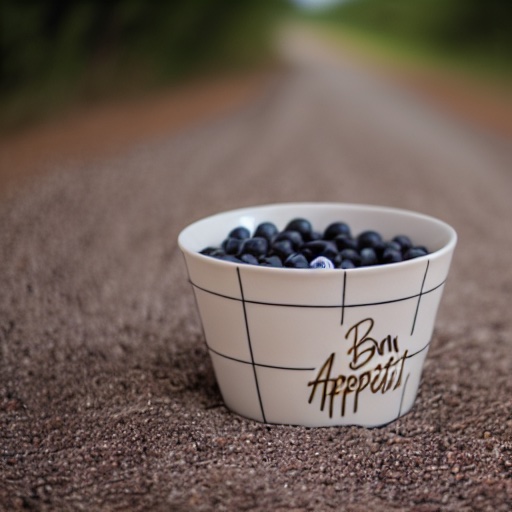}
    \end{tabular}}
    \end{tabular}
    \label{fig:sub4}
\end{subfigure}
\begin{subfigure}{0.24\textwidth}
    \centering
    \setlength{\tabcolsep}{2pt}
    \begin{tabular}{cc}
    \includegraphics[width=0.6\textwidth]{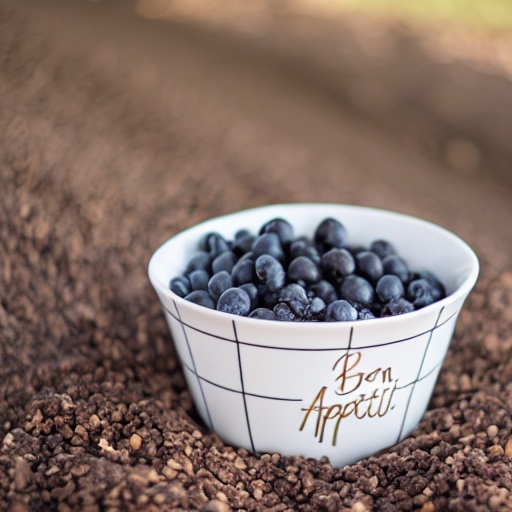} &
    \raisebox{0.32\textwidth} {\begin{tabular}[t]{@{}c@{}} 
    \includegraphics[width=0.28\textwidth]{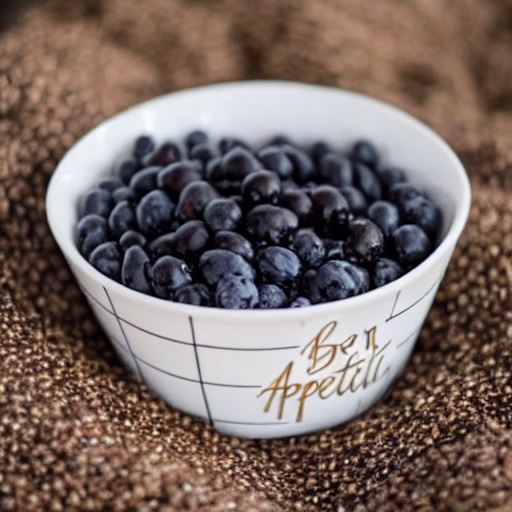} \\
    \includegraphics[width=0.28\textwidth]{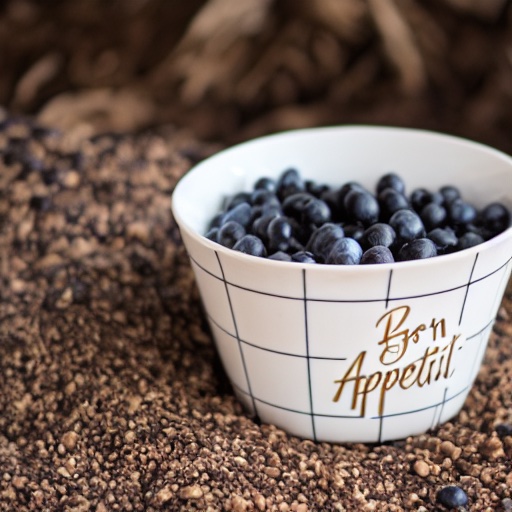}
    \end{tabular}}
    \end{tabular}
    \label{fig:sub4}
\end{subfigure}
\begin{subfigure}{0.24\textwidth}
    \centering
    \setlength{\tabcolsep}{2pt}
    \begin{tabular}{cc}
    \includegraphics[width=0.6\textwidth]{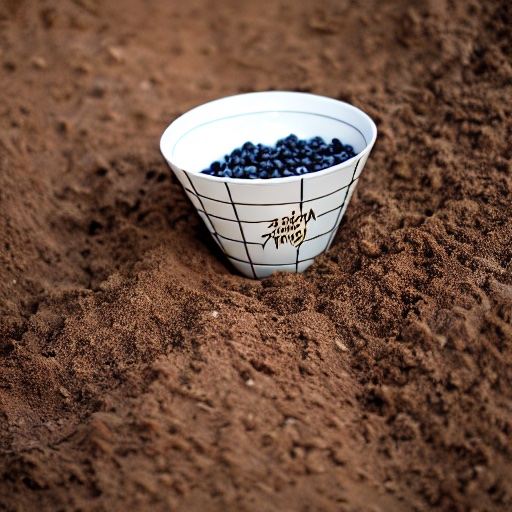} &
    \raisebox{0.32\textwidth} {\begin{tabular}[t]{@{}c@{}} 
    \includegraphics[width=0.28\textwidth]{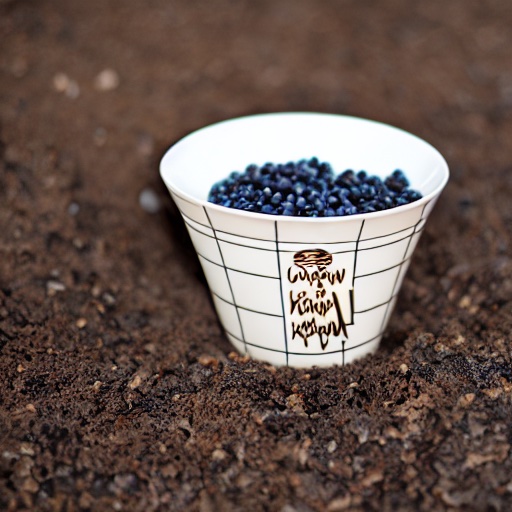} \\
    \includegraphics[width=0.28\textwidth]{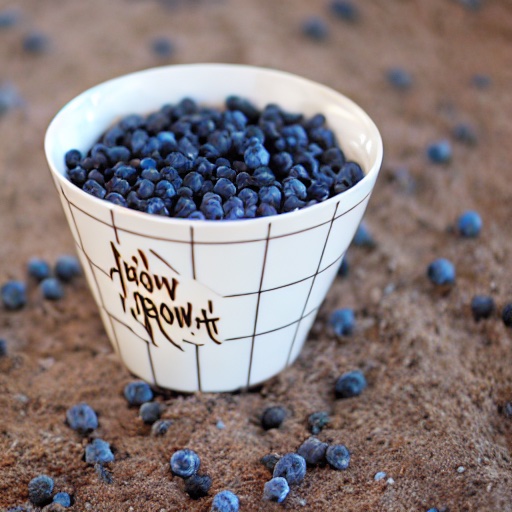}
    \end{tabular}}
    \end{tabular}
    \label{fig:sub4}
\end{subfigure}
\vspace{-12pt} 
\caption*{a $bowl^*$ on top of a dirt road}

\vspace{12pt} 
\begin{subfigure}{0.24\textwidth}
    \centering
    \setlength{\tabcolsep}{2pt}
    \begin{tabular}{cc}
    \includegraphics[width=0.6\textwidth]{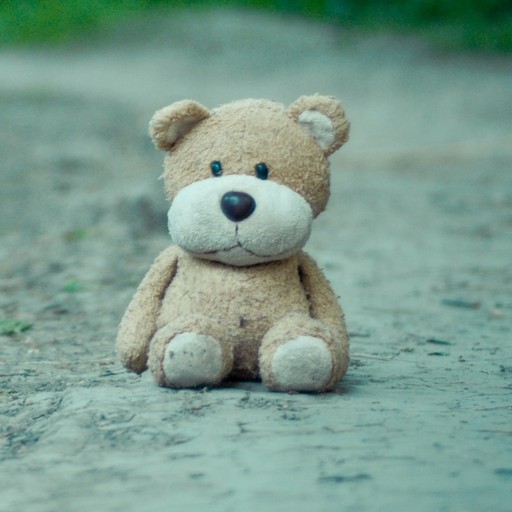} &
    \raisebox{0.32\textwidth} {\begin{tabular}[t]{@{}c@{}} 
    \includegraphics[width=0.28\textwidth]{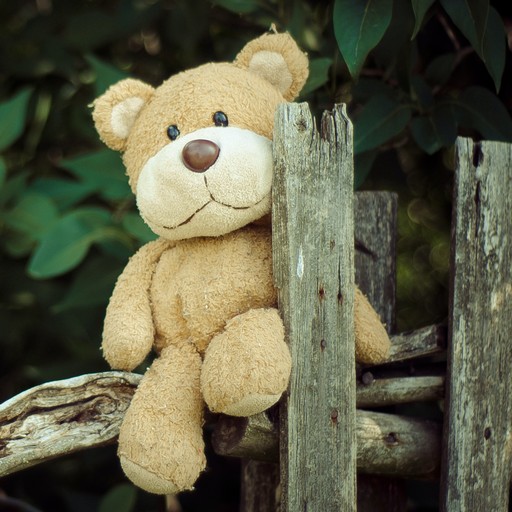} \\
    \includegraphics[width=0.28\textwidth]{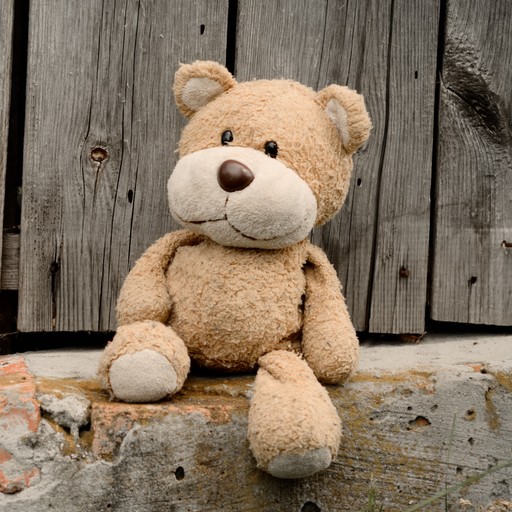}
    \end{tabular}}
    \end{tabular}
    \label{fig:sub4}
\end{subfigure}
\begin{subfigure}{0.24\textwidth}
    \centering
    \setlength{\tabcolsep}{2pt}
    \begin{tabular}{cc}
    \includegraphics[width=0.6\textwidth]{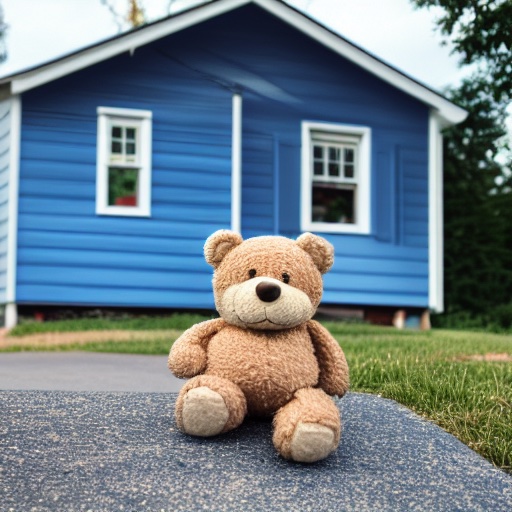} &
    \raisebox{0.32\textwidth} {\begin{tabular}[t]{@{}c@{}} 
    \includegraphics[width=0.28\textwidth]{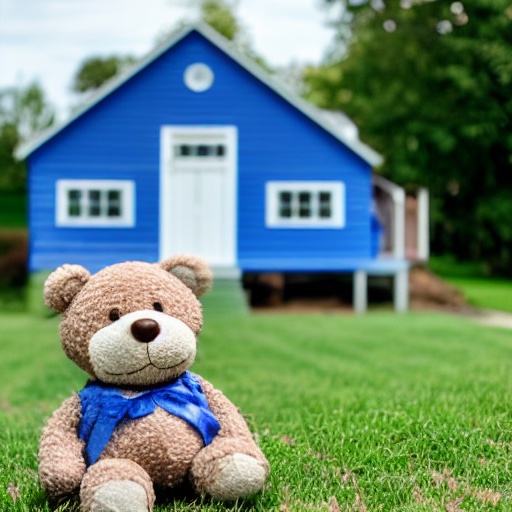} \\
    \includegraphics[width=0.28\textwidth]{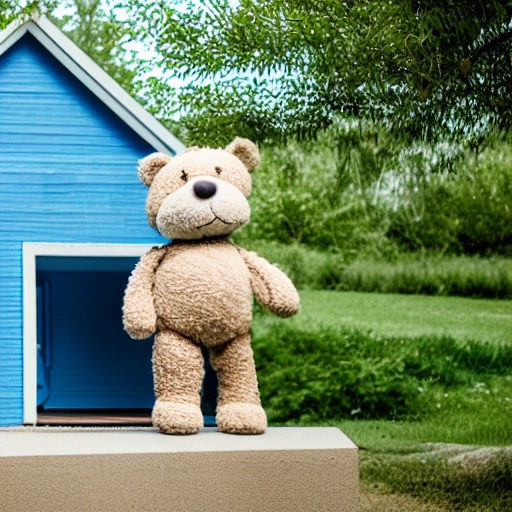}
    \end{tabular}}
    \end{tabular}
    \label{fig:sub4}
\end{subfigure}
\begin{subfigure}{0.24\textwidth}
    \centering
    \setlength{\tabcolsep}{2pt}
    \begin{tabular}{cc}
    \includegraphics[width=0.6\textwidth]{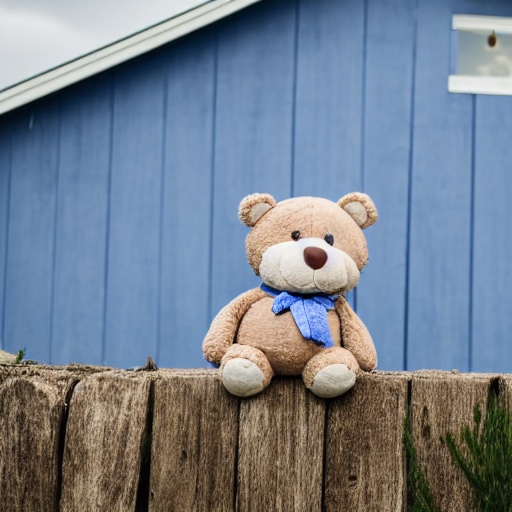} &
    \raisebox{0.32\textwidth} {\begin{tabular}[t]{@{}c@{}} 
    \includegraphics[width=0.28\textwidth]{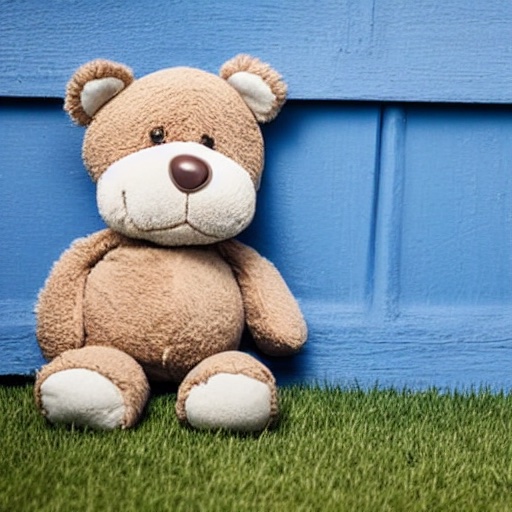} \\
    \includegraphics[width=0.28\textwidth]{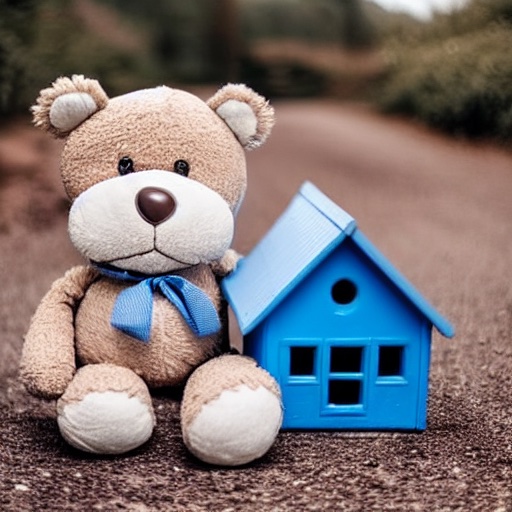}
    \end{tabular}}
    \end{tabular}
    \label{fig:sub4}
\end{subfigure}
\begin{subfigure}{0.24\textwidth}
    \centering
    \setlength{\tabcolsep}{2pt}
    \begin{tabular}{cc}
    \includegraphics[width=0.6\textwidth]{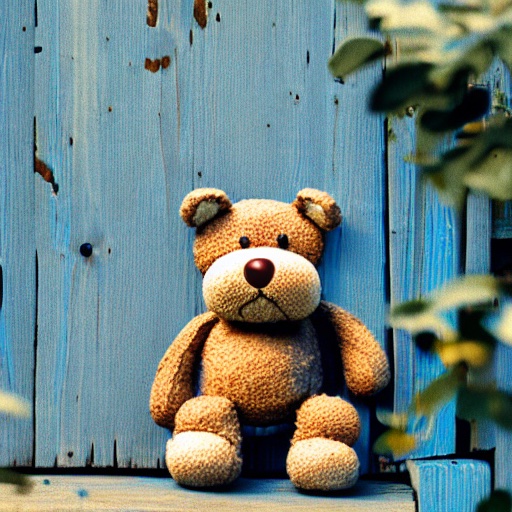} &
    \raisebox{0.32\textwidth} {\begin{tabular}[t]{@{}c@{}} 
    \includegraphics[width=0.28\textwidth]{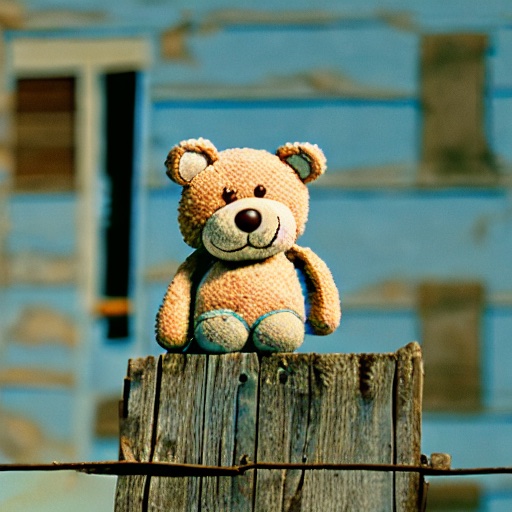} \\
    \includegraphics[width=0.28\textwidth]{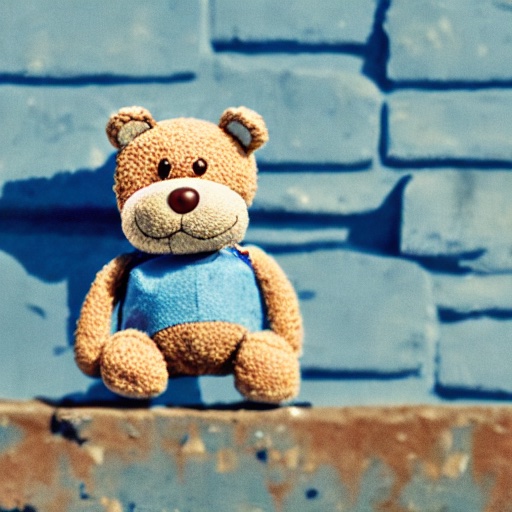}
    \end{tabular}}
    \end{tabular}
    \label{fig:sub4}
\end{subfigure}
\vspace{-12pt} 
\caption*{a $teddy^*$ with a blue house in the background}

\vspace{12pt} 
\begin{subfigure}{0.24\textwidth}
    \centering
    \setlength{\tabcolsep}{2pt}
    \begin{tabular}{cc}
    \includegraphics[width=0.6\textwidth]{images/backpack_input/02.jpg} &
    \raisebox{0.32\textwidth} {\begin{tabular}[t]{@{}c@{}} 
    \includegraphics[width=0.28\textwidth]{images/backpack_input/03.jpg} \\
    \includegraphics[width=0.28\textwidth]{images/backpack_input/05.jpg}
    \end{tabular}}
    \end{tabular}
    \label{fig:sub4}
\end{subfigure}
\begin{subfigure}{0.24\textwidth}
    \centering
    \setlength{\tabcolsep}{2pt}
    \begin{tabular}{cc}
    \includegraphics[width=0.6\textwidth]{images/comparisons/supplementary/c2/ours1.jpg} &
    \raisebox{0.32\textwidth} {\begin{tabular}[t]{@{}c@{}} 
    \includegraphics[width=0.28\textwidth]{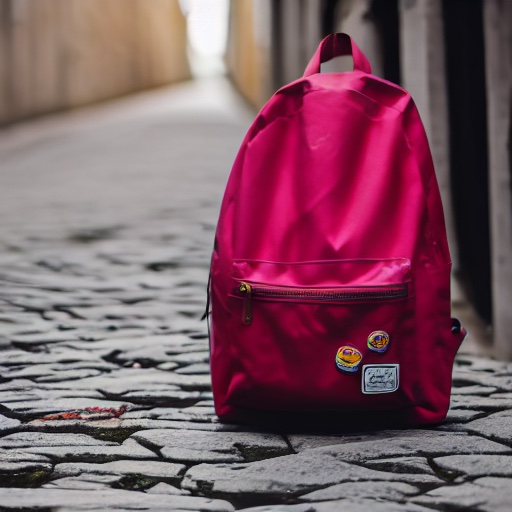} \\
    \includegraphics[width=0.28\textwidth]{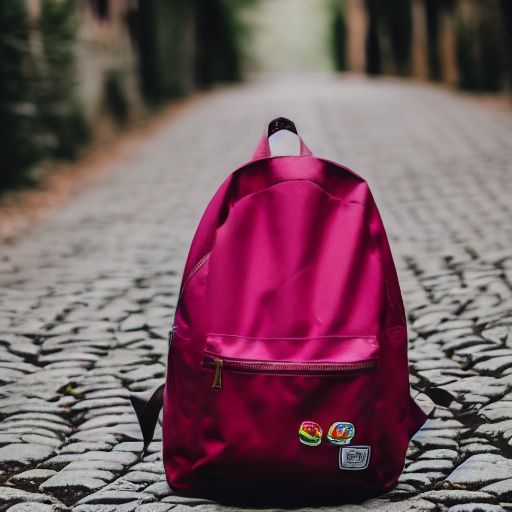}
    \end{tabular}}
    \end{tabular}
    \label{fig:sub4}
\end{subfigure}
\begin{subfigure}{0.24\textwidth}
    \centering
    \setlength{\tabcolsep}{2pt}
    \begin{tabular}{cc}
    \includegraphics[width=0.6\textwidth]{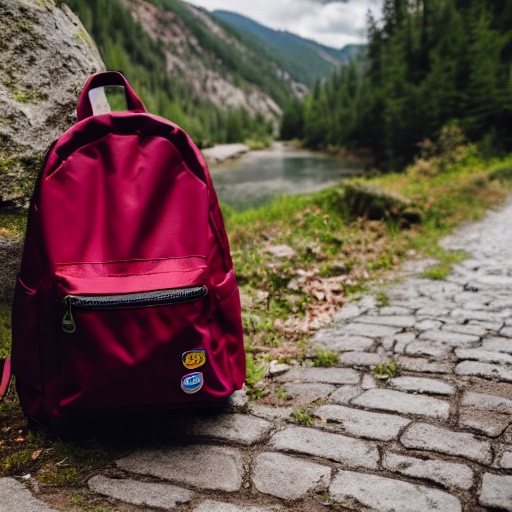} &
    \raisebox{0.32\textwidth} {\begin{tabular}[t]{@{}c@{}} 
    \includegraphics[width=0.28\textwidth]{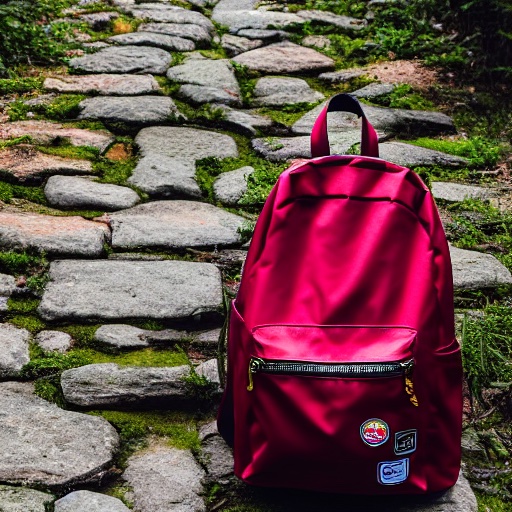} \\
    \includegraphics[width=0.28\textwidth]{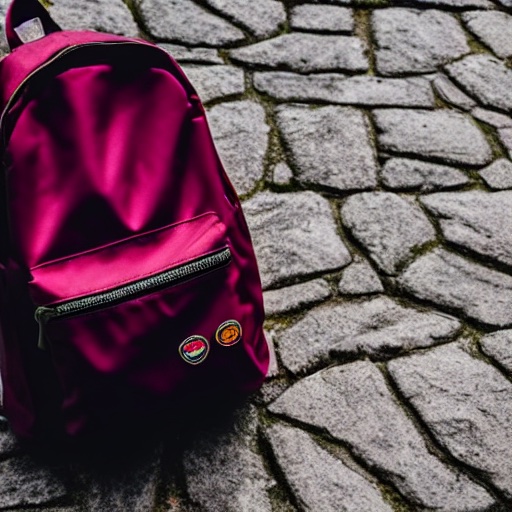}
    \end{tabular}}
    \end{tabular}
    \label{fig:sub4}
\end{subfigure}
\begin{subfigure}{0.24\textwidth}
    \centering
    \setlength{\tabcolsep}{2pt}
    \begin{tabular}{cc}
    \includegraphics[width=0.6\textwidth]{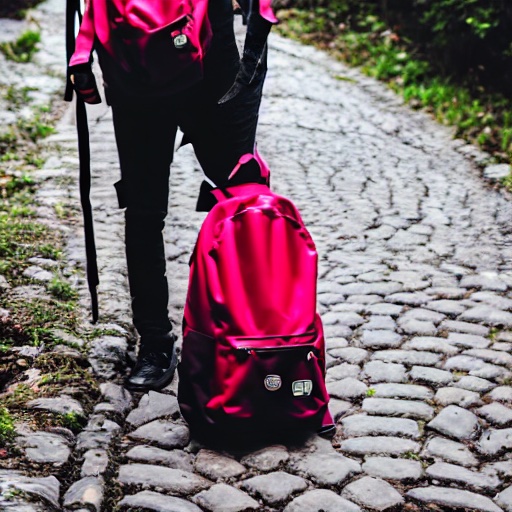} &
    \raisebox{0.32\textwidth} {\begin{tabular}[t]{@{}c@{}} 
    \includegraphics[width=0.28\textwidth]{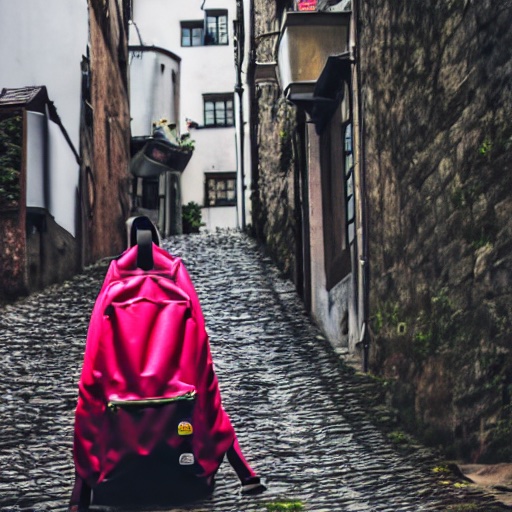} \\
    \includegraphics[width=0.28\textwidth]{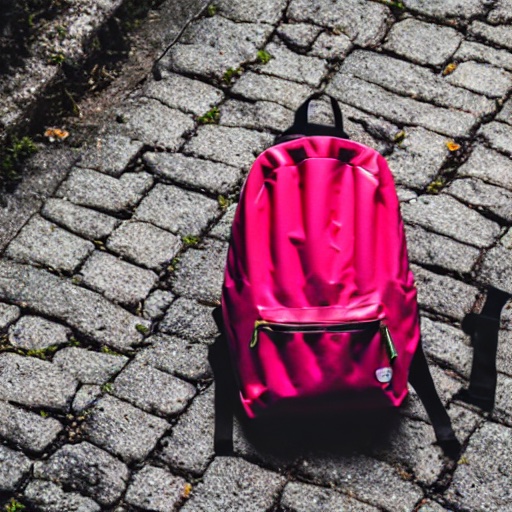}
    \end{tabular}}
    \end{tabular}
    \label{fig:sub4}
\end{subfigure}
\vspace{-12pt} 
\caption*{a $backpack^*$ on a cobblestone street}

\vspace{12pt} 
\begin{subfigure}{0.24\textwidth}
    \centering
    \setlength{\tabcolsep}{2pt}
    \begin{tabular}{cc}
    \includegraphics[width=0.6\textwidth]{images/candle_input/03.jpg} &
    \raisebox{0.32\textwidth} {\begin{tabular}[t]{@{}c@{}} 
    \includegraphics[width=0.28\textwidth]{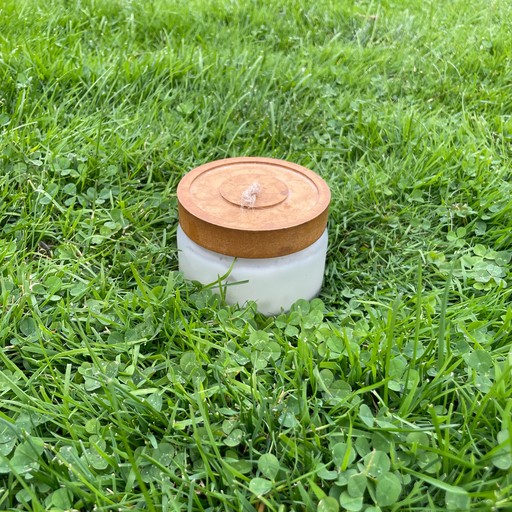} \\
    \includegraphics[width=0.28\textwidth]{images/candle_input/04.jpg}
    \end{tabular}}
    \end{tabular}
    \label{fig:sub4}
\end{subfigure}
\begin{subfigure}{0.24\textwidth}
    \centering
    \setlength{\tabcolsep}{2pt}
    \begin{tabular}{cc}
    \includegraphics[width=0.6\textwidth]{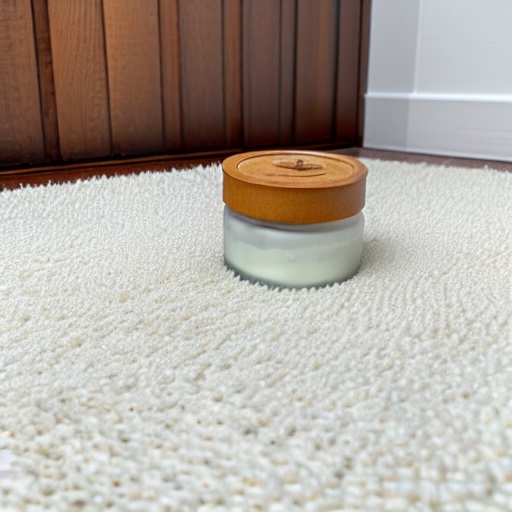} &
    \raisebox{0.32\textwidth} {\begin{tabular}[t]{@{}c@{}} 
    \includegraphics[width=0.28\textwidth]{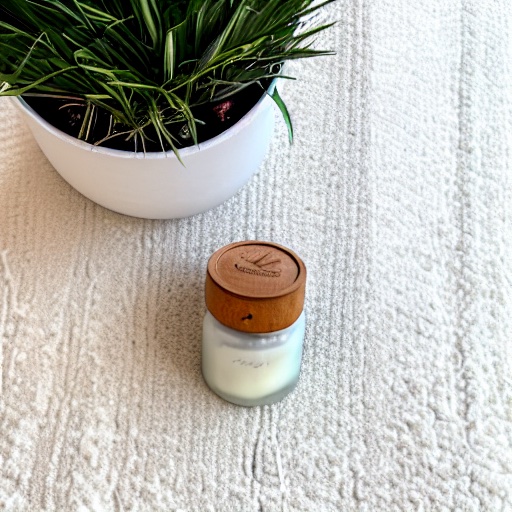} \\
    \includegraphics[width=0.28\textwidth]{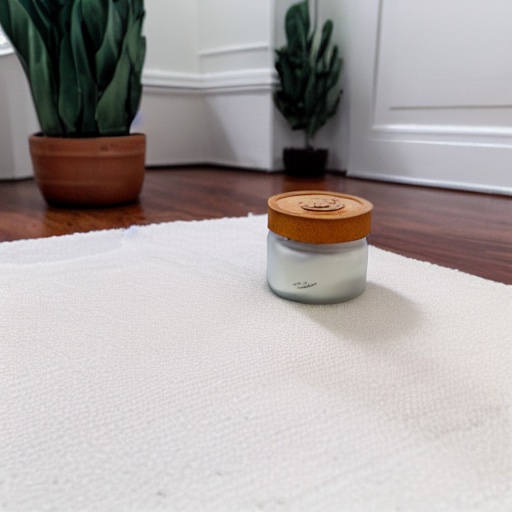}
    \end{tabular}}
    \end{tabular}
    \label{fig:sub4}
\end{subfigure}
\begin{subfigure}{0.24\textwidth}
    \centering
    \setlength{\tabcolsep}{2pt}
    \begin{tabular}{cc}
    \includegraphics[width=0.6\textwidth]{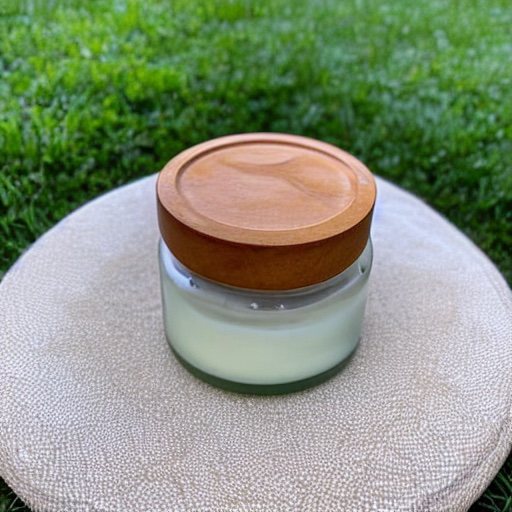} &
    \raisebox{0.32\textwidth} {\begin{tabular}[t]{@{}c@{}} 
    \includegraphics[width=0.28\textwidth]{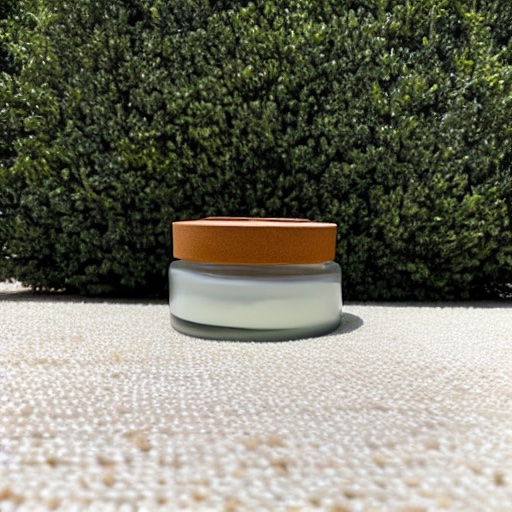} \\
    \includegraphics[width=0.28\textwidth]{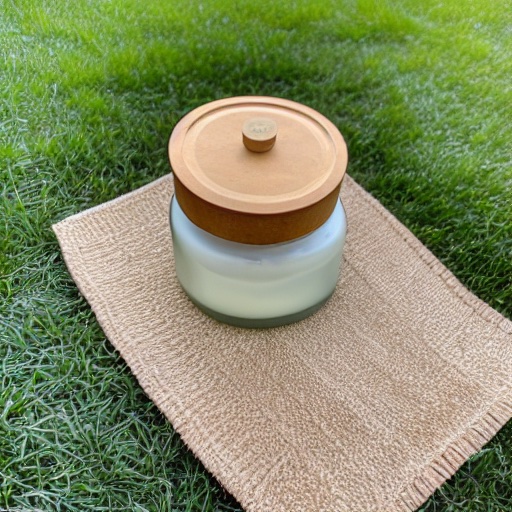}
    \end{tabular}}
    \end{tabular}
    \label{fig:sub4}
\end{subfigure}
\begin{subfigure}{0.24\textwidth}
    \centering
    \setlength{\tabcolsep}{2pt}
    \begin{tabular}{cc}
    \includegraphics[width=0.6\textwidth]{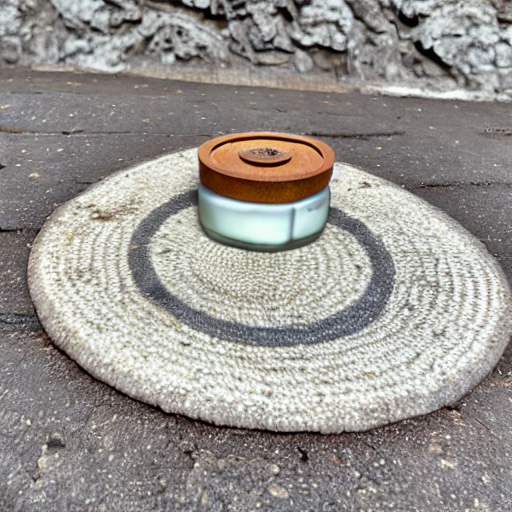} &
    \raisebox{0.32\textwidth} {\begin{tabular}[t]{@{}c@{}} 
    \includegraphics[width=0.28\textwidth]{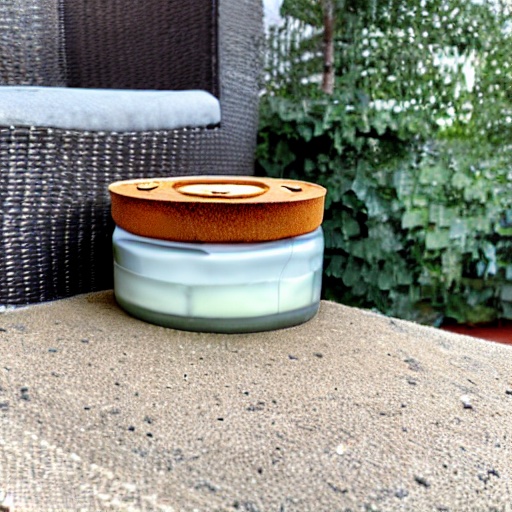} \\
    \includegraphics[width=0.28\textwidth]{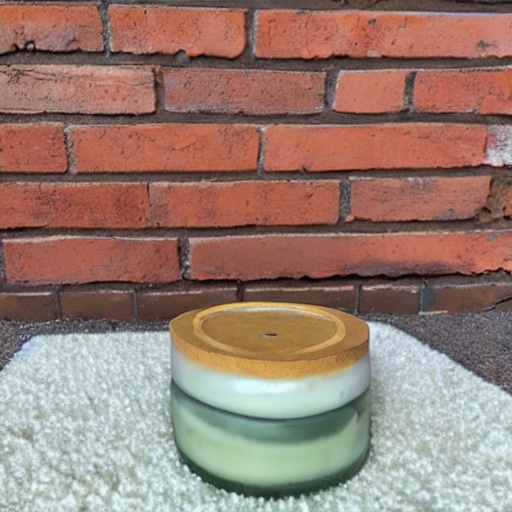}
    \end{tabular}}
    \end{tabular}
    \label{fig:sub4}
\end{subfigure}
\vspace{-12pt} 
\caption*{a $candle^*$ on top of a white rug}

\caption{Comparison with prior works: Our approach successfully generates images with better subject fidelity, prompt fidelity and diversity on challenging prompts.}
\label{fig:qualitative-comparison-2}
\end{figure*}

\begin{figure*}
\centering

\begin{subfigure}{0.49\textwidth}
    \centering
    \setlength{\tabcolsep}{1pt}
    \begin{tabular}{cccc}
    {Input} & {Overfit} & {Underfit} & {Ours} \\
     \includegraphics[width=0.25\textwidth]{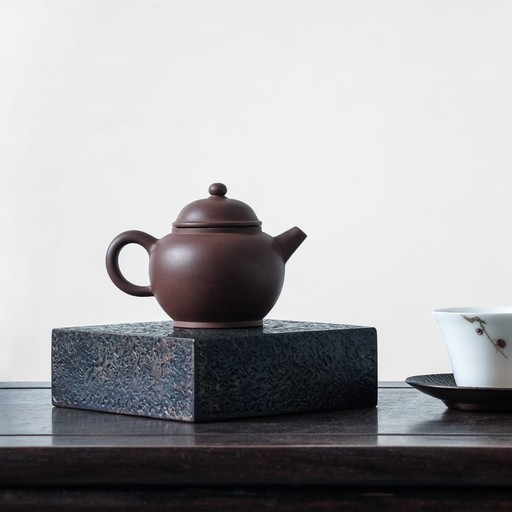} &
    \includegraphics[width=0.25\textwidth]{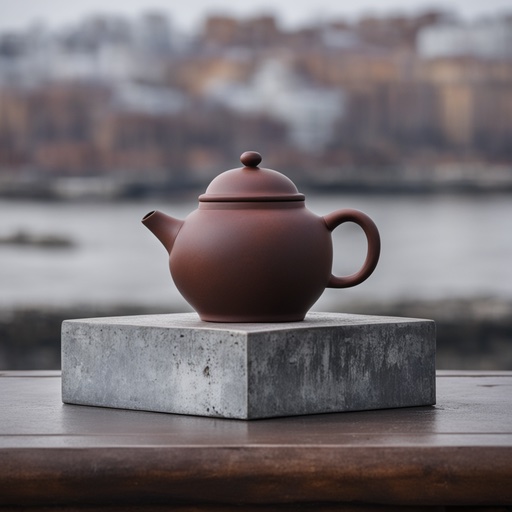} &
    \includegraphics[width=0.25\textwidth]{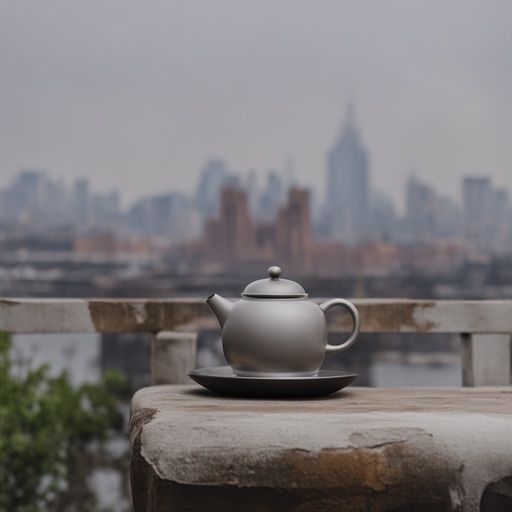} &
    \includegraphics[width=0.25\textwidth]{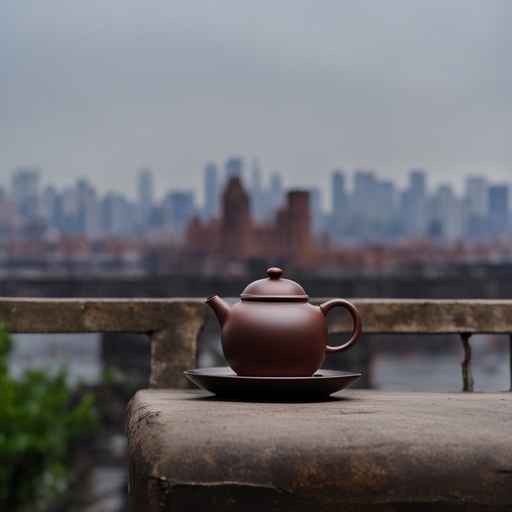} 
    \end{tabular}
    \vspace{-5pt} 
    \caption*{a $teapot^*$ with a city in the background}
    \label{fig:sub4}
\end{subfigure}


\begin{subfigure}{0.49\textwidth}
    \centering
    \setlength{\tabcolsep}{1pt}
    \begin{tabular}{cccc}
     \includegraphics[width=0.25\textwidth]{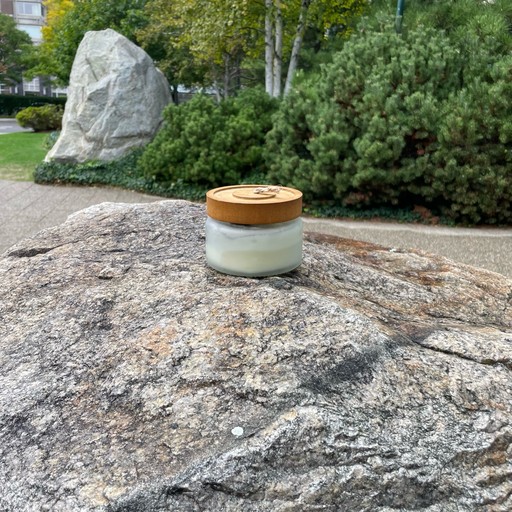} &
    \includegraphics[width=0.25\textwidth]{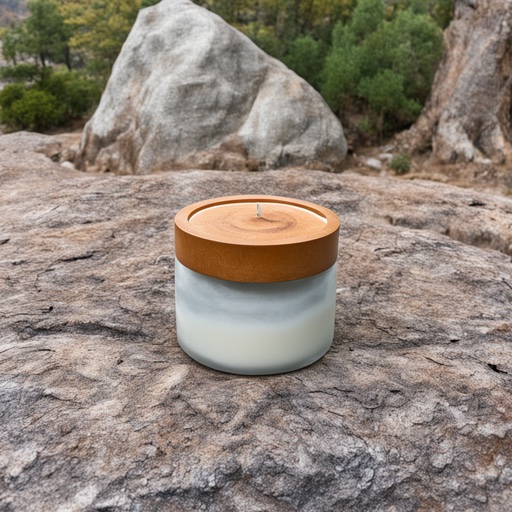} &
    \includegraphics[width=0.25\textwidth]{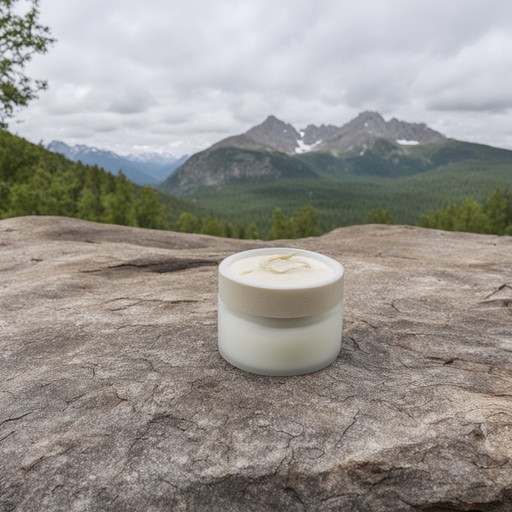} &
    \includegraphics[width=0.25\textwidth]{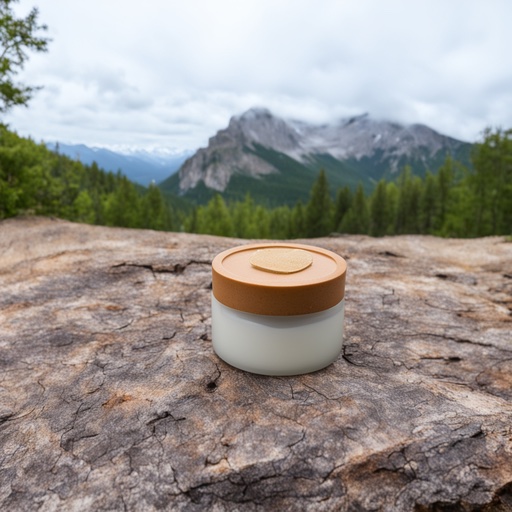} 
    \end{tabular}
    \vspace{-5pt} 
    \caption*{a $candle^*$ with a mountain in the background}
    \label{fig:sub4}
\end{subfigure}

\caption{Qualitative results on SDXL backbone}
\label{fig:sdxl-guided}

\vspace{-15pt}
\end{figure*}

In \cref{fig:guided-image-synthesis-2}, we present more results, in addition to \cref{fig:guided-image-synthesis} of the main text. In \cref{fig:qualitative-comparison-2}, we present more results in addition to \cref{fig:qualitative-comparison} of the main text. In \cref{fig:sdxl-guided}, we present more results with SDXL backbone, in addition to \cref{fig:other-backbones} of the main text.

\section{DreamBlend advances the pareto front}
\label{sec:more-quantitative-results}
\begin{figure*}
    \centering
    \begin{subfigure}{0.23\textwidth}
        \centering
        \includegraphics[width=\linewidth]{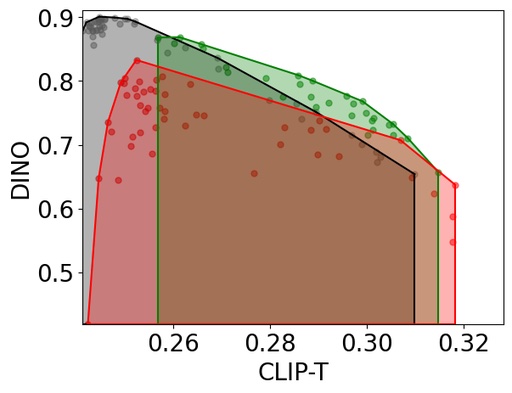}
        \caption{cat}
    \end{subfigure}
    \begin{subfigure}{0.23\textwidth}
        \centering
        \includegraphics[width=\linewidth]{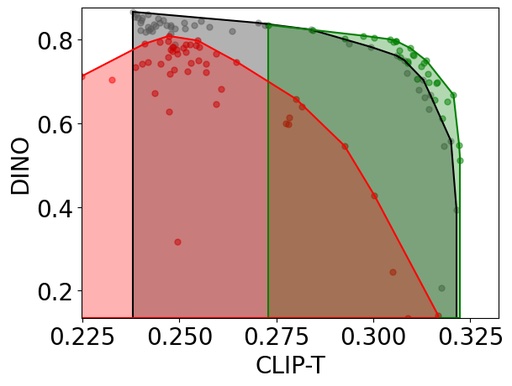}
        \caption{dog2}
    \end{subfigure}
    \begin{subfigure}{0.23\textwidth}
        \centering
        \includegraphics[width=\linewidth]{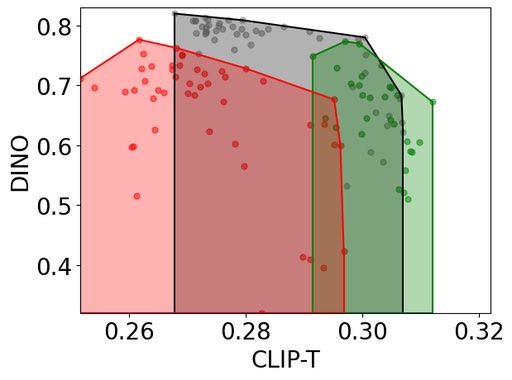}
        \caption{colorful sneaker}
    \end{subfigure}
    \begin{subfigure}{0.23\textwidth}
        \centering
        \includegraphics[width=\linewidth]{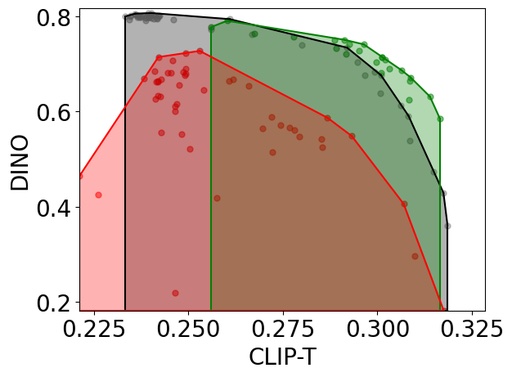}
        \caption{dog8}
    \end{subfigure}
    \begin{subfigure}{0.23\textwidth}
        \centering
        \includegraphics[width=\linewidth]{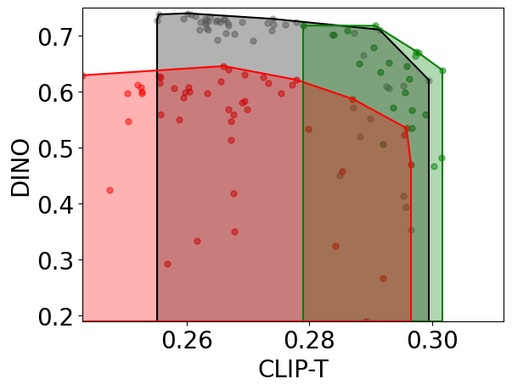}
        \caption{fancy boot}
    \end{subfigure}
    \begin{subfigure}{0.23\textwidth}
        \centering
        \includegraphics[width=\linewidth]{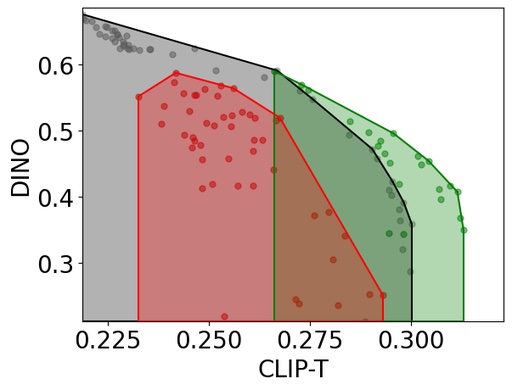}
        \caption{monster toy}
    \end{subfigure}
    \begin{subfigure}{0.23\textwidth}
        \centering
        \includegraphics[width=\linewidth]{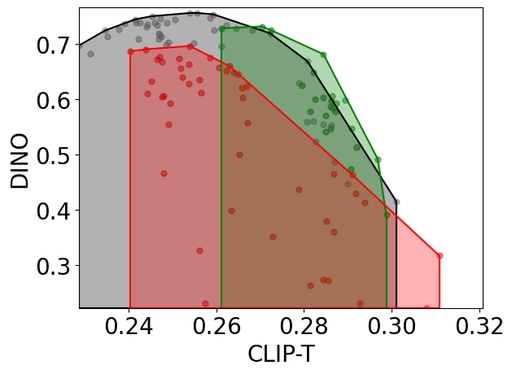}
        \caption{pink sunglasses}
    \end{subfigure}
    \begin{subfigure}{0.23\textwidth}
        \centering
        \includegraphics[width=\linewidth]{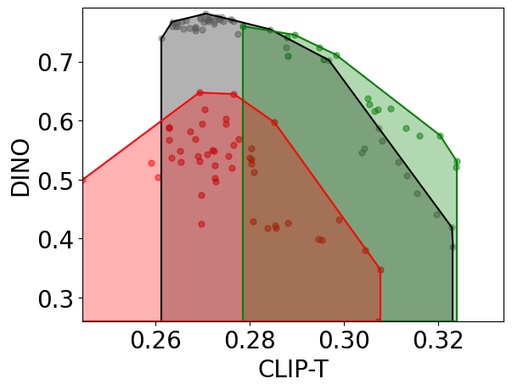}
        \caption{wolf plushie}
    \end{subfigure}
    \caption{Image alignment (DINO) - text alignment (CLIP-T) space spanned by densely sampled operating points of DreamBooth (gray), Custom Diffusion (red) and our method (green) for example subjects. Our method advances the pareto front, offering operating points unavailable to existing methods.}
    \label{fig:pareto-2}
\end{figure*}

In \cref{fig:pareto-2}, we visualize how DreamBlend advances the pareto front for more example subjects from the DreamBooth benchmark, in addition to \cref{fig:pareto} of the main text.

\section{Human preference study}
\label{sec:human-preference-study}
\begin{figure*}
\centering

\begin{subfigure}{0.49\textwidth}
    \centering
    \includegraphics[width=1.0\textwidth]{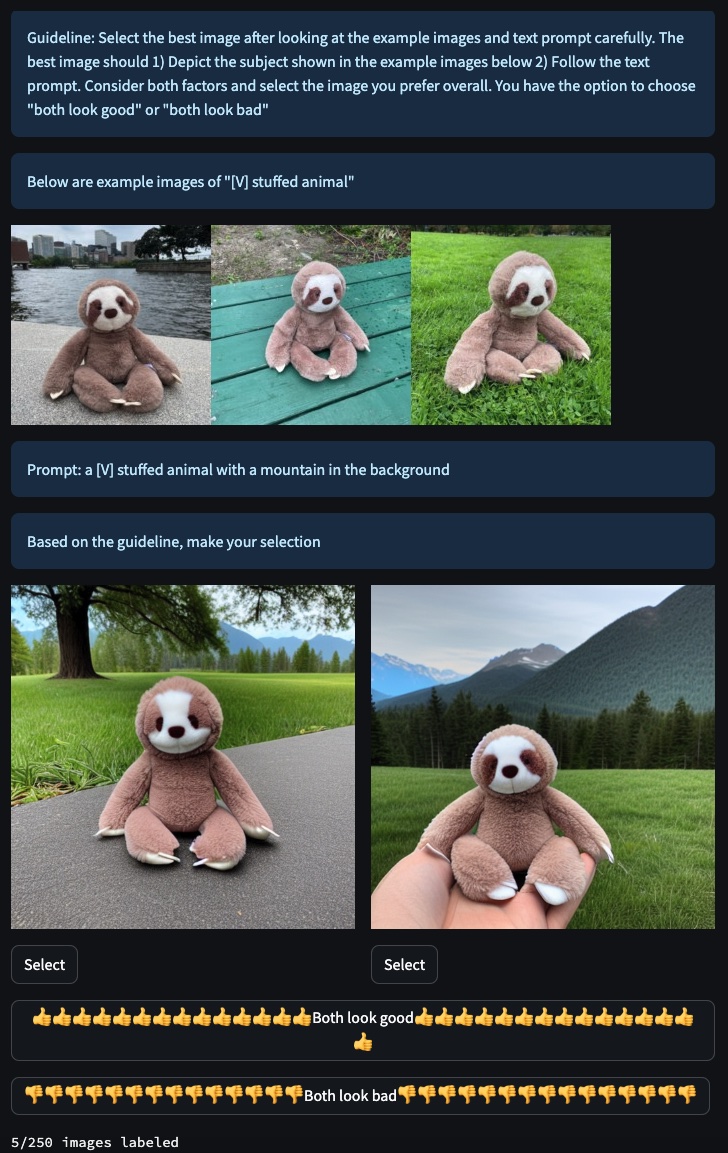}
   \caption{Overall preference study}
    \vspace{10pt} 
    \label{fig:user-study-overall}
\end{subfigure}
\begin{subfigure}{0.49\textwidth}
    \centering
    \includegraphics[width=1.0\textwidth]{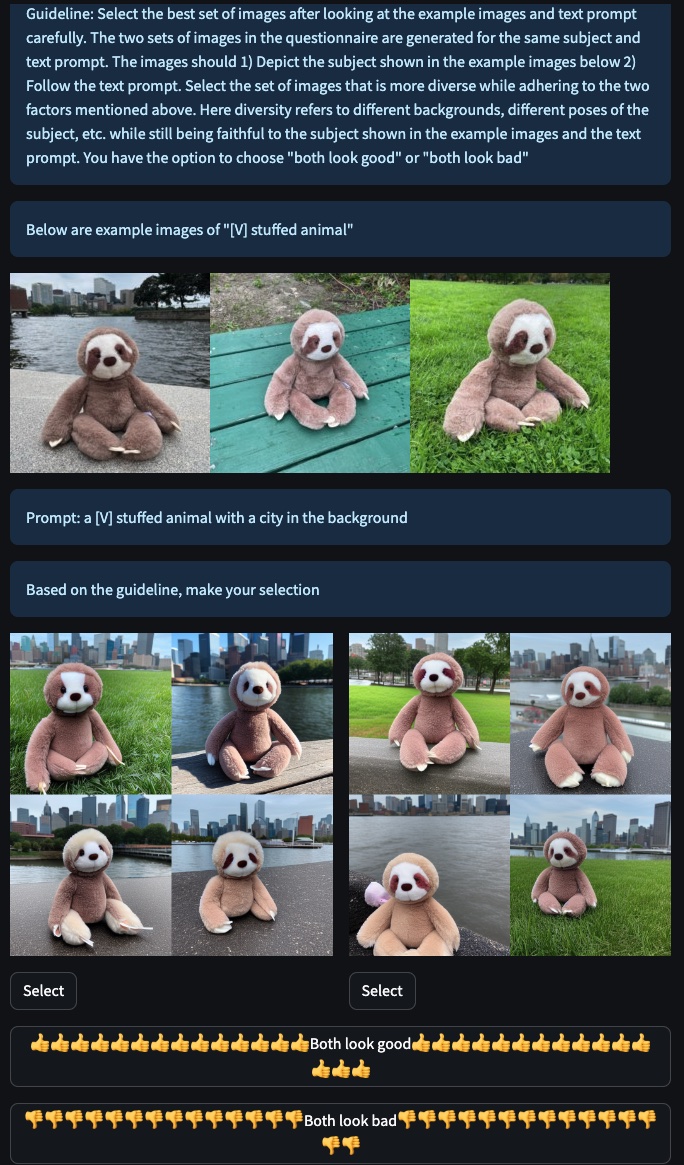}
   \caption{Diversity study}
    \vspace{10pt} 
    \label{fig:user-study-diversity}
\end{subfigure}
\caption{Human preference study interface}
\label{fig:user-study}
\vspace{50pt}
\end{figure*}

Two user studies were performed, assessing overall preference and diversity, comparing our approach to DreamBooth and Custom Diffusion. An example interface used for these studies is shown in \cref{fig:user-study}. In the overall preference study shown in \cref{fig:user-study-overall}, users chose between an image generated by our method and a baseline method for the same text prompt, considering both subject and prompt fidelity. In the diversity study shown in \cref{fig:user-study-diversity}, users selected the more diverse collection of four images between our method and a baseline. They were asked to consider both subject fidelity and prompt fidelity and select the collection of images which is more diverse in terms of backgrounds, subject poses, etc. For example, in \cref{fig:user-study-diversity}, the images in the left collection have very similar backgrounds while the images in the right collection are more diverse. The studies comprised of 1000 questions, each question was answered by an average of six people and the order was randomized.

Statistical tests were performed to verify the statistical significance of each study and all results were found to be statistically significant. The results of one-sample binomial test with confidence intervals are summarized in \cref{tab:binomial}. The results of Chi-square goodness of fit test are summarized in \cref{tab:chi-square}.

\section{Effect of varying cross attention guidance and classifier-free guidance}
\label{sec:gs_ags}
\begin{figure*}[h]
\centering
     \begin{subfigure}{0.6\textwidth}
    \centering
    \setlength{\tabcolsep}{2pt}
    \begin{tabular}{cccccc}
   Input & Overfit & Underfit & $\alpha=0.0$ & $\alpha=0.1$ & $\alpha=0.2$ \\
    \includegraphics[width=0.16\textwidth]{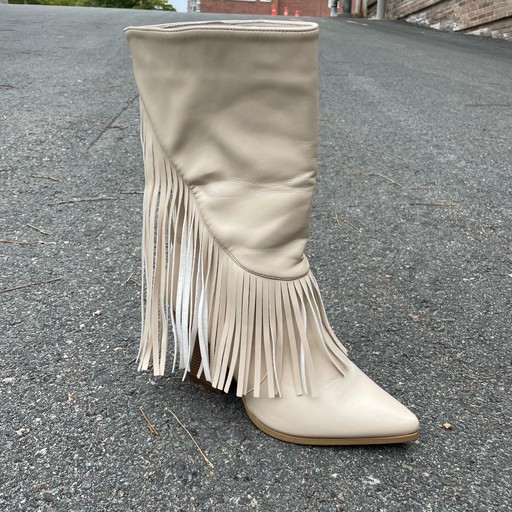} &
     \includegraphics[width=0.16\textwidth]{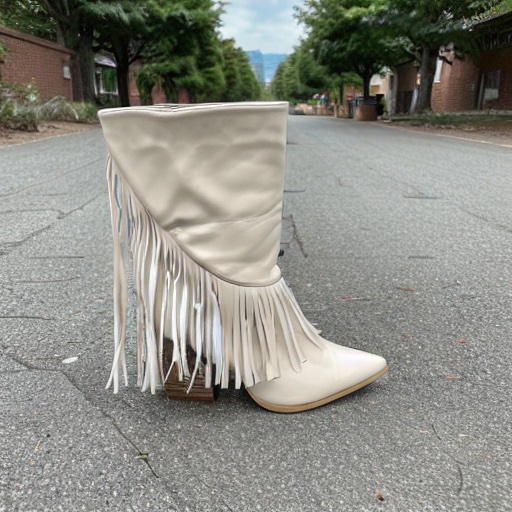} &
      \includegraphics[width=0.16\textwidth]{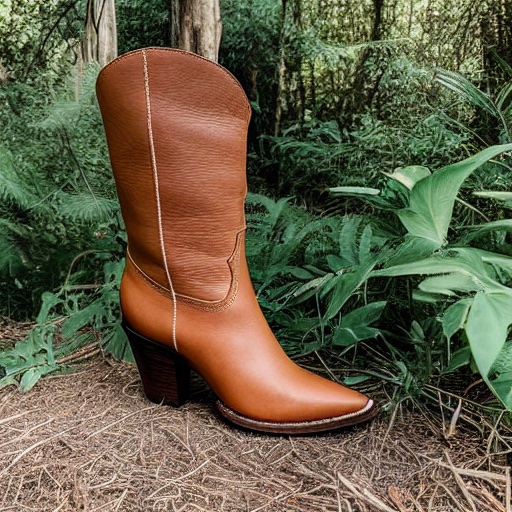} &
       \includegraphics[width=0.16\textwidth]{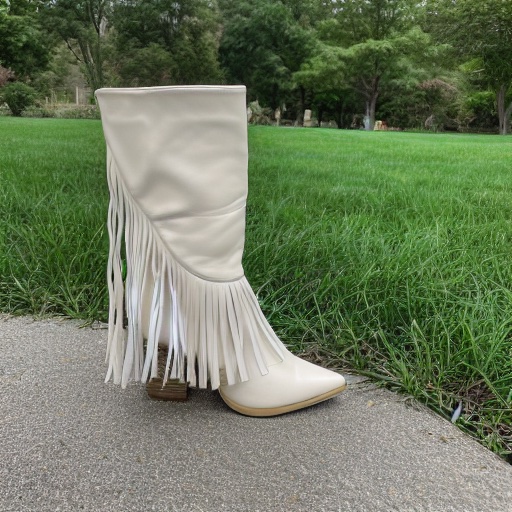} &
        \includegraphics[width=0.16\textwidth]{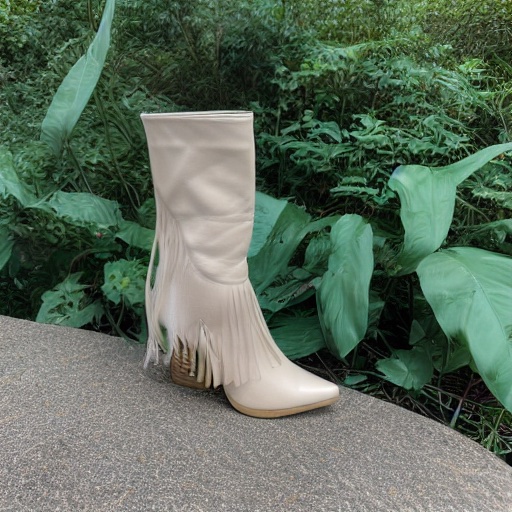} &
    \includegraphics[width=0.16\textwidth]{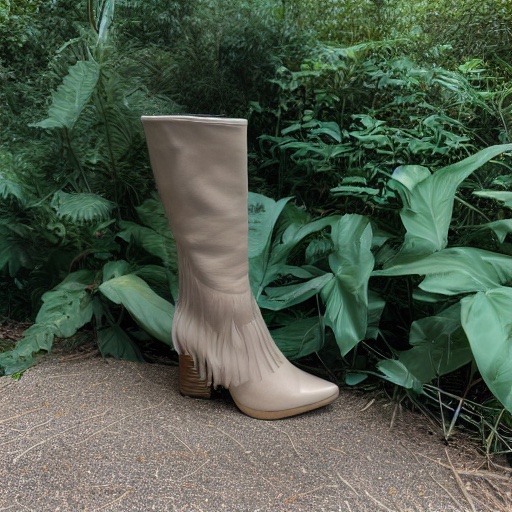} \\
    \end{tabular}
     \vspace{-5pt} 
     \caption*{a $boot^*$ in the jungle}
\end{subfigure}
\begin{subfigure}{0.6\textwidth}
    \centering
    \setlength{\tabcolsep}{2pt}
    \begin{tabular}{cccccc}
    \includegraphics[width=0.16\textwidth]{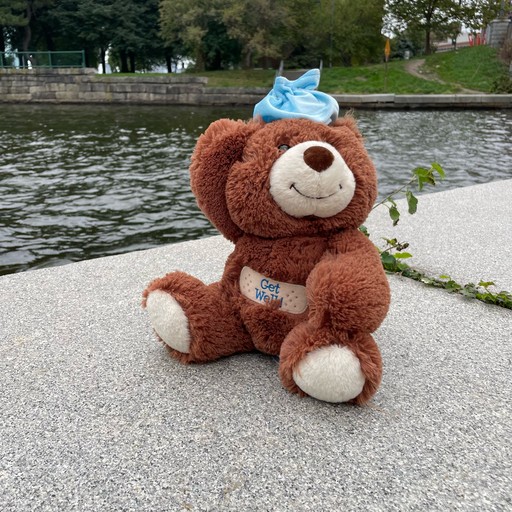} &
     \includegraphics[width=0.16\textwidth]{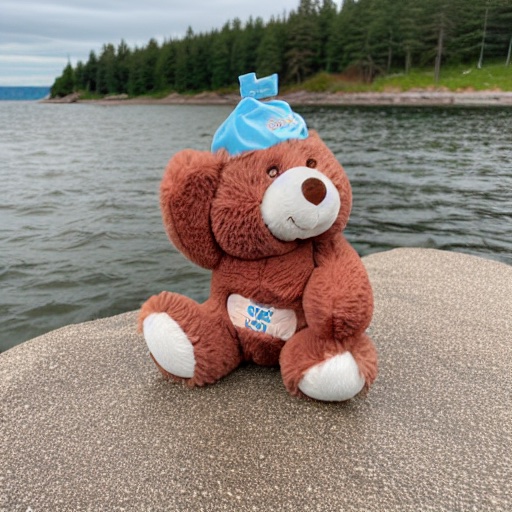} &
      \includegraphics[width=0.16\textwidth]{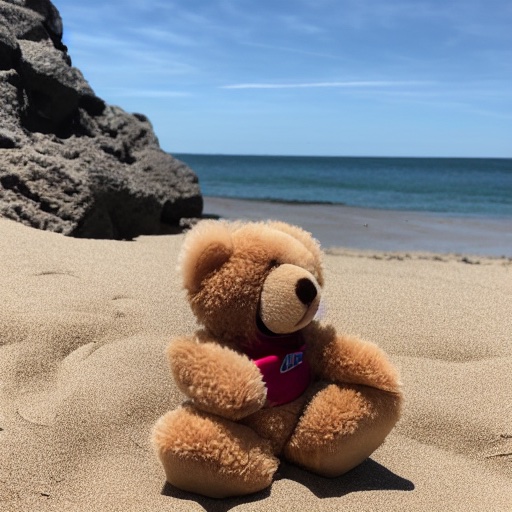} &
       \includegraphics[width=0.16\textwidth]{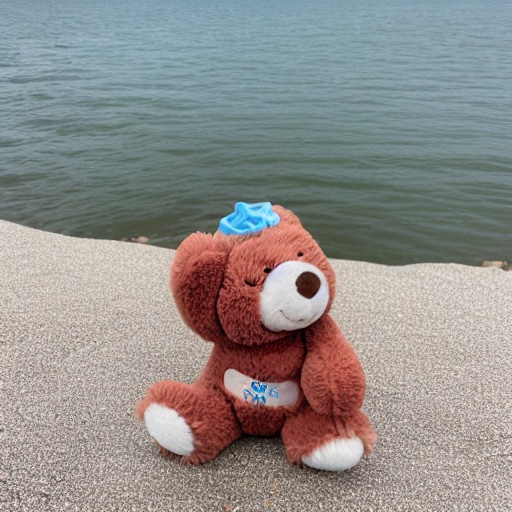} &
        \includegraphics[width=0.16\textwidth]{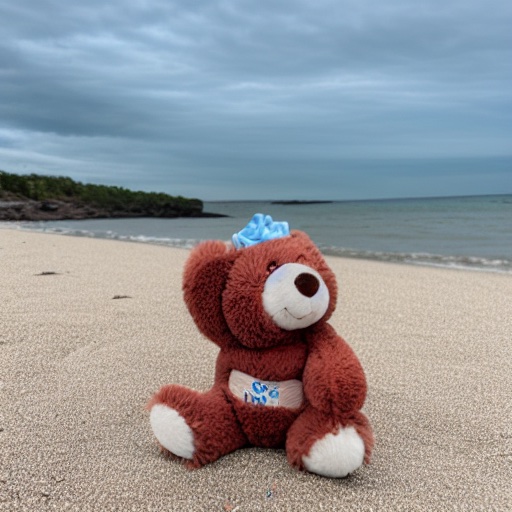} &
    \includegraphics[width=0.16\textwidth]{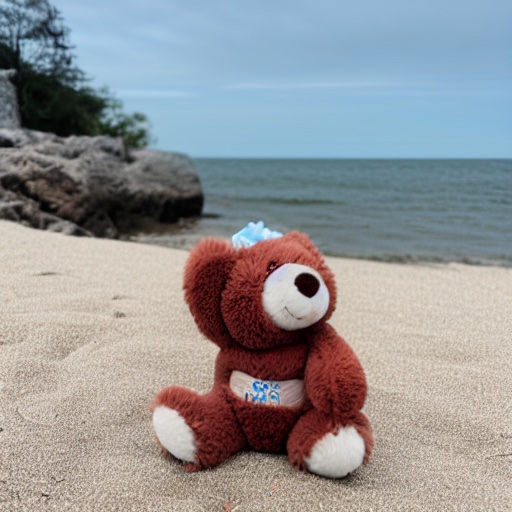} \\
    \end{tabular}
     \vspace{-5pt} 
     \caption*{a \textit{stuffed animal$^*$} on the beach}
\end{subfigure}
    \caption{Effect of cross attention guidance scale $\alpha$, for the same guidance and edit models and classifier-free guidance}
    \label{fig:cag_effect-2}
      \vspace{-10pt} 
\end{figure*}

\begin{figure*}
\centering

\begin{subfigure}{0.49\textwidth}
    \centering
    \setlength{\tabcolsep}{11pt}
    \begin{tabular}{cccc}
    Input & Overfit & Underfit \\
    \includegraphics[width=0.2\textwidth]{images/gs_and_cag/input.jpg} &
    \includegraphics[width=0.2\textwidth]{images/gs_and_cag/overfit.jpg} &
    \includegraphics[width=0.2\textwidth]{images/gs_and_cag/guidance.jpg} \\
    \end{tabular}
    \caption{Input training image, overfit (edit) image and underfit (guidance) image}
     \vspace{10pt} 
\end{subfigure}

 \begin{subfigure}{0.49\textwidth}
    \centering
    \setlength{\tabcolsep}{2pt}
    \begin{tabular}{cccc}
    & $\alpha=0.0$ & $\alpha=0.1$ & $\alpha=0.2$ \\
   gs = 2.0 &
   \includegraphics[width=0.2\linewidth]{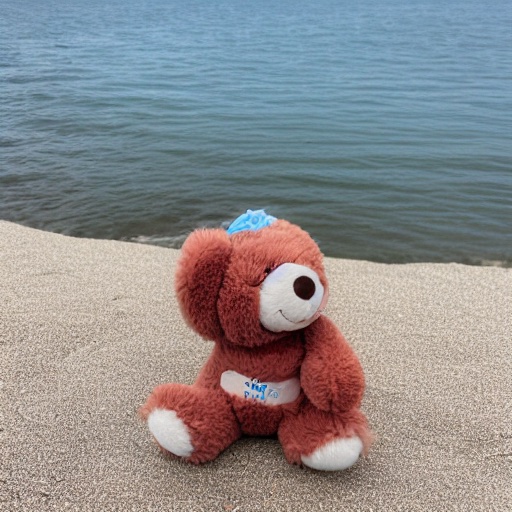} &
    \includegraphics[width=0.2\linewidth]{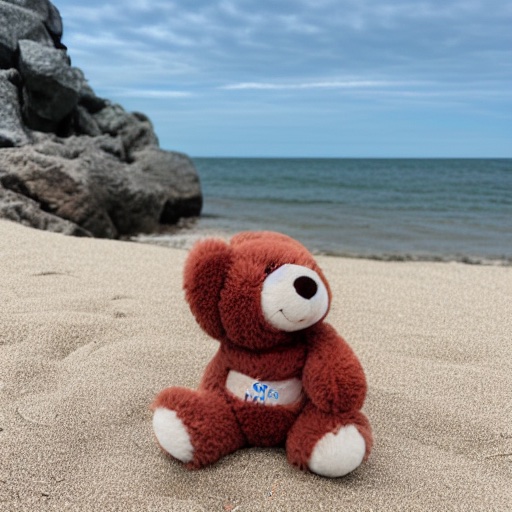} &
    \includegraphics[width=0.2\linewidth]{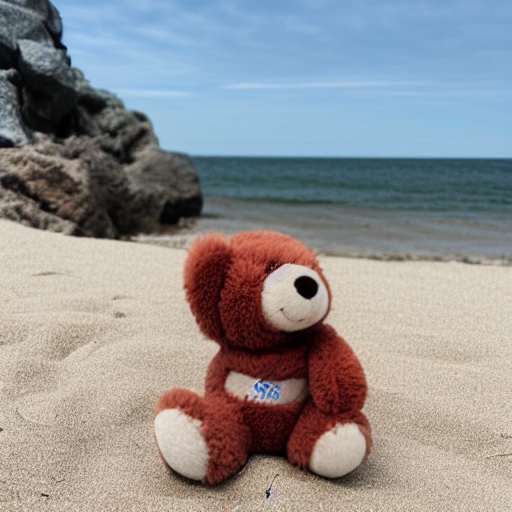} \\

   gs = 3.0 &
   \includegraphics[width=0.2\textwidth]{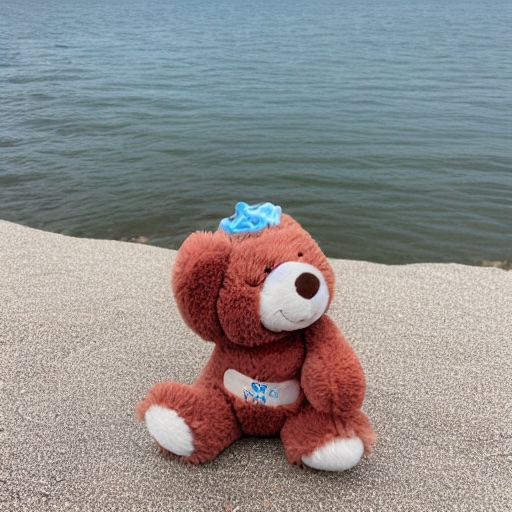} &
    \includegraphics[width=0.2\textwidth]{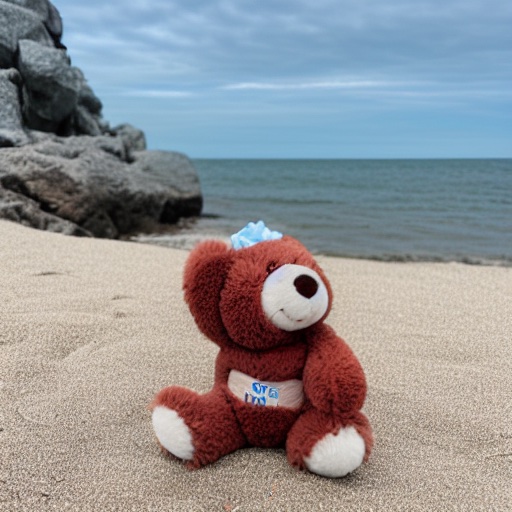} &
    \includegraphics[width=0.2\textwidth]{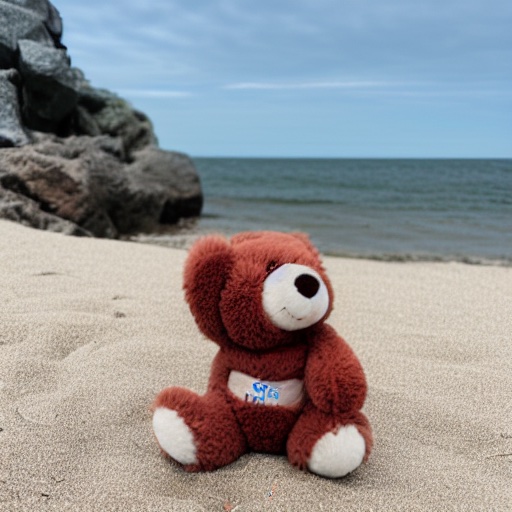} \\

   gs = 4.0 &
   \includegraphics[width=0.2\textwidth]{images/gs_and_cag/synthetic_image_9_translated_4.0_0.0.jpg} &
    \includegraphics[width=0.2\textwidth]{images/gs_and_cag/synthetic_image_9_translated_4.0_0.1.jpg} &
    \includegraphics[width=0.2\textwidth]{images/gs_and_cag/synthetic_image_9_translated_4.0_0.2.jpg} \\
    \end{tabular}
    \vspace{5pt} 
    \captionsetup{justification=centering}
    \caption{Our results varying classifier-free guidance scale (gs) and cross attention guidance scale ($\alpha$)}
\end{subfigure}
\caption{Effect of varying classifier-free guidance scale (gs) and cross attention guidance scale ($\alpha$) for the same guidance and edit models and prompt ``a \textit{stuffed animal$^*$} on the beach''. Increasing classifier-free guidance improves subject fidelity, while increasing cross attention guidance increases adherence to the layout of the underfit (guidance) image.}
\label{fig:gs_ags}
\end{figure*}

In \cref{fig:cag_effect-2}, we present more examples of the effect of varying cross attention guidance scale, in addition to \cref{fig:cag_effect} of the main text. In \cref{fig:gs_ags}, we present the effect of varying both cross attention guidance scale and classifier-free guidance, for the same guidance and edit models.

\section{Implementation details}
\label{sec:implementation-details}
For experiments in \cref{sec:experiments_and_results} of the main text, we use the pre-trained Stable Diffusion v1.5 model \cite{rombach2022high} and the SDXL model \cite{podell2023sdxl}. We use the HuggingFace Diffusers \cite{huggingface} implementation and the hyperparameters recommended by the authors. For DreamBooth, we use a learning rate of $5e^{-6}$ and the rare token ``sks'' to represent the specific subject during fine-tuning.  For Custom Diffusion, we use a learning rate of $1e^{-5}$, scaled with effective batch size. For regularization, we use 1000 images of the subject's category generated by the pre-trained model, with a prior preservation weight of 1.0. We use 50 steps of DDIM forward process for all methods. 

We apply our approach, DreamBlend, on results of classical DreamBooth tuning, full fine-tuning for SDv1.5 and LoRA for SDXL. For all subjects, we designate the models at step 100 and step 200 as edit models and all models with lower steps as guidance models. For the step 100 edit model, we use a classifier-free guidance scale of 3.0 and a cross attention guidance scale of 0.1 while for the step 200 edit model, we use 2.0 and 0.07, respectively. As the step 200 model has learnt the subject better, it can achieve higher subject fidelity with lower classifier-free guidance.

For calculating metrics, we use the CLIP \cite{radford2021learning} ViT-B/32 model for CLIP-I and CLIP-T and DINO \cite{caron2021emerging} ViT-S/16 model for DINO metric. Prior to computing text embeddings, we remove any rare token, such as ``sks" from the prompt.

\section{Comparison with non-fine-tuning based approaches}
\label{sec:non-fine-tuning-methods}
\begin{figure*}
\centering

\begin{subfigure}{0.24\textwidth}
    \centering
    \setlength{\tabcolsep}{2pt}
     \caption*{Input}
    \begin{tabular}{cc}
    \includegraphics[width=0.6\textwidth]{images/cat_statue_input/2.jpeg} &
    \raisebox{0.32\textwidth} {\begin{tabular}[t]{@{}c@{}} 
    \includegraphics[width=0.28\textwidth]{images/cat_statue_input/1.jpeg} \\
    \includegraphics[width=0.28\textwidth]{images/cat_statue_input/6.jpeg}
    \end{tabular}}
    \end{tabular}
    \label{fig:sub4}
\end{subfigure}
\begin{subfigure}{0.24\textwidth}
    \centering
    \setlength{\tabcolsep}{2pt}
    \caption*{Ours}
    \begin{tabular}{cc}
    \includegraphics[width=0.6\textwidth]{images/comparisons/supplementary/c3/ours1.jpg} &
    \raisebox{0.32\textwidth} {\begin{tabular}[t]{@{}c@{}} 
    \includegraphics[width=0.28\textwidth]{images/comparisons/supplementary/c3/ours2.jpg} \\
    \includegraphics[width=0.28\textwidth]{images/comparisons/supplementary/c3/ours3.jpg}
    \end{tabular}}
    \end{tabular}
    \label{fig:sub4}
\end{subfigure}
\begin{subfigure}{0.24\textwidth}
    \centering
    \setlength{\tabcolsep}{2pt}
    \caption*{Textual Inversion}
    \begin{tabular}{cc}
    \includegraphics[width=0.6\textwidth]{images/textual_inversion/cat_statue_blue_house/image1.jpg} &
    \raisebox{0.32\textwidth} {\begin{tabular}[t]{@{}c@{}} 
    \includegraphics[width=0.28\textwidth]{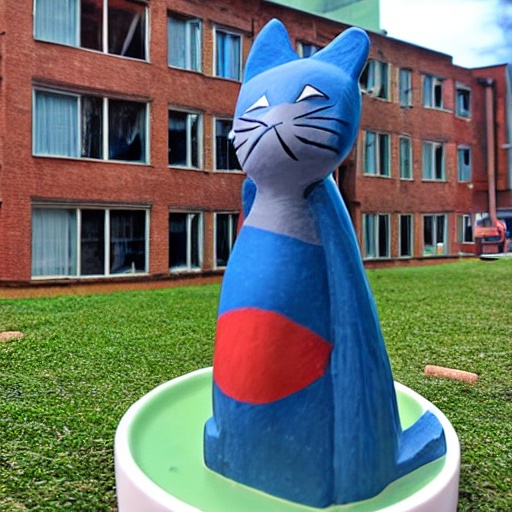} \\
    \includegraphics[width=0.28\textwidth]{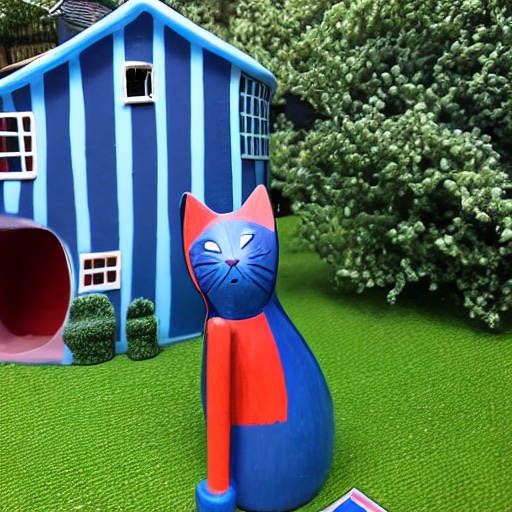}
    \end{tabular}}
    \end{tabular}
    \label{fig:sub4}
\end{subfigure}
\begin{subfigure}{0.24\textwidth}
    \centering
    \setlength{\tabcolsep}{2pt}
    \caption*{BLIP-Diffusion (ZeroShot)}
    \begin{tabular}{cc}
    \includegraphics[width=0.6\textwidth]{images/blip_diffusion/cat_statue_blue_house/image1.jpg} &
    \raisebox{0.32\textwidth} {\begin{tabular}[t]{@{}c@{}} 
    \includegraphics[width=0.28\textwidth]{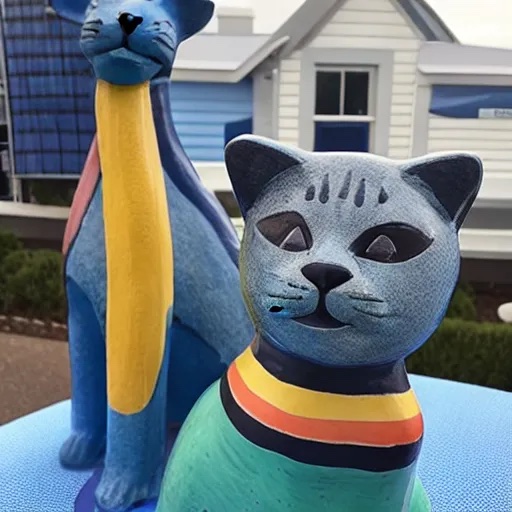} \\
    \includegraphics[width=0.28\textwidth]{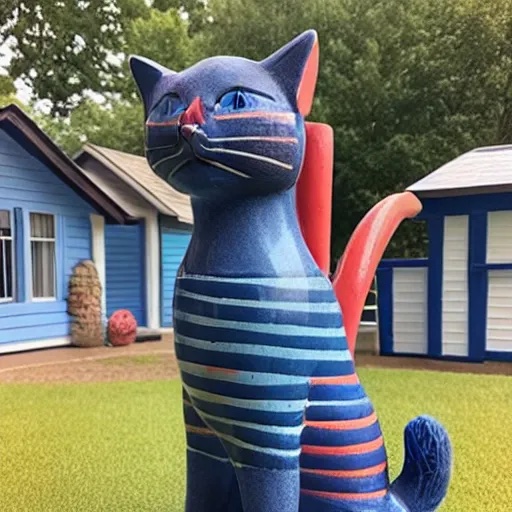}
    \end{tabular}}
    \end{tabular}
    \label{fig:sub4}
\end{subfigure}
\vspace{-12pt} 
\caption*{a $toy^*$ with a blue house in the background}

\vspace{12pt} 
\begin{subfigure}{0.24\textwidth}
    \centering
    \setlength{\tabcolsep}{2pt}
    \begin{tabular}{cc}
    \includegraphics[width=0.6\textwidth]{images/backpack_input/02.jpg} &
    \raisebox{0.32\textwidth} {\begin{tabular}[t]{@{}c@{}} 
    \includegraphics[width=0.28\textwidth]{images/backpack_input/03.jpg} \\
    \includegraphics[width=0.28\textwidth]{images/backpack_input/05.jpg}
    \end{tabular}}
    \end{tabular}
    \label{fig:sub4}
\end{subfigure}
\begin{subfigure}{0.24\textwidth}
    \centering
    \setlength{\tabcolsep}{2pt}
    \begin{tabular}{cc}
    \includegraphics[width=0.6\textwidth]{images/comparisons/supplementary/c2/ours1.jpg} &
    \raisebox{0.32\textwidth} {\begin{tabular}[t]{@{}c@{}} 
    \includegraphics[width=0.28\textwidth]{images/comparisons/supplementary/c2/ours2.jpg} \\
    \includegraphics[width=0.28\textwidth]{images/comparisons/supplementary/c2/ours3.jpg}
    \end{tabular}}
    \end{tabular}
    \label{fig:sub4}
\end{subfigure}
\begin{subfigure}{0.24\textwidth}
    \centering
    \setlength{\tabcolsep}{2pt}
    \begin{tabular}{cc}
    \includegraphics[width=0.6\textwidth]{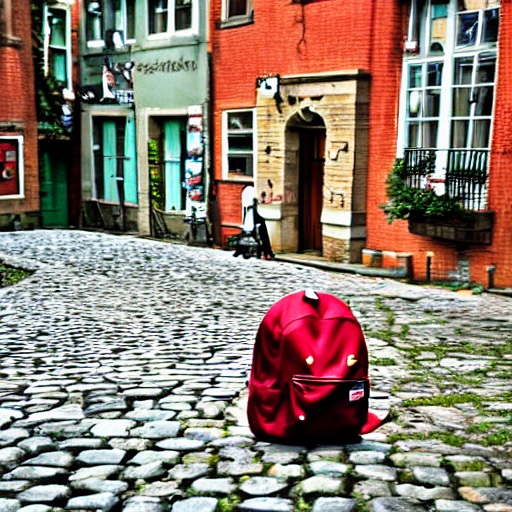} &
    \raisebox{0.32\textwidth} {\begin{tabular}[t]{@{}c@{}} 
    \includegraphics[width=0.28\textwidth]{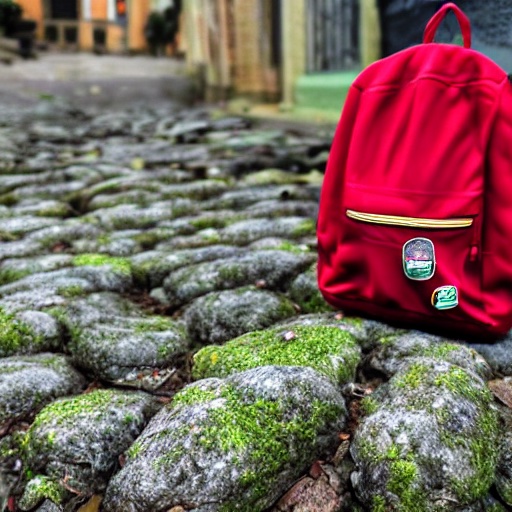} \\
    \includegraphics[width=0.28\textwidth]{images/textual_inversion/backpack_cobblestone/image2.jpg}
    \end{tabular}}
    \end{tabular}
    \label{fig:sub4}
\end{subfigure}
\begin{subfigure}{0.24\textwidth}
    \centering
    \setlength{\tabcolsep}{2pt}
    \begin{tabular}{cc}
    \includegraphics[width=0.6\textwidth]{images/blip_diffusion/backpack_cobblestone/image1.jpg} &
    \raisebox{0.32\textwidth} {\begin{tabular}[t]{@{}c@{}} 
    \includegraphics[width=0.28\textwidth]{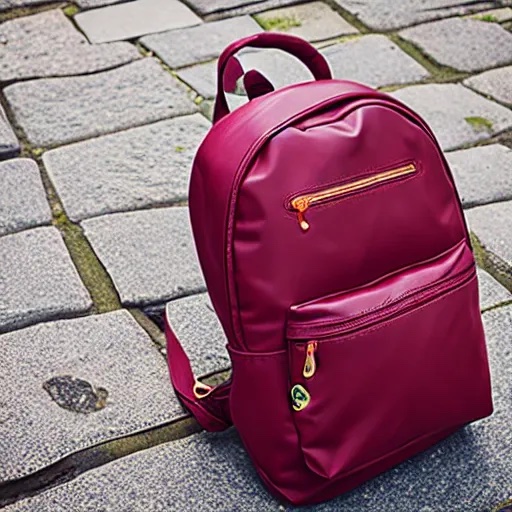} \\
    \includegraphics[width=0.28\textwidth]{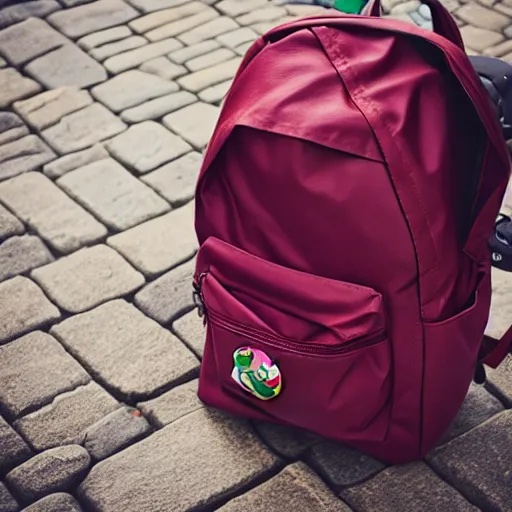}
    \end{tabular}}
    \end{tabular}
    \label{fig:sub4}
\end{subfigure}
\vspace{-12pt} 
\caption*{a $backpack^*$ on a cobblestone street}

\vspace{12pt} 
\begin{subfigure}{0.24\textwidth}
    \centering
    \setlength{\tabcolsep}{2pt}
    \begin{tabular}{cc}
    \includegraphics[width=0.6\textwidth]{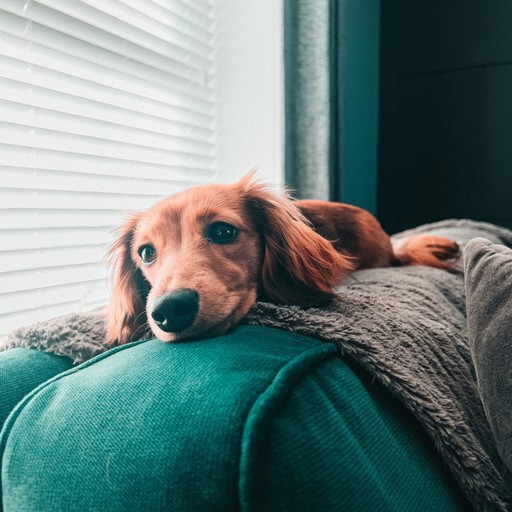} &
    \raisebox{0.32\textwidth} {\begin{tabular}[t]{@{}c@{}} 
    \includegraphics[width=0.28\textwidth]{images/dog5_input/02.jpg} \\
    \includegraphics[width=0.28\textwidth]{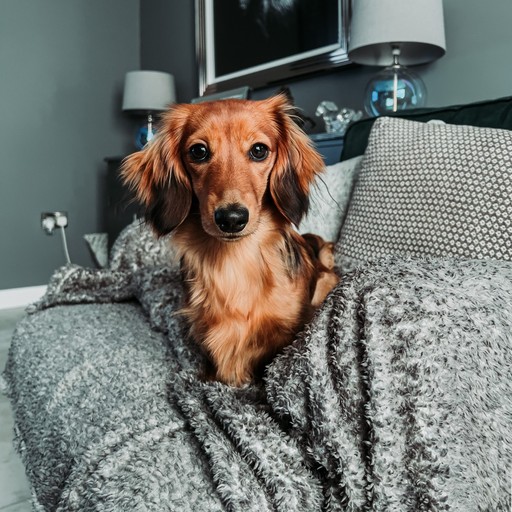}
    \end{tabular}}
    \end{tabular}
    \label{fig:sub4}
\end{subfigure}
\begin{subfigure}{0.24\textwidth}
    \centering
    \setlength{\tabcolsep}{2pt}
    \begin{tabular}{cc}
    \includegraphics[width=0.6\textwidth]{images/comparisons/dog5_beach_ours/image_1.jpg} &
    \raisebox{0.32\textwidth} {\begin{tabular}[t]{@{}c@{}} 
    \includegraphics[width=0.28\textwidth]{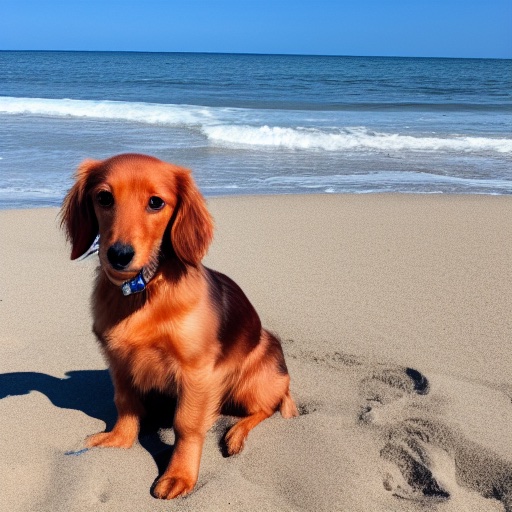} \\
    \includegraphics[width=0.28\textwidth]{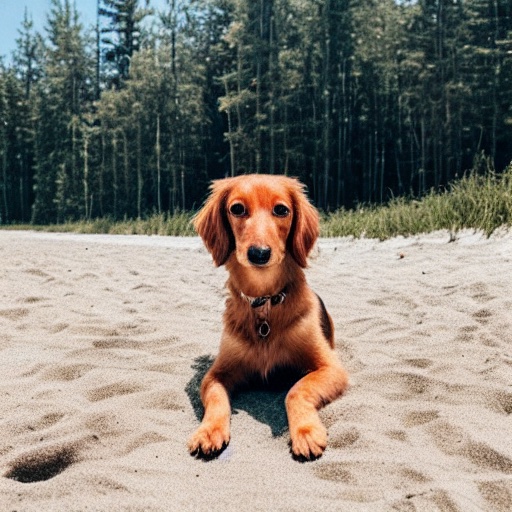}
    \end{tabular}}
    \end{tabular}
    \label{fig:sub4}
\end{subfigure}
\begin{subfigure}{0.24\textwidth}
    \centering
    \setlength{\tabcolsep}{2pt}
    \begin{tabular}{cc}
    \includegraphics[width=0.6\textwidth]{images/textual_inversion/dog5_beach/image2.jpg} &
    \raisebox{0.32\textwidth} {\begin{tabular}[t]{@{}c@{}} 
    \includegraphics[width=0.28\textwidth]{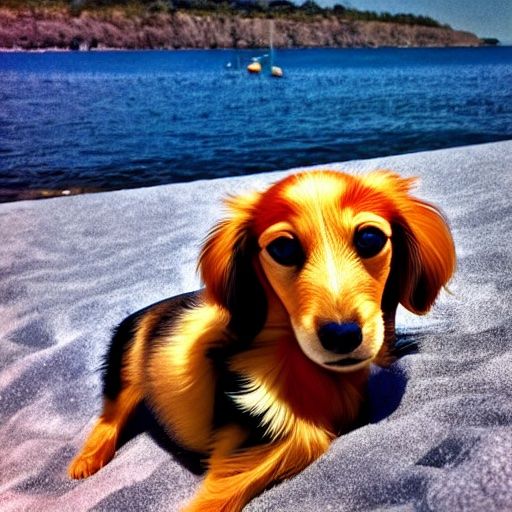} \\
    \includegraphics[width=0.28\textwidth]{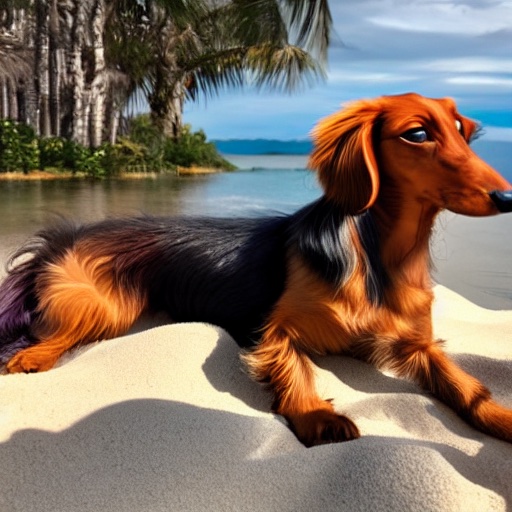}
    \end{tabular}}
    \end{tabular}
    \label{fig:sub4}
\end{subfigure}
\begin{subfigure}{0.24\textwidth}
    \centering
    \setlength{\tabcolsep}{2pt}
    \begin{tabular}{cc}
    \includegraphics[width=0.6\textwidth]{images/blip_diffusion/dog5_beach/image2.jpg} &
    \raisebox{0.32\textwidth} {\begin{tabular}[t]{@{}c@{}} 
    \includegraphics[width=0.28\textwidth]{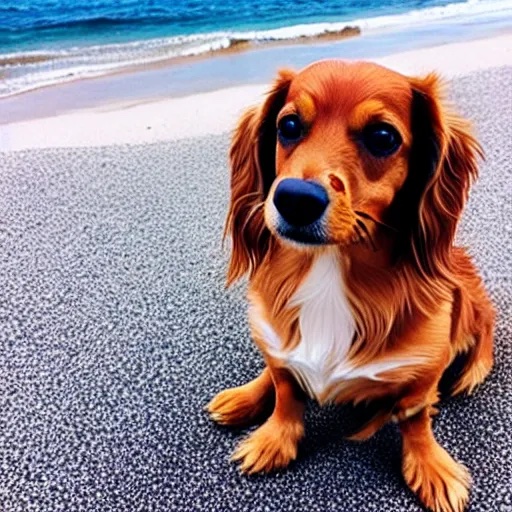} \\
    \includegraphics[width=0.28\textwidth]{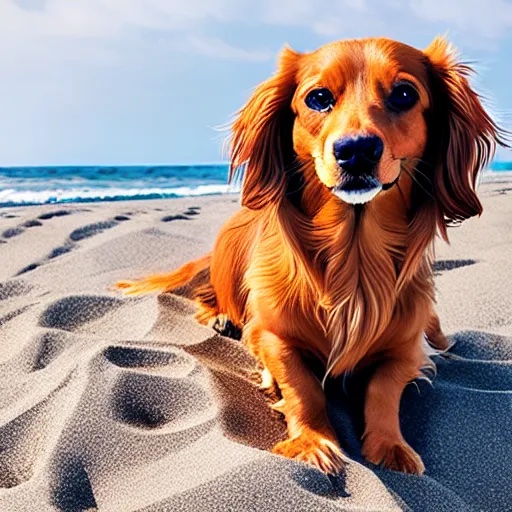}
    \end{tabular}}
    \end{tabular}
    \label{fig:sub4}
\end{subfigure}
\vspace{-12pt} 
\caption*{a $dog^*$ on a beach}

\caption{Comparison with non-fine-tuning based methods Textual Inversion and BLIP-Diffusion}
\label{fig:non-finetuning-methods}
\end{figure*}

\begin{figure*}
\centering

\begin{subfigure}{0.24\textwidth}
    \centering
    \setlength{\tabcolsep}{2pt}
     \caption*{Input}
    \begin{tabular}{cc}
    \includegraphics[width=0.6\textwidth]{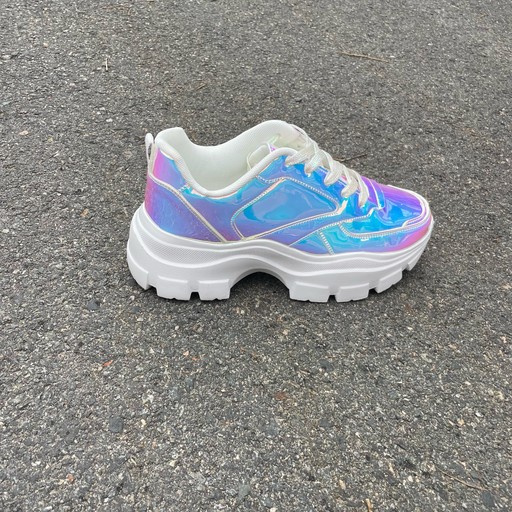} &
    \raisebox{0.32\textwidth} {\begin{tabular}[t]{@{}c@{}} 
    \includegraphics[width=0.28\textwidth]{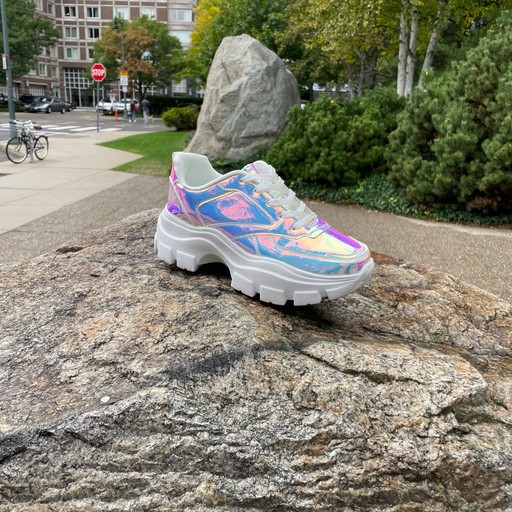} \\
    \includegraphics[width=0.28\textwidth]{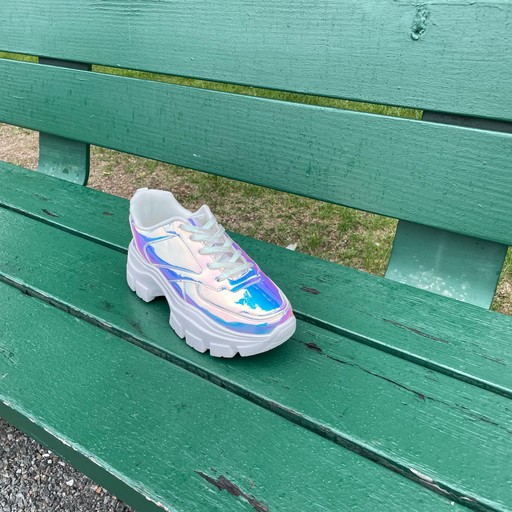}
    \end{tabular}}
    \end{tabular}
    \label{fig:sub4}
\end{subfigure}
\begin{subfigure}{0.24\textwidth}
    \centering
    \setlength{\tabcolsep}{2pt}
    \caption*{Ours}
    \begin{tabular}{cc}
    \includegraphics[width=0.6\textwidth]{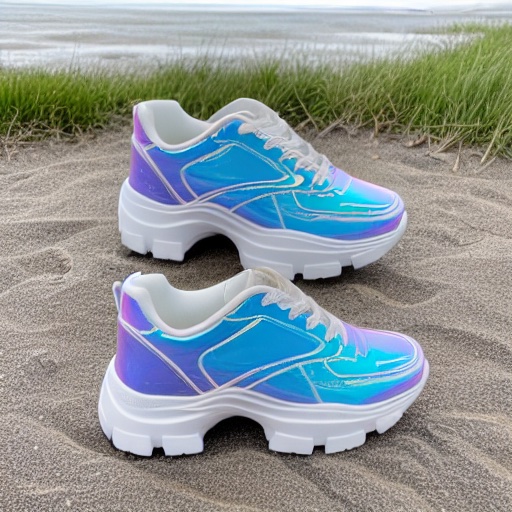} &
    \raisebox{0.32\textwidth} {\begin{tabular}[t]{@{}c@{}} 
    \includegraphics[width=0.28\textwidth]{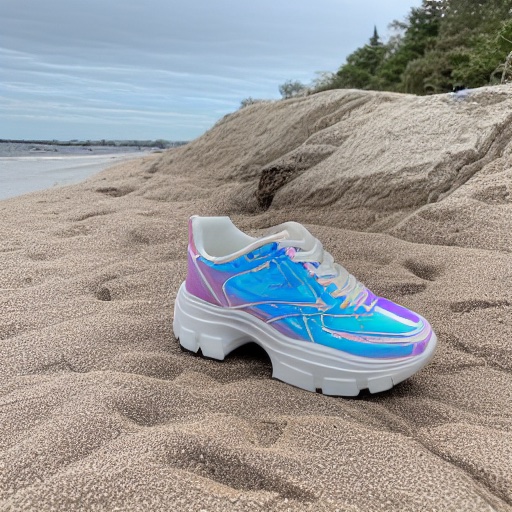} \\
    \includegraphics[width=0.28\textwidth]{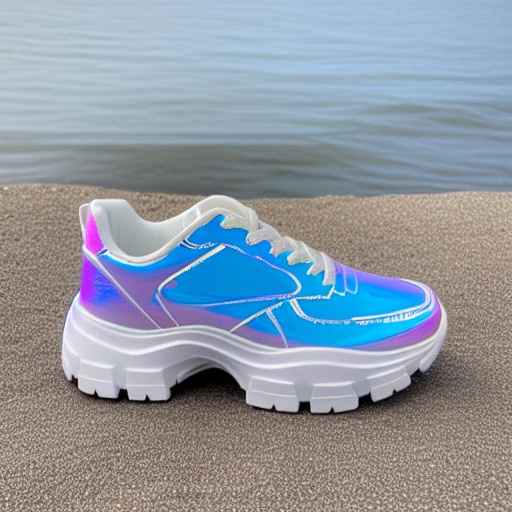}
    \end{tabular}}
    \end{tabular}
    \label{fig:sub4}
\end{subfigure}
\begin{subfigure}{0.24\textwidth}
    \centering
    \setlength{\tabcolsep}{2pt}
    \caption*{IP-Adapter}
    \begin{tabular}{cc}
    \includegraphics[width=0.6\textwidth]{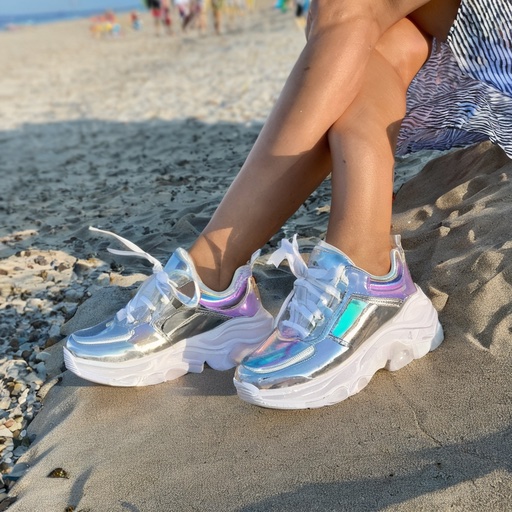} &
    \raisebox{0.32\textwidth} {\begin{tabular}[t]{@{}c@{}} 
    \includegraphics[width=0.28\textwidth]{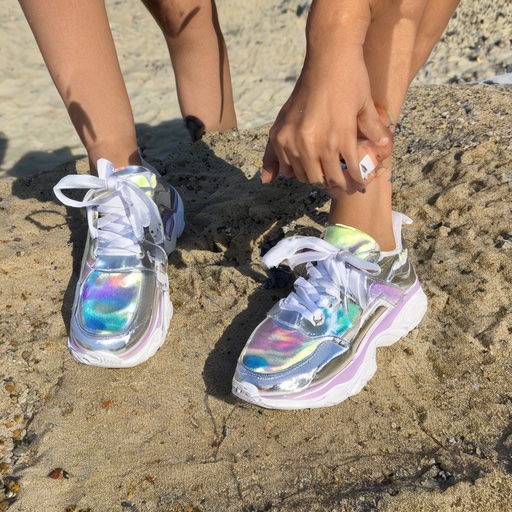} \\
    \includegraphics[width=0.28\textwidth]{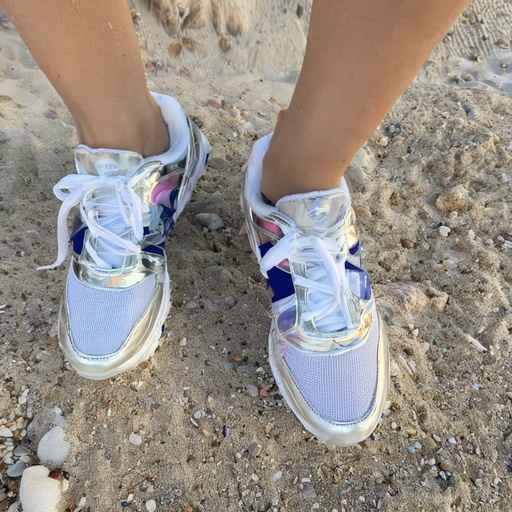}
    \end{tabular}}
    \end{tabular}
    \label{fig:sub4}
\end{subfigure}
\begin{subfigure}{0.24\textwidth}
    \centering
    \setlength{\tabcolsep}{2pt}
    \caption*{AnyDoor}
    \begin{tabular}{cc}
    \includegraphics[width=0.6\textwidth]{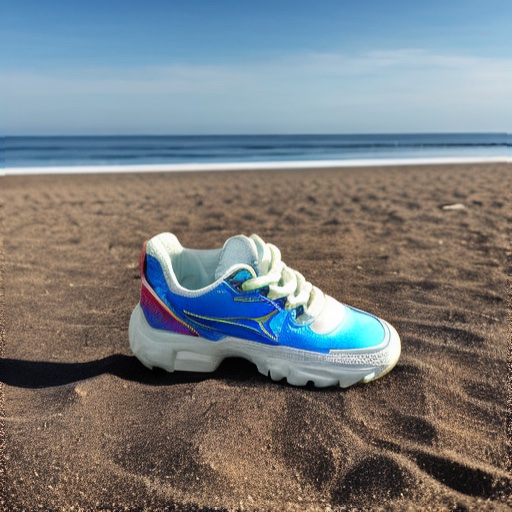} &
    \raisebox{0.32\textwidth} {\begin{tabular}[t]{@{}c@{}} 
    \includegraphics[width=0.28\textwidth]{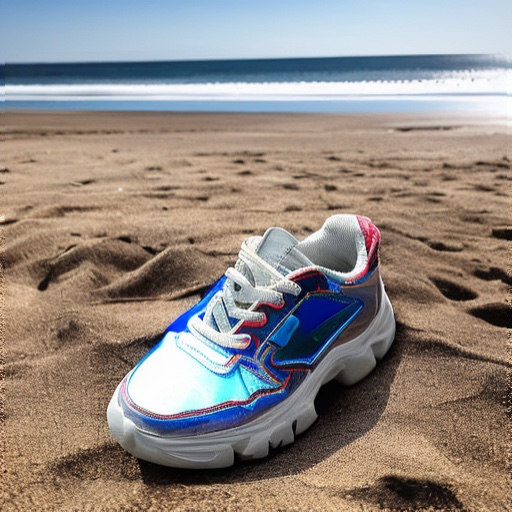} \\
    \includegraphics[width=0.28\textwidth]{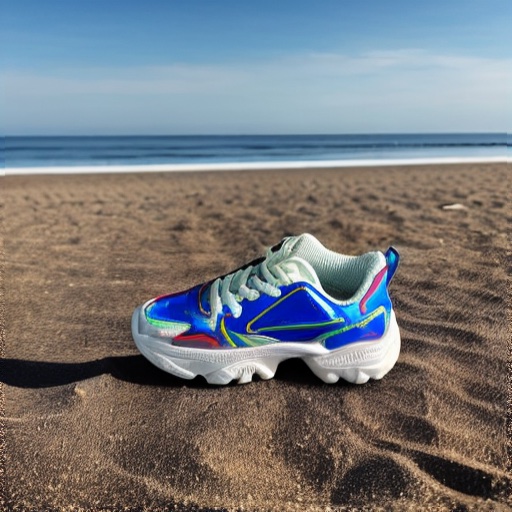}
    \end{tabular}}
    \end{tabular}
    \label{fig:sub4}
\end{subfigure}
\vspace{-12pt} 
\caption*{a $sneaker^*$ on the beach}

\vspace{12pt} 
\begin{subfigure}{0.24\textwidth}
    \centering
    \setlength{\tabcolsep}{2pt}
    \begin{tabular}{cc}
    \includegraphics[width=0.6\textwidth]{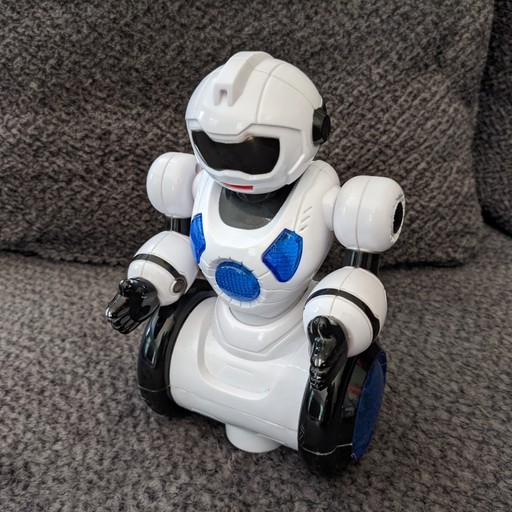} &
    \raisebox{0.32\textwidth} {\begin{tabular}[t]{@{}c@{}} 
    \includegraphics[width=0.28\textwidth]{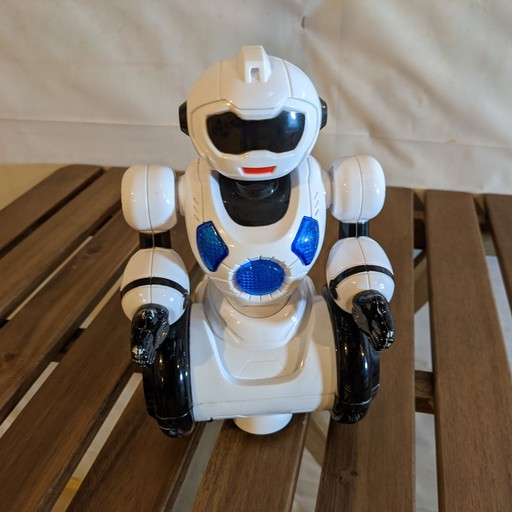} \\
    \includegraphics[width=0.28\textwidth]{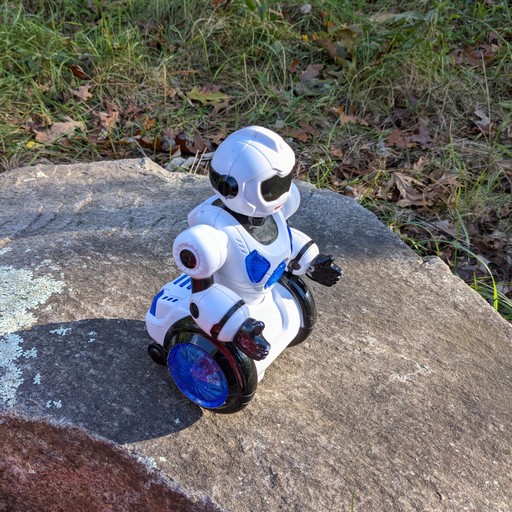}
    \end{tabular}}
    \end{tabular}
    \label{fig:sub4}
\end{subfigure}
\begin{subfigure}{0.24\textwidth}
    \centering
    \setlength{\tabcolsep}{2pt}
    \begin{tabular}{cc}
    \includegraphics[width=0.6\textwidth]{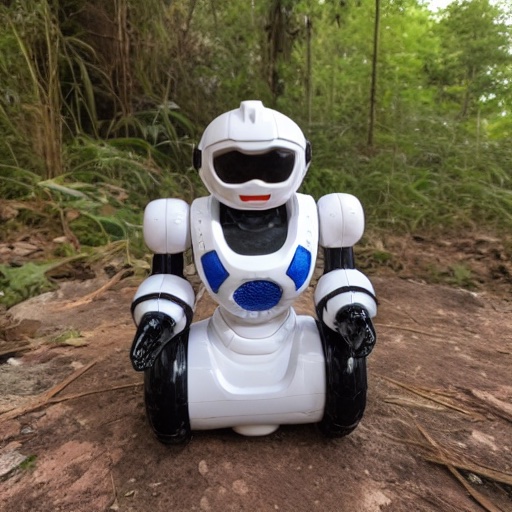} &
    \raisebox{0.32\textwidth} {\begin{tabular}[t]{@{}c@{}} 
    \includegraphics[width=0.28\textwidth]{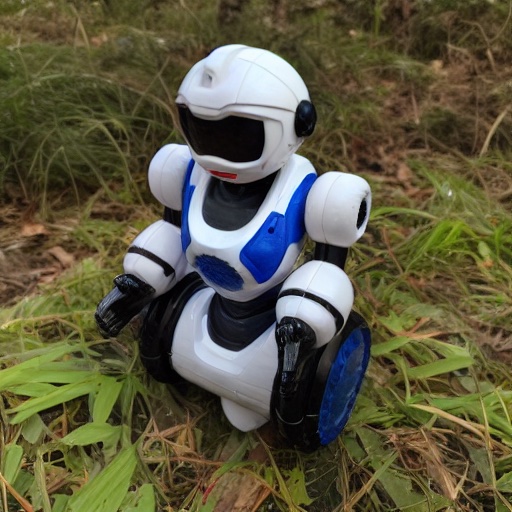} \\
    \includegraphics[width=0.28\textwidth]{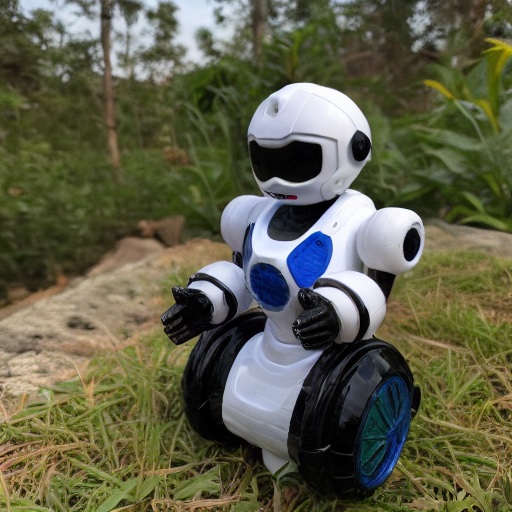}
    \end{tabular}}
    \end{tabular}
    \label{fig:sub4}
\end{subfigure}
\begin{subfigure}{0.24\textwidth}
    \centering
    \setlength{\tabcolsep}{2pt}
    \begin{tabular}{cc}
    \includegraphics[width=0.6\textwidth]{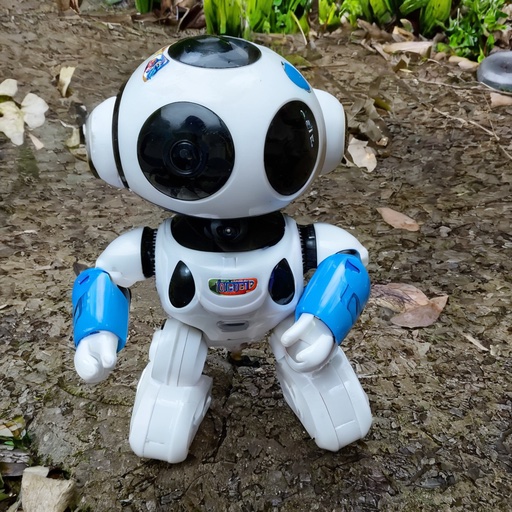} &
    \raisebox{0.32\textwidth} {\begin{tabular}[t]{@{}c@{}} 
    \includegraphics[width=0.28\textwidth]{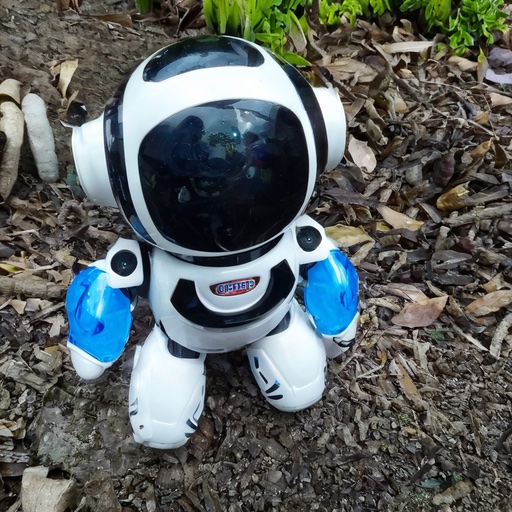} \\
    \includegraphics[width=0.28\textwidth]{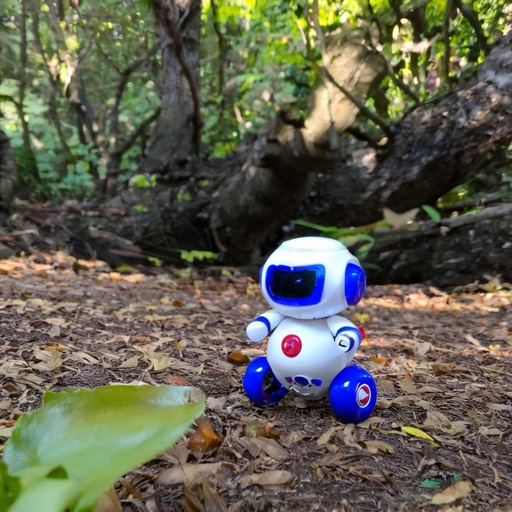}
    \end{tabular}}
    \end{tabular}
    \label{fig:sub4}
\end{subfigure}
\begin{subfigure}{0.24\textwidth}
    \centering
    \setlength{\tabcolsep}{2pt}
    \begin{tabular}{cc}
    \includegraphics[width=0.6\textwidth]{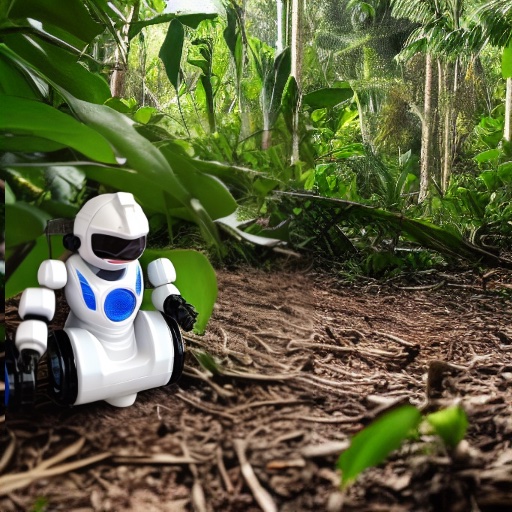} &
    \raisebox{0.32\textwidth} {\begin{tabular}[t]{@{}c@{}} 
    \includegraphics[width=0.28\textwidth]{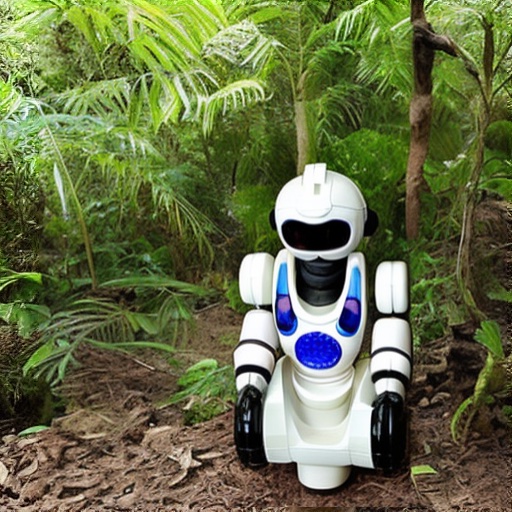} \\
    \includegraphics[width=0.28\textwidth]{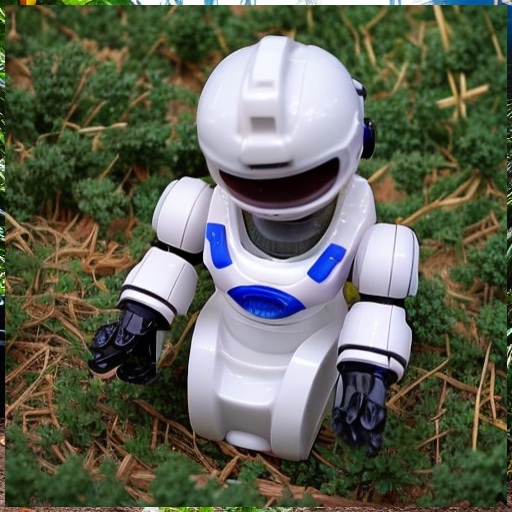}
    \end{tabular}}
    \end{tabular}
    \label{fig:sub4}
\end{subfigure}
\vspace{-12pt} 
\caption*{a $toy^*$ in the jungle}

\vspace{12pt} 
\begin{subfigure}{0.24\textwidth}
    \centering
    \setlength{\tabcolsep}{2pt}
    \begin{tabular}{cc}
    \includegraphics[width=0.6\textwidth]{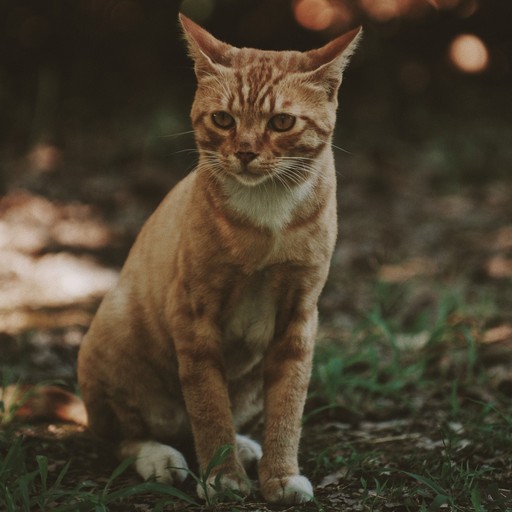} &
    \raisebox{0.32\textwidth} {\begin{tabular}[t]{@{}c@{}} 
    \includegraphics[width=0.28\textwidth]{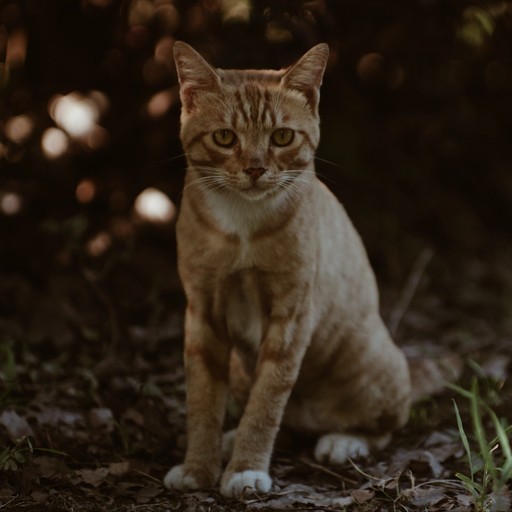} \\
    \includegraphics[width=0.28\textwidth]{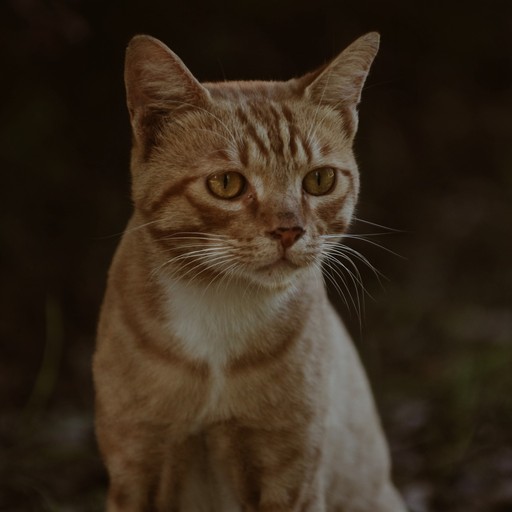}
    \end{tabular}}
    \end{tabular}
    \label{fig:sub4}
\end{subfigure}
\begin{subfigure}{0.24\textwidth}
    \centering
    \setlength{\tabcolsep}{2pt}
    \begin{tabular}{cc}
    \includegraphics[width=0.6\textwidth]{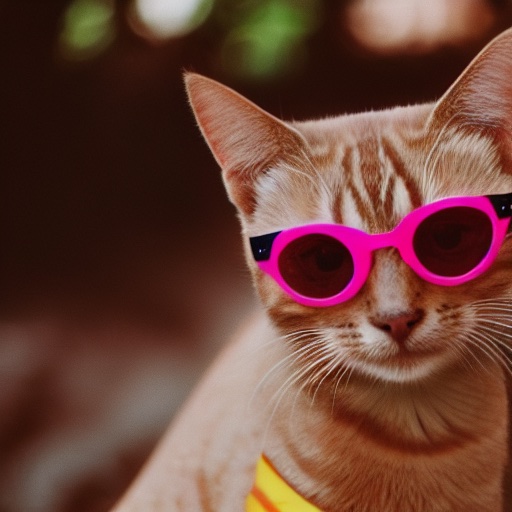} &
    \raisebox{0.32\textwidth} {\begin{tabular}[t]{@{}c@{}} 
    \includegraphics[width=0.28\textwidth]{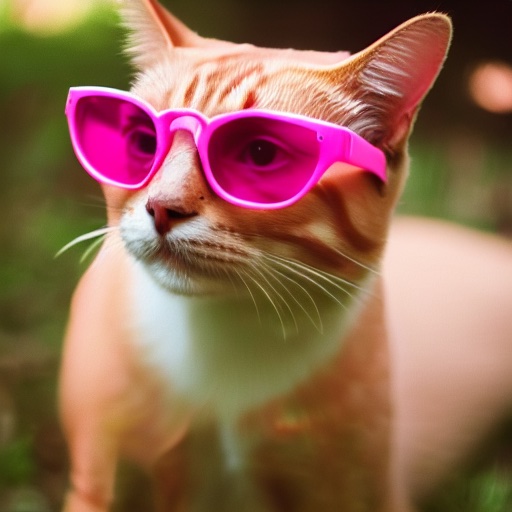} \\
    \includegraphics[width=0.28\textwidth]{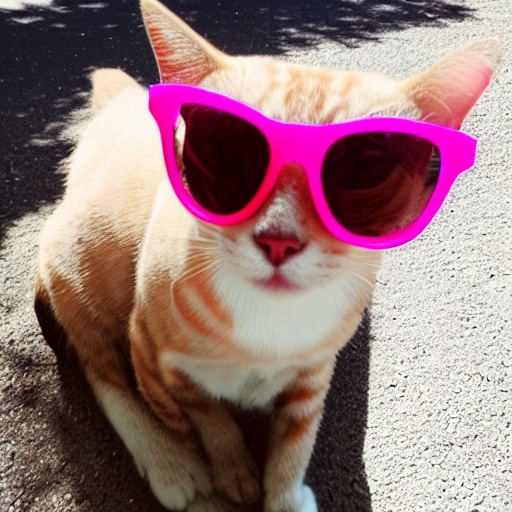}
    \end{tabular}}
    \end{tabular}
    \label{fig:sub4}
\end{subfigure}
\begin{subfigure}{0.24\textwidth}
    \centering
    \setlength{\tabcolsep}{2pt}
    \begin{tabular}{cc}
    \includegraphics[width=0.6\textwidth]{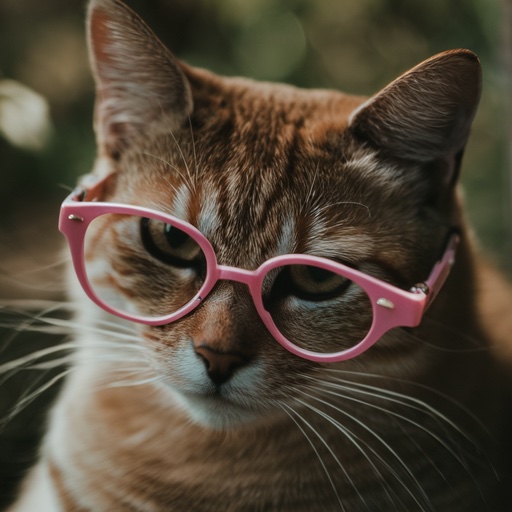} &
    \raisebox{0.32\textwidth} {\begin{tabular}[t]{@{}c@{}} 
    \includegraphics[width=0.28\textwidth]{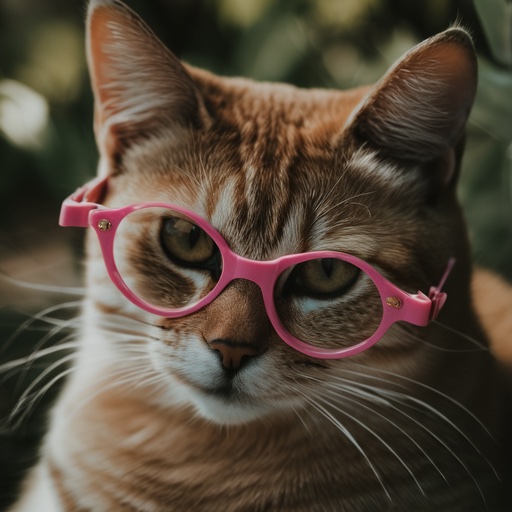} \\
    \includegraphics[width=0.28\textwidth]{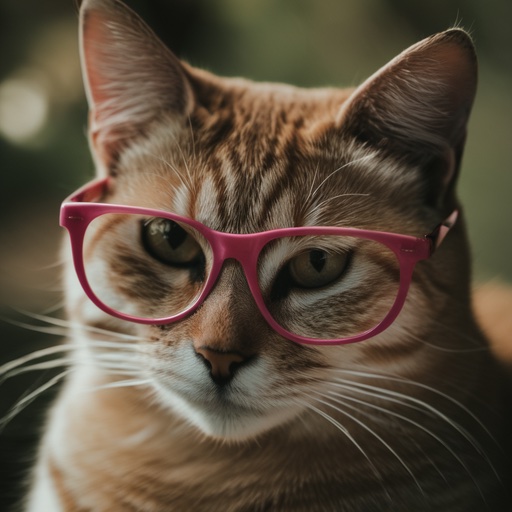}
    \end{tabular}}
    \end{tabular}
    \label{fig:sub4}
\end{subfigure}
\begin{subfigure}{0.24\textwidth}
    \centering
    \setlength{\tabcolsep}{2pt}
    \begin{tabular}{cc}
    \includegraphics[width=0.6\textwidth]{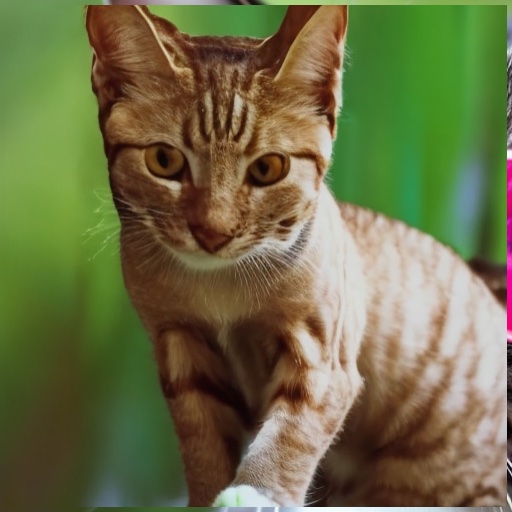} &
    \raisebox{0.32\textwidth} {\begin{tabular}[t]{@{}c@{}} 
    \includegraphics[width=0.28\textwidth]{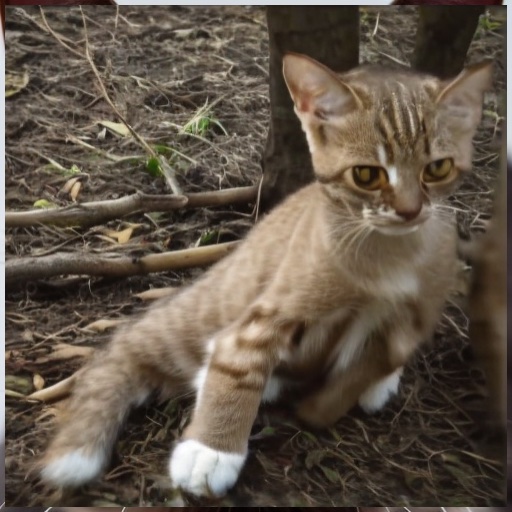} \\
    \includegraphics[width=0.28\textwidth]{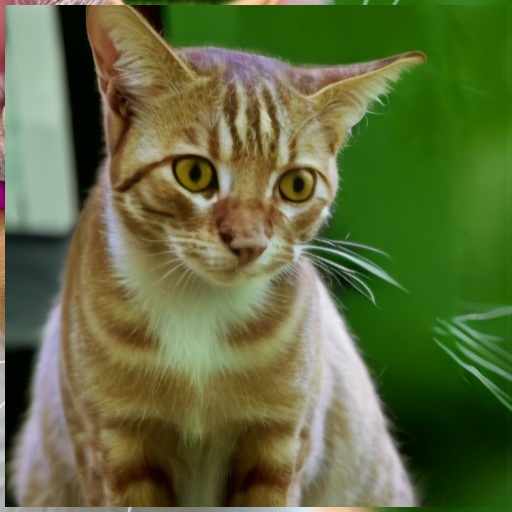}
    \end{tabular}}
    \end{tabular}
    \label{fig:sub4}
\end{subfigure}
\vspace{-12pt} 
\caption*{a $cat^*$ wearing pink glasses}

\caption{Comparison with non-fine-tuning based methods IP-Adapter and AnyDoor}
\label{fig:non-finetuning-methods-2}
\end{figure*}

In this section, we present qualitative comparisons to non-fine-tuning based methods, in addition to \cref{fig:non-finetune-comparison} of the main paper. Comparisons to Textual Inversion and BLIP-Diffusion are in \cref{fig:non-finetuning-methods}. Comparisons to IP-Adapter and AnyDoor are in \cref{fig:non-finetuning-methods-2}.
For Textual Inversion, we trained the word embedding for the recommended 3000 steps, logging results every 500 steps and present the best results. For AnyDoor, we generated the background images using the pre-trained StableDiffusion model and used CLIPSeg \cite{luddecke2022image} to generate the segmentation masks.

\begin{table*}
  \centering
  {\small{
  \begin{tabular}{@{}lccccc@{}}
    \toprule
     Preference study & Study result & Binomial p value & CI (ours) & CI (baseline) \\
    \midrule
    Ours over DB Overall & 61.11 & 2.51e-12 & [58.08, 64.13] & [35.87, 41.92] \\
    Ours over DB Diversity & 61.82 & 2.51e-09 & [58.05, 65.58] & [34.42, 41.95] \\
    Ours over CD Overall & 70.16 & 2.44e-20 & [66.21, 74.10] & [25.90, 33.79] \\
    Ours over CD Diversity & 72.70 & 1.46e-35 & [69.47, 75.94] & [24.06, 30.53] \\
    \bottomrule
  \end{tabular}
  }}
  \caption{Results of one-sample binomial tests on human preference study results. CI denotes the 95\% Adjusted Wald Confidence Intervals. The lower bound of the CI for our approach is greater than 50\% in all studies. Also, the confidence intervals for our approach and those for the baseline approach are well separated in all studies. Further, the exact binomial p value is very low in all studies.}
  \label{tab:binomial}
\end{table*}

\begin{table*}
  \centering
  {\small{
  \begin{tabular}{@{}lcc@{}}
    \toprule
     Preference study & Chi-square statistic & P-value\\
    \midrule
    Ours over DB Overall & 24.40 & 7.82e-07 \\
    Ours over DB Diversity & 17.62 & 2.70e-05 \\
    Ours over CD Overall & 42.86 & 5.88e-11 \\
    Ours over CD Diversity & 78.27 & 8.97e-19 \\
    \bottomrule
  \end{tabular}
  }}
  \caption{Results of Chi-square goodness of fit tests on human preference study results. The P-value is very low in all studies.}
  \label{tab:chi-square}
\end{table*}

\end{document}